\documentclass[sigconf]{acmart}
\pdfoutput=1

\copyrightyear{2023}
\acmYear{2023}
\setcopyright{acmlicensed}\acmConference[KDD '23]{Proceedings of the 29th
ACM SIGKDD Conference on Knowledge Discovery and Data Mining}{August
6--10, 2023}{Long Beach, CA, USA}
\acmBooktitle{Proceedings of the 29th ACM SIGKDD Conference on Knowledge
Discovery and Data Mining (KDD '23), August 6--10, 2023, Long Beach, CA,
USA}
\acmPrice{15.00}
\acmDOI{10.1145/3580305.3599258}
\acmISBN{979-8-4007-0103-0/23/08}

\usepackage{amsmath}
\usepackage{graphicx}
\usepackage{caption}
\usepackage{subcaption}
\usepackage[export]{adjustbox} 
\usepackage{bbding}
\usepackage{multirow}
\usepackage{enumitem}
\usepackage[ruled,vlined,linesnumbered]{algorithm2e}
\usepackage{xspace}
\usepackage{makecell}
\usepackage{lipsum}
\usepackage{balance} 

\AtBeginDocument{%
  \providecommand\BibTeX{{%
    \normalfont B\kern-0.5em{\scshape i\kern-0.25em b}\kern-0.8em\TeX}}}

\begin{document}
\newcommand{\ndatasets}{{25}\xspace}
\title{Anomaly Detection with Score Distribution Discrimination}

\author{Minqi Jiang}
\email{jiangmq95@163.com}
\orcid{0000-0003-1285-0208}
\affiliation{%
  \institution{AI Lab, Shanghai University of Finance and Economics}
  \city{Shanghai}
  \country{China}
}

\author{Songqiao Han}
\authornote{Corresponding author.}
\email{han.songqiao@shufe.edu.cn}
\orcid{0000-0002-2896-0607}
\affiliation{%
  \institution{AI Lab, Shanghai University of Finance and Economics}
  \city{Shanghai}
  \country{China}
}

\author{Hailiang Huang}
\authornotemark[1]
\email{hlhuang@shufe.edu.cn}
\orcid{0000-0002-0009-6677}
\affiliation{%
  \institution{AI Lab, Shanghai University of Finance and Economics}
  \city{Shanghai}
  \country{China}
}

\renewcommand{\shortauthors}{Minqi Jiang, Songqiao Han, \& Hailiang Huang}


\begin{abstract}
Recent studies give more attention to the anomaly detection (AD) methods that can leverage a handful of labeled anomalies along with abundant unlabeled data. These existing anomaly-informed AD methods rely on manually predefined score target(s), e.g., prior constant or margin hyperparameter(s), to realize discrimination in anomaly scores between normal and abnormal data. However, such methods would be vulnerable to the existence of anomaly contamination in the unlabeled data, and also lack adaptation to different data scenarios.

In this paper, we propose to optimize the anomaly scoring function from the view of score distribution, thus better retaining the diversity and more fine-grained information of input data, especially when the unlabeled data contains anomaly noises in more practical AD scenarios. We design a novel loss function called Overlap loss that minimizes the overlap area between the score distributions of normal and abnormal samples, 
which no longer depends on prior anomaly score targets and thus acquires adaptability to various datasets.
Overlap loss consists of \textit{Score Distribution Estimator} and \textit{Overlap Area Calculation}, which are introduced to overcome challenges when estimating arbitrary score distributions, and to ensure the boundness of training loss.
As a general loss component, Overlap loss can be effectively integrated into multiple network architectures for constructing AD models. Extensive experimental results indicate that Overlap loss based AD models significantly outperform their state-of-the-art counterparts, and achieve better performance on different types of anomalies.
\end{abstract}



\maketitle

\section{Introduction} \label{intro}
\begin{figure*}[h!]
    \begin{minipage}{0.12\textwidth}
        \centering
        \includegraphics[width=\textwidth,height=0.6cm]{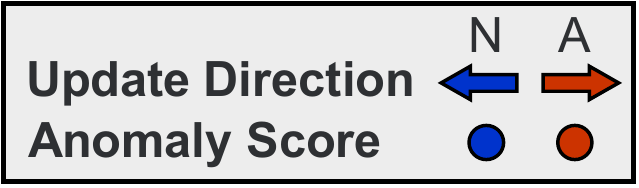}
    \end{minipage}
    \begin{minipage}{0.87\textwidth}
     \centering
     \begin{subfigure}[t]{0.28\textwidth}
         \centering
         \includegraphics[width=\textwidth,valign=t]{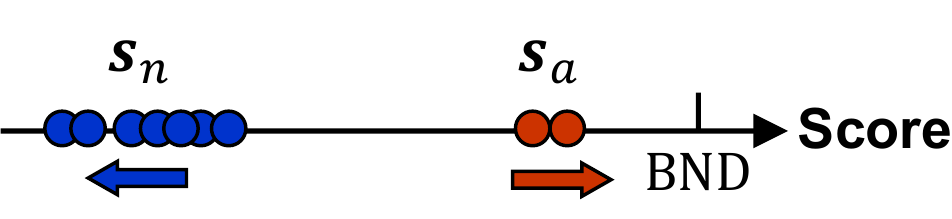}
         \caption{Minus loss}
         \label{fig:minus loss}
     \end{subfigure} 
     \hspace{5mm}
     \begin{subfigure}[t]{0.28\textwidth}
         \centering
         \includegraphics[width=\textwidth,valign=t]{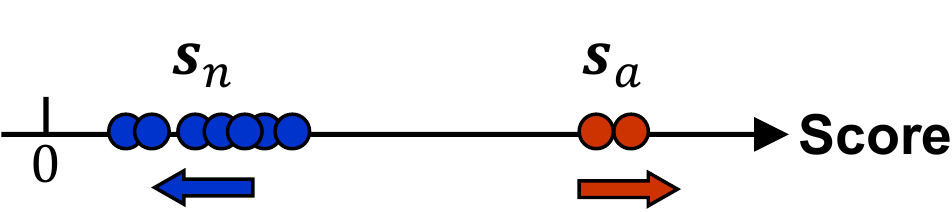}
         \caption{Inverse loss}
         \label{fig:inverse loss}
     \end{subfigure} 
     \hspace{5mm}
     \begin{subfigure}[t]{0.28\textwidth}
         \centering
         \includegraphics[width=\textwidth,valign=t]{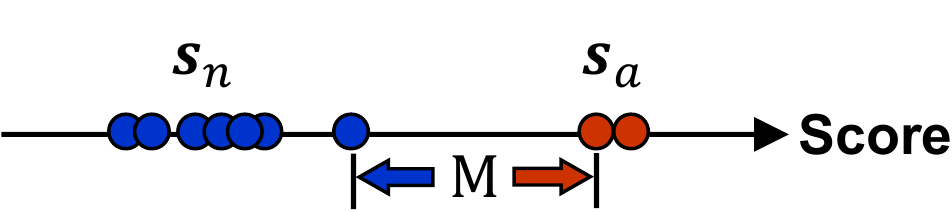}
         \caption{Hinge loss}
         \label{fig:hinge loss}
     \end{subfigure} 
     \begin{subfigure}[b]{0.28\textwidth}
         \centering
         \includegraphics[width=\textwidth]{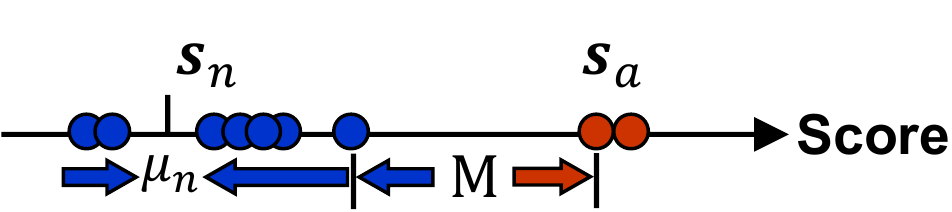}
         \caption{Deviation loss}
         \label{fig:deviation loss}
     \end{subfigure} 
     \hspace{5mm}
     \begin{subfigure}[b]{0.28\textwidth}
         \centering
         \includegraphics[width=\textwidth]{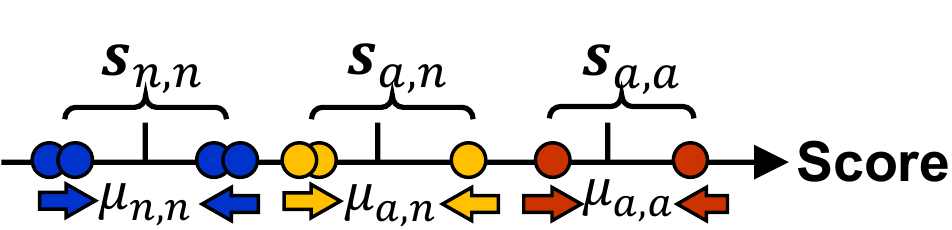}
         \caption{Ordinal loss}
         \label{fig:ordinal loss}
     \end{subfigure} 
     \hspace{5mm}
     \begin{subfigure}[b]{0.28\textwidth}
         \centering
         \includegraphics[width=\textwidth]{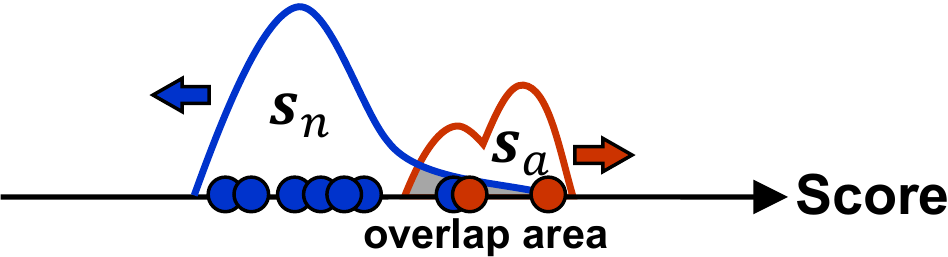}
         \caption{Overlap loss (proposed)}
         \label{fig:Overlap loss}
     \end{subfigure} 
    \end{minipage}
    \vspace{-0.1in}
    \caption{Loss function comparison. Unlike previous AD loss functions that suffer unbounded training loss or rely on specific anomaly score target(s) to guide model training, Overlap loss ensures a bounded training loss without requiring predefined target(s) by first estimating score distributions of normal and abnormal data, and further minimizing their overlap area.}
    \label{fig:loss_comparison}
\end{figure*}

Anomaly detection (AD) is the task of identifying unusual instances that deviate significantly from the majority of data, which has been applied in wide-ranging domains, such as social media analysis \cite{zhao2020multi,malik2022deep}, rare disease detection \cite{zhao2021suod,rekatsinas2015sourceseer}, intrusion detection \cite{lazarevic2003comparative}, and financial fraud detection \cite{pourhabibi2020fraud,hilal2021review}.
Previous research efforts \cite{liu2008isolation, CBLOF, ECOD, SVDD, DAGMM} focus on unsupervised AD which does not require any labeled training data, but unsupervised methods lack any guidance of true anomalies \cite{pang-ultra-high, devnet}. Therefore recent studies propose to learn valuable distinguishing features from a few labeled anomalies that may be identified by domain experts in practice, which is termed as the "anomaly-informed" AD methods \cite{devnet,han2022adbench}.



Current anomaly-informed methods, including semi- and weakly-supervised AD algorithms, mainly devise specific forms of loss functions to leverage such limited label information.  These involve representation learning based minus loss in Unlearning \cite{Unlearning}, inverse loss in DeepSAD \cite{ruff2019deep}, and hinge loss in REPEN \cite{pang-ultra-high}, as we summarized in Figure~\ref{fig:minus loss}\textasciitilde\ref{fig:hinge loss}. In these methods, an anomaly score is generated based on the learned feature transformation of input data, such as the reconstruction error or the embedding distance. However, optimization in the representation space would lead to data-inefficient learning and suboptimal anomaly scoring \cite{devnet, devnet_explainable}. 
Therefore several works \cite{devnet, pang2019deep_weak, FEAWAD} fulfill an end-to-end learning fashion of anomaly score to obtain better performance, designing loss functions to map input instances to their corresponding anomaly score target(s), or to predefine a margin hyperparameter to realize the difference in anomaly scores between unlabeled samples and labeled anomalies, as shown in Figure~\ref{fig:deviation loss}\textasciitilde\ref{fig:ordinal loss}. Nevertheless, such a predefined score target or margin would constrain the model's adaptability to different datasets, and further tuning these hyperparameters for realizing adaptation is often difficult, considering the scarcity of labeled data in practical AD scenarios.


In this paper, we aim to acquire an anomaly scoring function capable of realizing \textit{adaptive anomaly score discrimination} for diverse data scenarios, thus addressing the above issues in previous anomaly-informed loss functions.
Notably, we devise the Overlap loss that minimizes score distribution overlap between normal samples and anomalies, depending on the model itself to decide the suitable distribution of output anomaly scores. This kind of adaptability eliminates the dependency on predefined anomaly score targets. However, a non-trivial challenge is to estimate arbitrary distributions of anomaly scores, which are caused by the scarcity of labeled anomalies and anomaly noises in the unlabeled data. In the Overlap loss, we design a simple and effective method to estimate the overlap area of arbitrary score distributions, while ensuring a correct order in the output anomaly scores and the boundness of training loss to better achieve stability in model training.

The main contributions of this paper can be summarized as follows:
\textbf{(1)} We propose the Overlap loss for the AD community, which can achieve adaptive score distribution discrimination in input data, realizing sufficient global insight of anomaly scores in an end-to-end gradient update fashion.
\textbf{(2)} We verify the effectiveness of the proposed Overlap loss on several network architectures covering both AD and classification tasks. Extensive results on \ndatasets datasets suggest that the proposed Overlap loss could be served as a basis for further development in AD tasks. We open-source the proposed method, related codes, and all testing datasets for AD communities at \url{https://github.com/Minqi824/Overlap}.
\textbf{(3)} We decouple the loss functions from several popular AD algorithms and analyze them in a unified framework, including embedding variation and network parameter changes. Moreover, we investigate the detection performance of different loss functions on various types of anomalies, therefore further exploring the pros and cons of these methods.

\vspace{-0.1in}
\section{Related Work} \label{related}
A desirable anomaly detection approach should produce not only a binary output (normal or abnormal) but also assign a degree of being an anomaly (anomaly score) to each observation \cite{zhang2018adaptive}. Prior literature can be divided into two categories, i.e., AD algorithms without or with supervision. The former assumes that no labeled data is available during the model training stage and is proposed with different assumptions of data distribution \cite{aggarwal2017introduction}, whereas the latter leverages a limited number of labeled samples which may be verified by some domain experts or automatic detecting systems.


\textbf{AD Algorithms without Supervision}. 
Typical anomaly detection methods are constructed for learning anomaly patterns in an unsupervised manner. These include shallow unsupervised models like CBLOF \cite{CBLOF} and ECOD \cite{ECOD}, or ensemble method Isolation Forest \cite{liu2008isolation}. More recently, deep learning (DL) techniques like DeepSVDD \cite{SVDD} and GAN-based MO-GAAL \cite{SOGAAL_MOGAAL} have been proposed for improving the performance of unsupervised AD tasks.

\textbf{AD Algorithms with Supervision}.
Unsupervised methods can not achieve satisfactory performance in practical applications without the guidance of labeled data. Therefore several studies have also investigated utilizing partially labeled data to improve detection performance, which can be summarized into the following three categories:

(i) AD methods that are trained only on labeled normal samples, and detect anomalies that deviate from the normal representation learned in the training process \cite{ALAD, EGBAD, GANomaly, Skip-GANomaly}.

(ii) AD methods that additionally leverage a limited number of labeled anomalies.
A common problem of the above methods using only normal samples is that many of the anomalies they identify are data noises or uninteresting data instances due to the lack of prior knowledge about the abnormal behaviors \cite{pang-ultra-high, devnet}. This results in the development of semi- and weakly-supervised AD methods, which not only learn from numerous unlabeled data but also utilize limited information of labeled anomalies.

Among them, Unlearning \cite{Unlearning} uses the minus loss form to provide an opposite direction of gradient update between the reconstruction error of the normal data and anomalies.
DeepSAD \cite{ruff2019deep} employs the inverse loss to penalize the inverse of the embedding distance such that the representation of anomalies must be mapped further away from the initial center of the hypersphere. 
REPEN \cite{pang-ultra-high} introduces a ranking model-based framework, which applies the hinge loss to encourage a distance separation of low-dimensional representation between normal samples and anomalies.

Several works are proposed to realize end-to-end learning of anomaly score, as they indicate that the above representation learning based AD methods would lead to data-inefficient learning and suboptimal detection performance \cite{devnet_explainable}. Specified with the deviation loss, DevNet \cite{devnet} leverages a prior probability and a margin hyperparameter to enforce significant deviations in anomaly scores between normal and abnormal data. FEAWAD \cite{FEAWAD} incorporates the DAGMM \cite{DAGMM} network architecture with the deviation loss, for the better use of the information among hidden representation, reconstruction residual vector and reconstruction error transformed by the auto-encoder \cite{autoencoders}. 
PReNet \cite{pang2019deep_weak} formulates the scoring function as a pairwise relation learning task, where it defines three constant targets to enforce large margins among the anomaly scores of three types of instance pairs.

(iii) Fully-supervised methods are not specific for AD tasks in general \cite{gornitz2013toward}. Previous studies \cite{anton2018evaluation,omar2013machine} often use existing binary classifiers for this purpose such as Random Forest and MLP. One known risk of supervised methods is that ground truth labels maybe not necessarily accurate enough (i.e., there often exist some unlabeled anomaly noises in normal samples) to capture all types of anomalies, therefore these supervised methods may fail to detect unknown types of anomalies \cite{han2022adbench, ruff2021unifying}.

In summary, some of the above anomaly-informed methods \cite{Unlearning,ruff2019deep,pang-ultra-high} perform an indirect representation learning of anomaly score, while other methods \cite{devnet, FEAWAD, pang2019deep_weak} mainly rely on predetermined training target(s) to realize the score discrepancy between the normal and abnormal data. Our proposed Overlap loss adaptively achieves the score discrimination from a distribution view, thus alleviating the need to define hyperparameter(s) as anomaly score target(s) in model training.

\textbf{Distribution Overlap}.
Our idea is inspired by some recent studies of out-of-distribution (OOD) or multi-classification tasks in the CV field, whereas they usually consider the overlap of class distribution \textit{only as a measurement} to describe the characteristics of datasets or to evaluate model quality \cite{overlap_for_measure_3,overlap_for_measure_1,overlap_for_measure_4,overlap_for_measure_5}.
%
Considering research more closely related to our work, Magnet loss \cite{Magnet} is proposed to achieve local discrimination by penalizing class distribution overlap, as to realize explicit modeling of the distributions of different classes in representation space. Based on the entropy minimization principle \cite{grandvalet2004semi}, MA-DNN \cite{chen2018semi} minimizes the model entropy in the feature space and penalizes inconsistent network predictions at the class level. 
Nevertheless, these two methods are mainly devised for distance metric learning (DML) that presents optimization in the representation space. We propose to
directly optimize distribution overlap in the anomaly scoring space to realize adaptive score distribution discrimination in input instances, which is tailored for the AD problem where there exist data noises and only a limited number of labeled anomalies.

\vspace{-0.1in}
\section{Methodology} \label{setting}
\subsection{Problem Statement}
Assume the training dataset
$\mathcal{D}=\left\{\boldsymbol{x}_{1}^{n}, \ldots, \boldsymbol{x}_{k}^{n},\left(\boldsymbol{x}_{k+1}^{a}, y_{k+1}^{a}\right), \ldots\right.$,
$\left.\left(\boldsymbol{x}_{k+m}^{a}, y_{k+m}^{a}\right)\right\}$
collects both unlabeled instances $\mathcal{D}_{n}=\left\{\boldsymbol{x}_{i}^{n}\right\}_{i=1}^{k}$ and a handful of labeled anomalies $\mathcal{D}_{a}=\left\{\left(\boldsymbol{x}_{j}^{a}, y_{j}^{a}\right)\right\}_{j=1}^{m}$, where $\boldsymbol{x} \in \mathbb{R}^{d}$ represents the input feature and $y_{j}^{a}$ is the label of identified anomalies. Usually, we have $m \ll k$, since only limited prior knowledge of anomalies is available. Such data assumption is more practical for AD problems, and has been studied in recent works \cite{devnet, pang2019deep_weak, FEAWAD, score_guided_network}. 
Given such a dataset, our goal is to train a model, to effectively assign higher anomaly score for the abnormal data.

\subsection{Overview of the Proposed Overlap Loss}
Overlap loss first employs a \textit{Score Distribution Estimator} for estimating the unknown probability density function (PDF) of the output anomaly scores in neural networks and then conducts \textit{Overlap Area Calculation} between the anomaly score distributions of the unlabeled samples and labeled anomalies. Finally, Overlap loss minimizes the calculated overlap area of score distributions to provide the gradient for backpropagation in neural networks.
The proposed Overlap loss fulfills the following properties: 
(i) the boundness of training loss for better convergence in the model training.
(ii) eliminating explicit training target of anomaly score (e.g., constant or margin hyperparameter(s)) to enhance the model adaptability to different datasets.
(iii) optimizing the entire anomaly score distribution, instead of pointwise optimization between the estimated anomaly scores and their corresponding targets.
\vspace{-0.1in}

\subsection{Overlap Loss for Score Distribution Discrimination} \label{method}
In the following subsections, we illustrate two main parts of proposed Overlap loss: \textit{Score Distribution Estimator} and \textit{Overlap Area Calculation}, along with their corresponding basic ideas and challenges, as complements to our final solutions.

\subsubsection{Score Distribution Estimator} \label{subsec: score distribution estimator}
Instead of pointwise optimization of the output anomaly scores, here we consider optimizing the anomaly score from a distribution view.
Let $Q\in\mathbb{R}^{M}$ be the hidden representation space, an end-to-end anomaly scoring network $\phi(\cdot ; \Theta): \boldsymbol{x} \mapsto \mathbb{R}$ can be defined as a combination of a feature representation learner $\psi\left(\cdot ; \Theta_{t}\right): \boldsymbol{x} \mapsto Q$ and an anomaly scoring function $\eta\left(\cdot ; \Theta_{s}\right): Q \mapsto \mathbb{R}$, in which $\Theta=\left\{\Theta_{t}, \Theta_{s}\right\}$.
If we denote the anomaly score of normal data as $\phi(\boldsymbol{x}^{n} ; \Theta)=s_{n}$ and that of abnormal data as $\phi(\boldsymbol{x}^{a} ; \Theta)=s_{a}$, a density estimator $f(\cdot)$ is then applied to estimate the PDFs of both $\boldsymbol{s}_{n}$ and $\boldsymbol{s}_{a}$ in a training batch.

\begin{figure*}[h!]
     \centering
     \begin{subfigure}[b]{0.27\textwidth}
         \centering
         \includegraphics[width=\textwidth]{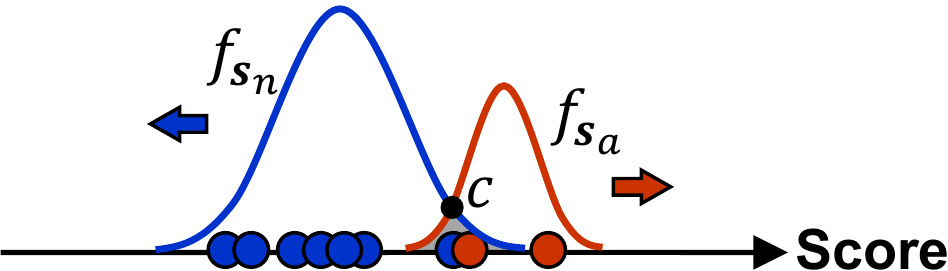}
         \caption{Gaussian score distribution overlap}
         \label{fig:challenges_1}
     \end{subfigure}
     \hspace{5mm}
     \begin{subfigure}[b]{0.27\textwidth}
         \centering
         \includegraphics[width=\textwidth]{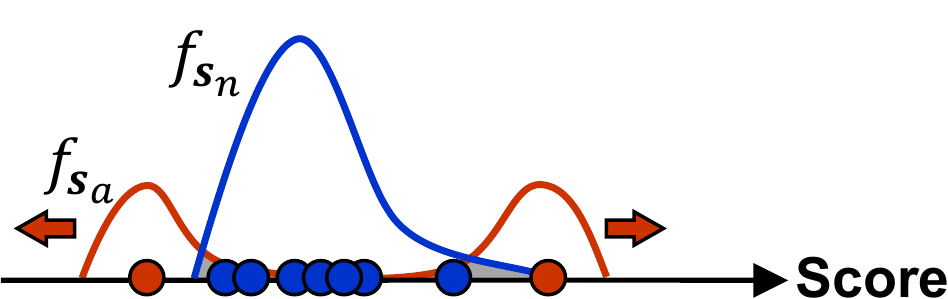}
         \caption{Arbitrary score distribution overlap}
         \label{fig:challenges_2}
     \end{subfigure} 
     \hspace{5mm}
     \begin{subfigure}[b]{0.27\textwidth}
         \centering
         \includegraphics[width=\textwidth]{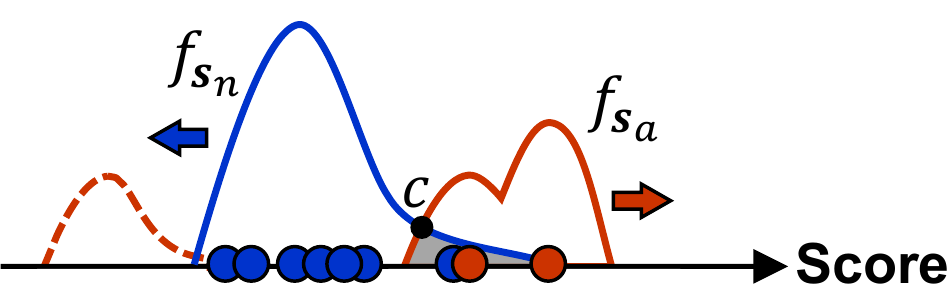}
         \caption{Proposed score distribution overlap}
         \label{fig:challenges_proposed}
     \end{subfigure} 
     \vspace{-0.1in}
    \caption{Anomaly score distribution overlaps. (a) The prior assumption of Gaussian distributions limits the representational ability of neural networks. (b) Overlap of arbitrary score distributions leads to the disorder in anomaly scores. (c) The proposed Overlap loss minimizes the overlap area of arbitrary score distributions while ensuring correct order in anomaly scores.}
    \label{fig:Challenge}
\end{figure*}

A straightforward idea is to employ a prior distribution, e.g., the Gaussian distribution, as the score distribution estimator. Gaussian distribution inherits several good properties. For instance, the intersection point $c$ used for estimating the score distribution overlap in Figure~\ref{fig:challenges_1} can be calculated by the following formula \cite{inman1989overlapping}:
\vspace{-0.2in}

\begin{equation} \label{Eq:intersection_point_Gaussian}
c=\frac{\mu_{a} \sigma_{n}^{2}-\sigma_{a}\left(\mu_{n} \sigma_{a}+\sigma_{n} \sqrt{\left(\mu_{n}-\mu_{a}\right)^{2}+2\left(\sigma_{n}^{2}-\sigma_{a}^{2}\right) \log \left(\frac{\sigma_{n}}{\sigma_{a}}\right)}\right)}{\sigma_{n}^{2}-\sigma_{a}^{2}}
\end{equation}

where $\mu$ and $\sigma$ are their corresponding mean and variance of score distributions, respectively. The main challenge of this basic idea is that the number of labeled anomalies is usually too small to satisfy the Gaussian distribution assumption according to the central limit theorem \cite{central_limit_theory}, while enforcing the anomaly scores to follow this Gaussian prior would limit the representational ability of neural networks and further distort the anomaly scoring space, resulting in suboptimal performance.

To address the above challenges, we employ a score distribution estimator that is capable of estimating \textit{arbitrary} distribution of output anomaly scores. In this paper, we use the non-parametric Kernel Density Estimation (KDE) method for estimating the arbitrary anomaly score distribution that may be caused by the scarcity of labeled data or the anomaly contamination in the unlabeled data. Actually, other differentiable density estimators can also be applied into our proposed Overlap loss.

If we denote the output anomaly score as $\phi(\boldsymbol{x} ; \Theta)=s$, the empirical cumulative distribution function (ECDF) can be defined as $\hat{F}_{N}(s)=\frac{1}{N} \sum_{i=1}^{N} \mathbf{1}_{s_{i} \leq s}$, where $\mathbf{1}$ is the indicator function and $N$ is the number of partitions. $\hat{F}_{N}(s)$ is an unbiased estimator \cite{dekking2005modern} of the cumulative distribution function (CDF) $F(s)$, and can be further used for estimating the PDF by the following equation:
\vspace{-0.2in}

\begin{equation}
\begin{aligned}
\hat{f}(s)&=\lim _{h \rightarrow 0} \frac{\hat{F}_{N}(s+h)-\hat{F}_{N}(s-h)}{2 h} \approx \frac{1}{2 N h} \sum_{i=1}^{N}\left(\mathbf{1}_{s_{i} \leq s+h}-\mathbf{1}_{s_{i} \leq s-h}\right) \\
&=\frac{1}{2 N h} \sum_{i=1}^{N}\left(\mathbf{1}_{s-h \leq s_{i} \leq s+h}\right)=\frac{1}{N h} \sum_{i=1}^{N} \frac{1}{2} \mathbf{1}\left(\frac{\left|s-s_{i}\right|}{h} \leq 1\right)
\end{aligned}
\end{equation}

where $h$ is the bandwidth.
If we denote the kernel function as $K(s)=\frac{1}{2} \mathbf{1}(s \leq 1)$, the estimated PDF can be rewritten as:
\begin{equation}
\hat{f}(s)=\frac{1}{N h} \sum_{i=1}^{N} K\left(\frac{\left|s-s_{i}\right|}{h}\right)=\frac{1}{N h} \sum_{i=1}^{N} K\left(\frac{s-s_{i}}{h}\right)
\end{equation}

where $K(\cdot)$ is symmetric. We use Gaussian kernel in KDE, i.e., $K(s ; h) \propto \exp \left(-\frac{s^{2}}{2 h^{2}}\right)$, for estimating the unknown PDFs of anomaly scores, where the PDFs are further utilized to calculate the overlap area of score distributions, as described in the following subsection.

\subsubsection{Overlap Area Calculation} \label{subsec: overlap area calculation}
Once we obtain the estimated score distributions (i.e., PDFs), the score distribution overlap can be calculated as the overlap area of PDFs between normal samples and the abnormal ones.



An optional method is to directly use the integral to approximate the score distribution overlap, as illustrated in Eq. \ref{Eq:overlap_wo_order}. The overlap area between the PDFs of $\boldsymbol{s}_{n}$ and $\boldsymbol{s}_{a}$ is formulated as the integral of the one with the smaller probability density:
\vspace{-0.05in}

\begin{equation} \label{Eq:overlap_wo_order}
O\left(\boldsymbol{s}_{n}, \boldsymbol{s}_{a}\right)=\int_{\min \left(\boldsymbol{s}_{n}, \boldsymbol{s}_{a}\right)}^{\max \left(\boldsymbol{s}_{n}, \boldsymbol{s}_{a}\right)} \min \left(\hat{f}_{\boldsymbol{s}_{n}}(t), \hat{f}_{\boldsymbol{s}_{a}}(t)\right) d t
\end{equation}


The main challenge of the above basic idea is that such a method \textit{does not} necessarily guarantee a correct gradient update direction for anomaly scores, as illustrated in Figure~\ref{fig:challenges_2}. The neural networks could minimize the overlap area of the score distributions, while mistakenly assigning lower anomaly scores for the anomalies (e.g., the left side in the score distribution of the anomalies in Figure~\ref{fig:challenges_2}) instead of the normal ones. 
This problem can be remedied through the multi-task learning form by combining Eq.\ref{Eq:overlap_wo_order} with a ranking loss term \cite{wang2019ranked}, as shown in Eq.\ref{Eq:overlap_w_order}. However, although it ensures the order in anomaly scores, i.e., the anomaly scores of abnormal data should be further ranked higher than that of normal data, such a method may suffer from the difficult optimization problem in multi-task learning, sometimes leading to worse performance and data inefficiency compared to learning tasks individually \cite{yu2020gradient,parisotto2016actor}.
\vspace{-0.1in}


\begin{equation} \label{Eq:overlap_w_order}
O\left(\boldsymbol{s}_{n}, \boldsymbol{s}_{a}\right)=\int_{\min \left(\boldsymbol{s}_{n}, \boldsymbol{s}_{a}\right)}^{\max \left(\boldsymbol{s}_{n}, \boldsymbol{s}_{a}\right)} \min \left(\hat{f}_{\boldsymbol{s}_{n}}(t), \hat{f}_{\boldsymbol{s}_{a}}(t)\right) d t + \max \left(0, \boldsymbol{s}_{n}-\boldsymbol{s}_{a}\right)
\end{equation}

Our proposed Overlap loss aims to calculate the overlap area of \textit{arbitrary} score distributions while ensuring the correct \textit{order} in anomaly scores. We manage to acquire the intersection point $c$ of these arbitrary score distributions (see Figure~\ref{fig:challenges_proposed}), and the score distribution overlap between $\boldsymbol{s}_{n}$ and $\boldsymbol{s}_{a}$ in a training batch can be further formulated as Eq.\ref{Eq:overlap}, where $\hat{F}_{\boldsymbol{s}_{n}}(\cdot)$ and $\hat{F}_{\boldsymbol{s}_{a}}(\cdot)$ are the estimated CDF of normal and abnormal data, respectively.
\vspace{-0.18in}

\begin{equation} \label{Eq:overlap}
O\left(\boldsymbol{s}_{n}, \boldsymbol{s}_{a}\right)=P\left(\boldsymbol{s}_{n}>c\right)+P\left(\boldsymbol{s}_{a}<c\right)=1-\hat{F}_{\boldsymbol{s}_{n}}(c)+\hat{F}_{\boldsymbol{s}_{a}}(c)
\end{equation}

As shown in Figure~\ref{fig:challenges_proposed}, the Overlap loss formulated in Eq.\ref{Eq:overlap} guarantees the order in output anomaly scores. A small overlap area with correct score order means a close to zero loss of $O\left(\boldsymbol{s}_{n}, \boldsymbol{s}_{a}\right)$. If the anomaly scores of abnormal data are smaller than that of normal data, $O\left(\boldsymbol{s}_{n}, \boldsymbol{s}_{a}\right)$ would penalize this disorder and be close to 2, since both $P\left(\boldsymbol{s}_{n}>c\right)$ and $P\left(\boldsymbol{s}_{a}<c\right)$ are close to 1, respectively. Moreover, $O\left(\boldsymbol{s}_{n}, \boldsymbol{s}_{a}\right)$ is naturally bounded to $[0, 2]$ due to the property of PDF.

\begin{figure}[h!]
     \centering
     \begin{subfigure}[b]{0.14\textwidth}
         \centering
         \includegraphics[width=\textwidth]{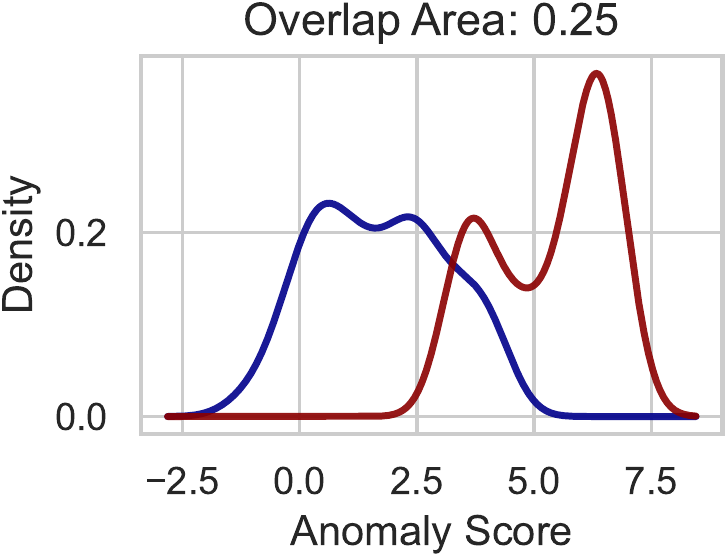}
         \caption{PDFs w.r.t. the number of $c=1$}
         \label{fig:IP_1}
     \end{subfigure} \hfill
     \begin{subfigure}[b]{0.14\textwidth}
         \centering
         \includegraphics[width=\textwidth]{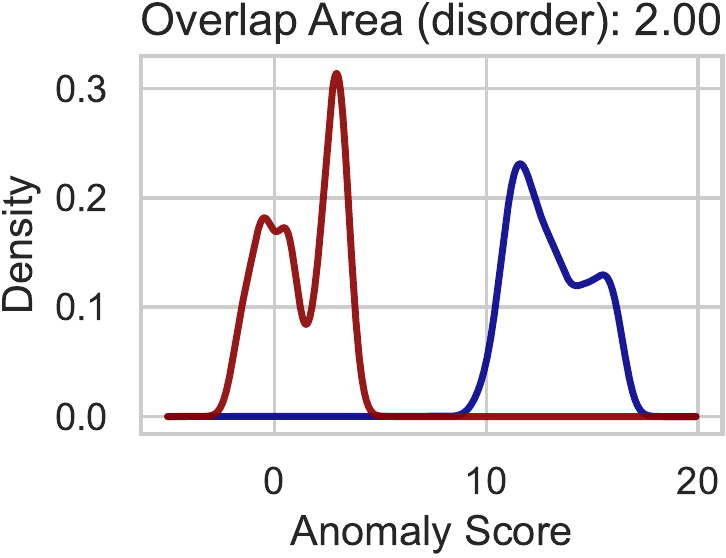}
         \caption{PDFs w.r.t. the number of $c=0$}
         \label{fig:IP_0}
     \end{subfigure} \hfill
     \begin{subfigure}[b]{0.14\textwidth}
         \centering
         \includegraphics[width=\textwidth]{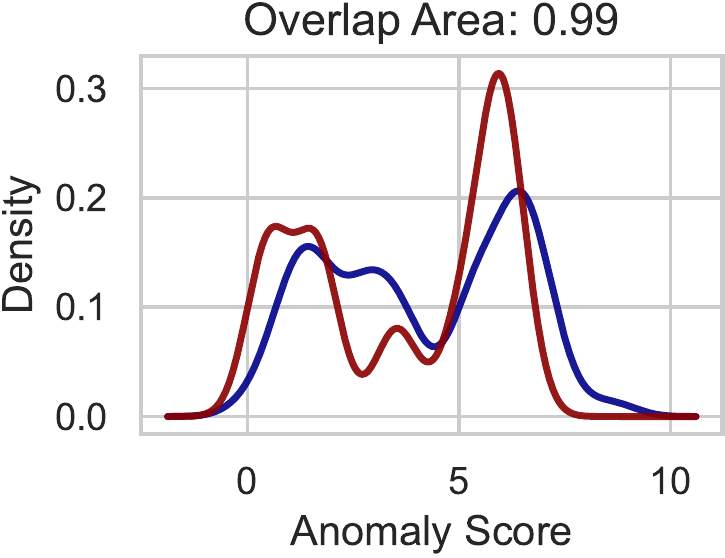}
         \caption{PDFs w.r.t. the number of $c>1$}
         \label{fig:IP_2}
     \end{subfigure} \hfill
     \begin{subfigure}[b]{0.14\textwidth}
         \centering
         \includegraphics[width=\textwidth]{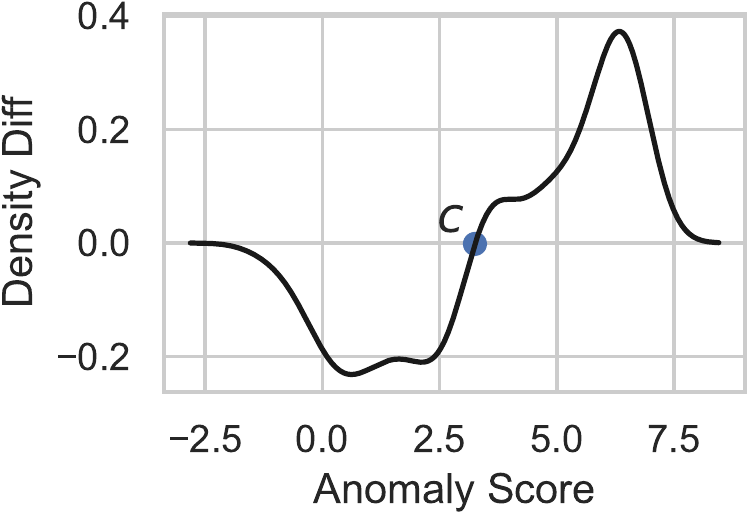}
         \caption{PDFs diff w.r.t. the number of $c=1$}
         \label{fig:IP_diff_1}
     \end{subfigure} \hfill
     \begin{subfigure}[b]{0.14\textwidth}
         \centering
         \includegraphics[width=\textwidth]{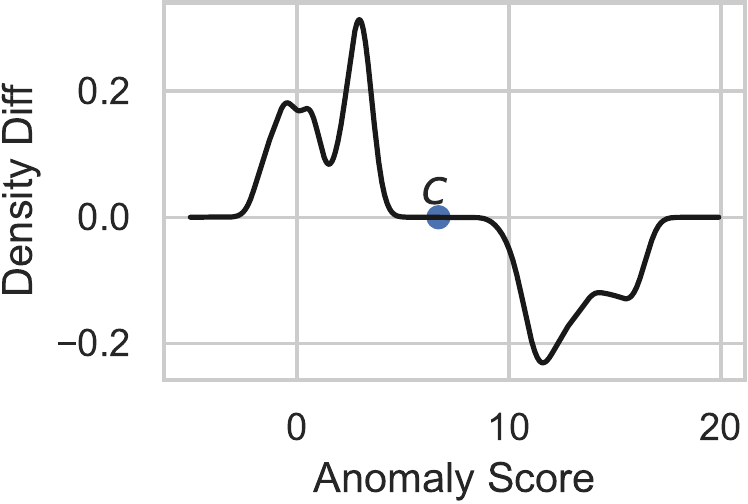}
         \caption{PDFs diff w.r.t. the number of $c=0$}
         \label{fig:IP_diff_0}
     \end{subfigure} \hfill
     \begin{subfigure}[b]{0.14\textwidth}
         \centering
         \includegraphics[width=\textwidth]{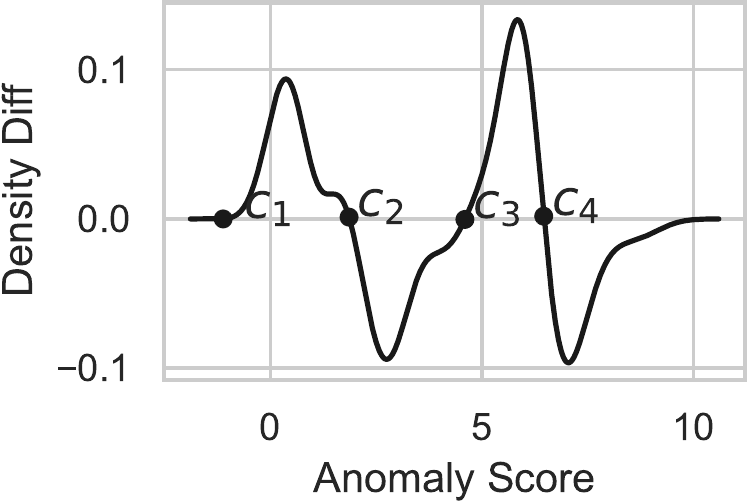}
         \caption{PDFs diff w.r.t. the number of $c>1$}
         \label{fig:IP_diff_2}
     \end{subfigure} \hfill
    \vspace{-0.1in}
    \caption{Calculation of intersection point(s) for arbitrary anomaly score distributions of $f_{\boldsymbol{s}_n}(\cdot)$ (blue) and $f_{\boldsymbol{s}_a}(\cdot)$ (red). (a)\textasciitilde(c) correspond to the situations of one, zero, and multiple intersection point(s), respectively. (d)\textasciitilde(f) are their corresponding PDF differences.
}
    \label{fig:IP}
\end{figure}

However, for two arbitrary score distributions, we can not directly calculate the intersection point $c$ by the formula suitable for Gaussian distribution in Eq.\ref{Eq:intersection_point_Gaussian}. Instead, we acquire the intersection point $c$ as the corresponding x value of the non-zero element of $d_{k}^{s}$ in Eq.\ref{Eq:intersection} for the arbitrary score distribution scenario, where $s_{k}$ is generated by the arithmetic sequence $s_k=\min \left(\boldsymbol{s}_n, \boldsymbol{s}_a\right)+(k-1) \frac{\max \left(\boldsymbol{s}_n, \boldsymbol{s}_a\right)-\min \left(\boldsymbol{s}_n, \boldsymbol{s}_a\right)}{N}$ and $k=1, \ldots, N$.
%
%
In other words, we compare the PDF differences between two adjacent points of the score distributions $\boldsymbol{s}_{a}$ and $\boldsymbol{s}_{n}$, as shown in Figure~\ref{fig:IP}.
\vspace{-0.2in}

\begin{equation} \label{Eq:intersection}
\begin{aligned}
d_{k}^{s}=\operatorname{sgn}\left(\hat{f}_{\boldsymbol{s}_{a}}\left(s_{k+1}\right)-\hat{f}_{\boldsymbol{s}_{n}}\left(s_{k+1}\right)\right)-\operatorname{sgn}\left(\hat{f}_{\boldsymbol{s}_{a}}\left(s_{k}\right)-\hat{f}_{\boldsymbol{s}_{n}}\left(s_{k}\right)\right)
\end{aligned}
\end{equation}

Figure~\ref{fig:IP} shows toy examples of calculating the intersection point(s) $c$. For most cases where there is only one intersection point between $\hat{f}_{\boldsymbol{s}_n}(\cdot)$ and $\hat{f}_{\boldsymbol{s}_a}(\cdot)$, $c$ is regarded as the x value of the sign change point of PDF differences, as shown in Figure~\ref{fig:IP_1} and \ref{fig:IP_diff_1}. Even if the two score distributions are far apart (see Figure~\ref{fig:IP_0}), we could still extend the x range of their PDFs and acquire $c$, as shown in Figure~\ref{fig:IP_diff_0}. It is worth noting that for the case in Figure~\ref{fig:IP_0}, Overlap loss would reach its upper bound with an overlap area of 2 for penalizing the disorder in estimated anomaly scores, as illustrated in Eq.\ref{Eq:overlap}. For the case where there exist multiple intersection points (shown in Figure\ref{fig:IP_2} and \ref{fig:IP_diff_2}), we randomly choose one of them as $c$. We show in the Appendix that the detection performance of this strategy is very close to that of ensembling different intersection points while improving the efficiency of model training.

After that, the integral of CDF $F_{\boldsymbol{s}}(c)$ can be approximated via the trapezoidal rule \cite{trapezoidal} in Eq.\ref{Eq:CDF}, where $\Delta s_{k}$ is adjusted based on the intersection point as $\Delta s_{k}=\left[c-\min \left(\boldsymbol{s}_{n}, \boldsymbol{s}_{a}\right)\right] / N$.
\vspace{-0.2in}

\begin{equation} \label{Eq:CDF}
\begin{aligned}
\hat{F}_{\boldsymbol{s}}(c)=\int_{-\infty}^{c} \hat{f}_{\boldsymbol{s}}(t) d t&=\int_{\min \left(\boldsymbol{s}_{n}, \boldsymbol{s}_{a}\right)}^{c} \hat{f}_{\boldsymbol{s}}(t) d t
\approx \sum_{k=1}^{N} \frac{\hat{f}_{\boldsymbol{s}}\left(s_{k}\right)+\hat{f}_{\boldsymbol{s}}\left(s_{k+1}\right)}{2} \Delta s_{k}
\end{aligned}
\end{equation}

Based on the above notations, Overlap loss is defined as:
\vspace{-0.2in}

\begin{equation} \label{Eq:objective}
\begin{aligned}
&\mathcal{L}_{\text {Overlap }}(\boldsymbol{x} \mid \Theta)=\;O\left(\boldsymbol{s}_{n}, \boldsymbol{s}_{a}\right)=\;\\&1-\sum_{k=1}^{N} \frac{\hat{f}_{\boldsymbol{s}_{n}}\left(s_{n, k}\right)+\hat{f}_{\boldsymbol{s}_{n}}\left(s_{n, k+1}\right)}{2} \Delta s_{n, k}+
\sum_{k=1}^{N} \frac{\hat{f}_{\boldsymbol{s}_{a}}\left(s_{a, k}\right)+\hat{f}_{\boldsymbol{s}_{a}}\left(s_{a, k+1}\right)}{2} \Delta s_{a, k}
\end{aligned}
\end{equation}

\subsection{Network Architecture}
\begin{figure*}[t!]
    \centering
    \includegraphics[width=16cm]{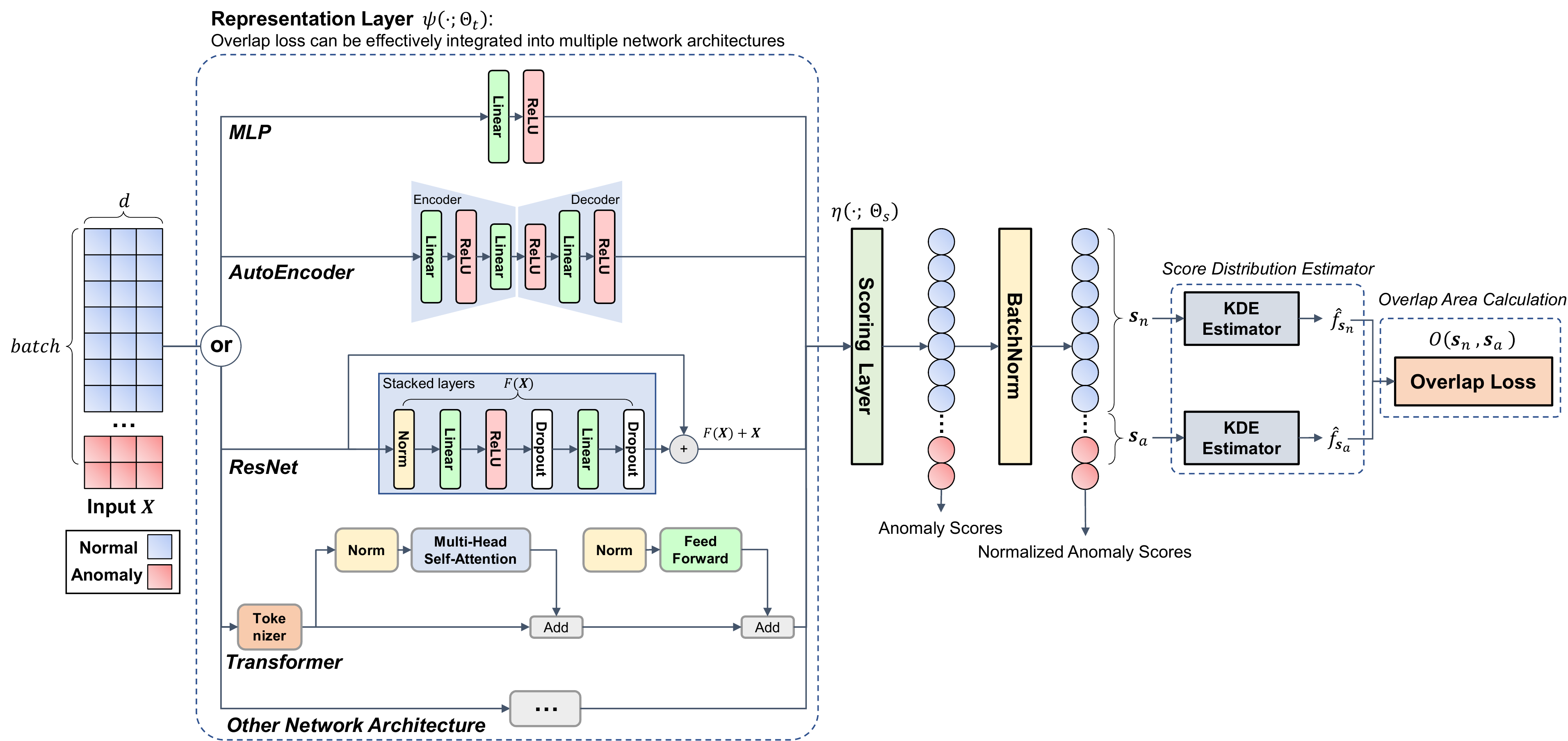}
    \caption{AD model instantiated by the proposed Overlap loss, which consists of a representation layer $\psi\left(\cdot ; \Theta_{t}\right)$ and a scoring layer $\eta\left(\cdot ; \Theta_{s}\right)$ with batch normalization. The output anomaly scores are used for estimating the score distributions (PDFs) of normal samples $\hat{f}_{\boldsymbol{s}_{n}}(\cdot)$ and that of anomalies $\hat{f}_{\boldsymbol{s}_{a}}(\cdot)$ via the KDE estimators. Finally, the calculated overlap area of anomaly score distributions is minimized.}
    \label{fig:network}
\end{figure*}

Overlap loss is instantiated into an end-to-end neural network that consists of a feature representation layer $\psi\left(\cdot ; \Theta_{t}\right)$ and a scoring layer $\eta\left(\cdot ; \Theta_{s}\right)$. The BatchNorm layer is applied after the scoring layer to normalize the output anomaly scores. After that, score distributions of both normal data and anomalies are estimated by the KDE estimators, where their score distribution overlap is further calculated via the proposed Overlap loss, as shown in Figure~\ref{fig:network}.

We point out that the proposed Overlap loss can be effectively integrated into multiple popular network architectures, including the widely-used MLP and AutoEncoder in AD tasks, and some cutting-edge architectures like ResNet and Transformer in the classification tasks. Algorithm \ref{ADSD algorithm} provides detailed steps of instantiated models based on our proposed Overlap loss.
\begin{algorithm}[!h]
\footnotesize
\SetAlgoLined
  \textbf{Input}: Unlabeled instances $\mathcal{D}_{n}$, a limited number of identified anomalies $\mathcal{D}_{a}$\\
 \textbf{Output}: Anomaly scores $\boldsymbol{s}$\\
 Initialize network parameters of both feature representation layer $\Theta_{t}$ and scoring layer $\Theta_{s}$\\
 \For{epoch=1:$n_{epoch}$}{
 \For{batch=1:$n_{batch}$}{
 Randomly sample unlabeled instances $\boldsymbol{x}_{batch}^{n}$ from $\mathcal{D}_{n}$ and labeled anomalies $\boldsymbol{x}_{batch}^{a}$ from $\mathcal{D}_{a}$\\
 (1) Acquire the anomaly scores with $\phi(\boldsymbol{x}_{batch}^{n}; \Theta)=\boldsymbol{s}_{batch}^{n}$ and $\phi(\boldsymbol{x}_{batch}^{a}; \Theta)=\boldsymbol{s}_{batch}^{a}$\\
 (2) Use the KDE method to estimate the PDF $\hat{f}_{\boldsymbol{s}_{batch}^{n}}(\cdot)$ and $\hat{f}_{\boldsymbol{s}_{batch}^{a}}(\cdot)$ of the output anomaly scores\\
 (3) Calculate the intersection point $c$ by Eq.~\ref{Eq:intersection}\\
 (4) Approximate the CDFs by the trapezoidal rule in Eq.~\ref{Eq:CDF}\\
 (5) Calculate and minimize the score distribution overlap $O\left(\boldsymbol{s}_{batch}^{n}, {s}_{batch}^{a}\right)$ via Eq.~\ref{Eq:objective}\\
 (6) Perform backpropagation and update network parameters $\Theta=\left\{\Theta_{t}, \Theta_{s}\right\}$
 }
 }
  Output anomaly scores via learned scoring function $\phi(\boldsymbol{x} ; \Theta)=\boldsymbol{s}$
 \caption{AD model instantiated by the Overlap loss}
 \label{ADSD algorithm}
\end{algorithm}

\section{Experiments} \label{experiment}
\subsection{Experiment Setting}
\textbf{Datasets}. We apply \ndatasets publicly available real-world datasets for model evaluation. These datasets include several domains such as disease diagnosis, speech recognition, and image identification. Detailed dataset description is illustrated in Appendix~\ref{subsec: appendix_dataset_description}. For each dataset, 70\% data is split as the training set and the remaining 30\% as the testing set, where the same proportion of anomalies is kept by the stratified sampling. We discuss the model performance in Section \ref{model_performance} w.r.t. different ratios of labeled anomalies to all true anomalies $\gamma_{l}=m /(k+m)$ in the training set, where $m$ labeled anomalies are sampled from the entire anomaly data and the rest of $k$ instances remain unlabeled.

\textbf{Baselines}. We compare the proposed method with the following baselines\footnote{Unlearning \cite{Unlearning} is not included here since it is originally proposed for the time-series task, while we explore the Minus loss of Unlearning in Section \ref{additional results}. 
We do not include the results of DAGMM \cite{DAGMM} for comparison as it may not converge on some datasets. We exclude the semi-supervised Dual-MGAN \cite{li2022dual} method for comparison since it is too computationally expensive.}, and summarize them according to their network architectures and levels of supervision, as is shown in Table~\ref{tab:model_comparison_AUCPR}.

\begin{itemize}[leftmargin=*]
    \item \textbf{Iforest} \cite{liu2008isolation}. An ensemble of binary trees defines the anomaly score as the closeness of an individual instance to the root.
    \item \textbf{ECOD} \cite{ECOD}. A parameter-free method that estimates the empirical cumulative distribution of input features and regards tail probabilities as the anomaly score.
    \item \textbf{DeepSVDD} \cite{SVDD}. A neural network based model that describes the anomaly score as the distance of transformed embedding to the center of the hypersphere.
    \item \textbf{GANomaly} \cite{GANomaly}. A GAN-based method that defines the reconstruction error of the input instance as the anomaly score.
    \item \textbf{DeepSAD} \cite{ruff2019deep}. A deep semi-supervised one-class method that improves the unsupervised DeepSVDD.
    \item \textbf{REPEN} \cite{pang-ultra-high}. A neural network based model that leverages transformed low-dimensional representation for random distance-based detectors.
    \item \textbf{DevNet} \cite{devnet}. A neural network based model that uses a prior probability to enforce the statistical deviation score of input instances.
    \item \textbf{PReNet} \cite{pang2019deep_weak}. A neural network based model that defines a two-stream ordinal regression to learn the relation of instance pairs.
    \item \textbf{FEAWAD} \cite{FEAWAD}. A neural network based model that incorporates the network architecture of DAGMM \cite{DAGMM} with the deviation loss of DevNet. We compare our proposed method with both its weakly- and fully-supervised versions.
    \item \textbf{ResNet} \cite{gorishniy2021revisiting}. ResNet-like architecture turns out to be a strong baseline that is often missing in prior tabular AD tasks.
    \item \textbf{FTTransformer} \cite{gorishniy2021revisiting}. A 
    Transformer architecture implements with Feature Tokenizer. FTTransformer has been proven to be better than other DL solutions on tabular tasks.
\end{itemize}

\textbf{Metrics}. We evaluate the above models by two metrics:  the AUC-ROC (Area Under Receiver Operating Characteristic Curve) and the AUC-PR (Area Under Precision-Recall Curve) values. We mainly report the AUC-PR results due to the space limit, and demonstrate the AUC-ROC results in the Appendix. We find that the results of these two metrics are generally consistent.
Besides, we apply the pairwise Wilcoxon signed rank test \cite{wilcoxon} to examine the significance of proposed methods against its competitors.

\textbf{Training details}
For the proposed Overlap loss based AD models, we use the SGD optimizer with 0.001 learning rate and 0.7 momentum. The weight decay is set to 0.01. The bandwidth $h$ in the KDE method is set to 1. The $N$ in Eq.\ref{Eq:intersection}\textasciitilde\ref{Eq:objective} is set to 1000 by default. We train the Overlap loss based MLP and AutoEncoder models (namely MLP-Overlap and AE-Overlap) for 20 epochs, where batch size of 256 is used. For ResNet and FTTransformer architectures, we train the Overlap loss based models (namely ResNet-Overlap and FTTransformer-Overlap) 100 training epochs just as their original paper. We provide the training details of compared baselines in Appendix~\ref{subsec: appendix_training_details}, which are mainly according to their original papers. All the experiments are run on a Tesla V100 GPU accelerator.

\begin{table*}[h!]
\footnotesize
\caption{Average AUC-PR performance over \ndatasets real-world datasets. Each experiment is repeated 5 times. $\gamma_{l}$ stands for the ratio of labeled anomalies to all true anomalies in the training set. $\Delta$ Perf. shows the relative improvement of Overlap loss based models over their corresponding counterparts. $^{***}$, $^{**}$ and $^{*}$ denote statistical significance at $1\%$, $5\%$ and $10\%$ of Wilcoxon signed rank test, respectively. The best results are in \textbf{bold}.}
\label{tab:model_comparison_AUCPR}
  \centering
    \begin{tabular}{ccc|cc|cc|cc}
    \toprule
    \multirow{2}{*}{\textbf{Architecture}} & \multirow{2}{*}{\textbf{Model}} & \multirow{2}{*}{\textbf{Supervision}} & \multicolumn{2}{c}{$\gamma_{l}=5\%$} & \multicolumn{2}{c}{$\gamma_{l}=10\%$} & \multicolumn{2}{c}{$\gamma_{l}=20\%$} \\
\cmidrule{4-9}          &       &       & AUC-PR & $\Delta$ Perf. & AUC-PR & $\Delta$ Perf. & AUC-PR & $\Delta$ Perf. \\
    \midrule
    \multirow{6}[4]{*}{\textbf{Typical}} & Iforest &Unsup       & 0.389±0.295 & /     & 0.389±0.295 & /     & 0.389±0.295 & / \\
          & ECOD  &Unsup       & 0.315±0.239 & /     & 0.315±0.239 & /     & 0.315±0.239 & / \\
          & DeepSVDD &Unsup       & 0.147±0.120 & /     & 0.147±0.120 & /     & 0.147±0.120 & / \\
\cmidrule{2-9}          & GANomaly &Semi       & 0.297±0.191 & /     & 0.296±0.195 & /     & 0.306±0.201 & / \\
          & DeepSAD &Semi       & 0.506±0.253 & /     & 0.601±0.275 & /     & 0.675±0.284 & / \\
          & REPEN & Weak      & 0.560±0.300 & /     & 0.603±0.308 & /     & 0.639±0.306 & / \\
    \midrule
    \multirow{3}[1]{*}{\textbf{MLP}} & DevNet & Weak      & 0.606±0.311 & $+2.89\%$ & 0.626±0.307 & $+7.75\%^{**}$ & 0.652±0.305 & $+6.74\%^{*}$ \\
          & PReNet & Weak      & 0.612±0.305 & $+1.82\%$ & 0.638±0.307 & $+5.67\%^{*}$ & 0.660±0.303 & $+5.49\%$ \\
          & MLP-Overlap (ours) &Weak       & \textbf{0.623±0.291} & /     & \textbf{0.674±0.286} & /     & \textbf{0.696±0.288} & / \\
    \midrule
    \multirow{3}[0]{*}{\textbf{AutoEncoder}} & FEAWAD & Sup      & 0.509±0.269 & $+28.04\%^{***}$ & 0.620±0.270 & $+12.05\%^{***}$ & 0.678±0.270 & $+5.17\%^{**}$ \\
    & FEAWAD & Weak      & 0.596±0.286 & $+9.29\%^{***}$ & 0.645±0.293 & $+7.71\%^{***}$ & 0.682±0.283 & $+4.56\%^{**}$ \\
          & AE-Overlap (ours) & Weak      & \textbf{0.652±0.290} & /     & \textbf{0.695±0.294} & /     & \textbf{0.713±0.296} & / \\
    \midrule
    \multirow{2}[0]{*}{\textbf{ResNet}} & ResNet & Sup      & 0.401±0.241 & $+56.30\%^{***}$ & 0.483±0.224 & $+44.81\%^{***}$ & 0.598±0.235 & $+23.92\%^{***}$ \\
          & ResNet-Overlap (ours) &Weak       & \textbf{0.627±0.297} & /     & \textbf{0.699±0.289} & /     & \textbf{0.742±0.283} & / \\
    \midrule
    \multirow{2}[1]{*}{\textbf{Transformer}} & FTTransformer &Sup       & 0.594±0.299 & $+5.50\%^{*}$ & 0.644±0.308 & $+6.61\%^{*}$ & 0.691±0.305 & $+5.65\%^{*}$ \\
          & FTTransformer-Overlap (ours) & Weak      & \textbf{0.627±0.277} & /     & \textbf{0.686±0.282} & /     & \textbf{0.730±0.285} & / \\
    \bottomrule
    \end{tabular}%
\end{table*}%

\subsection{Experimental Results}\label{subsec: experimental results}

\subsubsection{Model Performance} \label{model_performance}
Table~\ref{tab:model_comparison_AUCPR} shows the average model performance over \ndatasets real-world datasets, and we report the full results in the Appendix of supplementary materials.
Above all, we verify the effectiveness of the proposed Overlap loss on various network architectures, including MLP, AutoEncoder, ResNet, and Transformer. The Overlap loss based AD models generally outperform corresponding baselines w.r.t. the ratios of labeled anomalies $\gamma_{l}=5\%$, $\gamma_{l}=10\%$ and $\gamma_{l}=20\%$.

Specifically, experimental results show that the MLP-Overlap achieves a relative improvement $\Delta$ Perf. of AUC-PR over its counterpart DevNet $2.89\%$ and PReNet $1.82\%$ w.r.t. $\gamma_{l}=5\%$. These results indicate that compared to the current state-of-the-art weakly-supervised AD methods, Overlap loss is still more effective when only a handful of labeled anomalies (say $5\%$ labeled anomalies) are available in the training process, considering that such limited label information would bring challenges for estimating anomaly score distribution. 
Besides, we show that end-to-end AD methods, including DevNet, PReNet, and our MLP-Overlap, statistically outperform those unsupervised (e.g. Iforest) or semi-supervised representational learning (e.g., DeepSAD) AD methods, since end-to-end anomaly score learning can leverage the data much more efficiently than the two-step AD approaches \cite{devnet, devnet_explainable}.

For $\gamma_{l}=10\%$, the relative improvement of Overlap loss based model is more significant, as more labeled anomalies are beneficial for the Overlap loss to estimate more accurate anomaly score distributions, and thus to better measure the score distribution overlap. The MLP-Overlap $\Delta$ Perf. of AUC-PR over DevNet is $7.75\%$ and $5.67\%$ over PReNet w.r.t. $\gamma_{l}=10\%$.
Furthermore, we prove the superiority of the proposed Overlap loss on other network architectures such as AutoEncoder and ResNet. In terms of AUC-PR, the $\Delta$ Perf. of AE-Overlap over fully- and weakly-supervised FEAWAD is $28.04\%$ and $9.29\%$, respectively, and $\Delta$ Perf. of ResNet-Overlap over ResNet is $56.30\%$, where all the relative improvements are significant at $1\%$ significance level. Although FTTransformer has been proven to be a strong solution for tabular-based tasks \cite{gorishniy2021revisiting}, we still observe $\Delta$ Perf. of FTTransformer-Overlap over FTTransformer $5.50\%$\textasciitilde$6.61\%$ on AUC-PR.

\subsubsection{Runtime Analysis}
\begin{figure}[h]
    \centering
    \includegraphics[width=8.2cm]{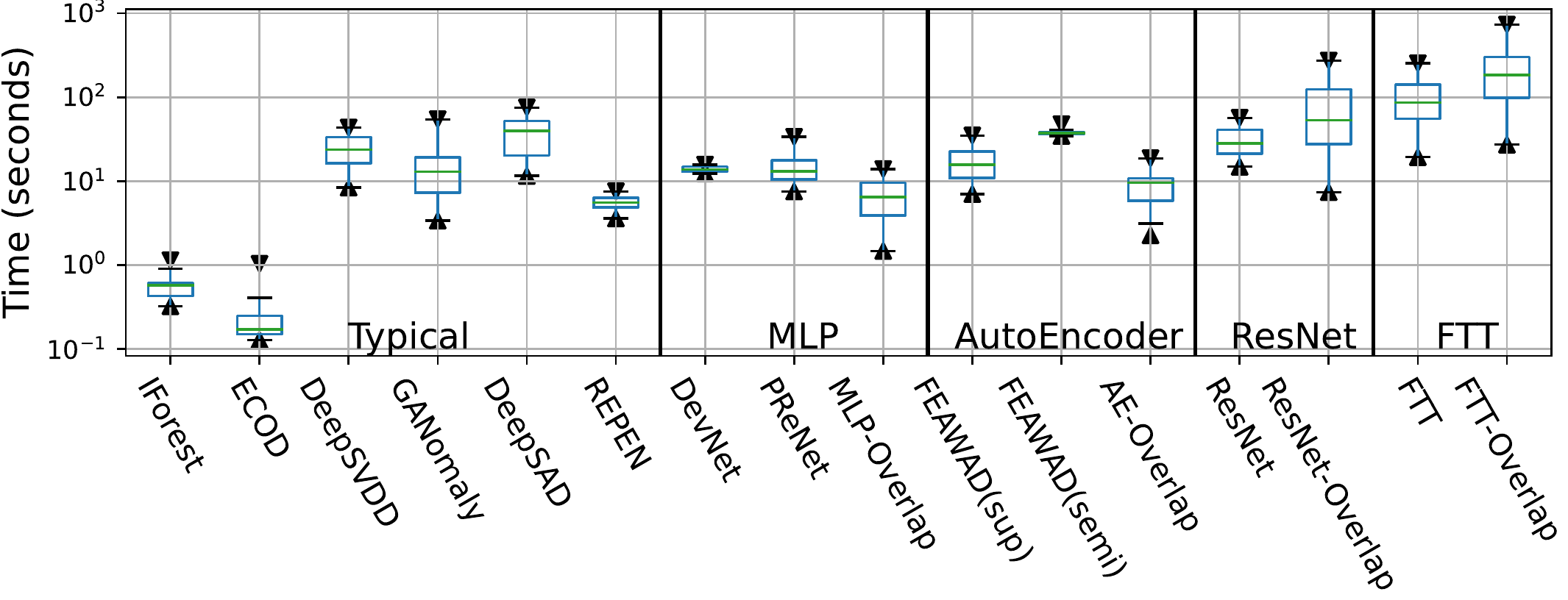}
    \vspace{-0.15in}
    \caption{Boxplot of model training time}
    \label{fig:runtime}
\end{figure}

We show the model training time in Figure~\ref{fig:runtime}. This result shows that ECOD is the fastest algorithm as it treats each feature independently. Both MLP-Overlap and AE-Overlap are faster than their counterparts, since our methods need fewer training epochs while achieving better detection performance. For ResNet and FTTransformer (FTT) architectures, our methods are comparable to or relatively slower than the counterparts. This is mainly due to the fact that Overlap loss requires more training epochs for more complex network architectures (especially for FTTransformer) than those simple architectures like MLP. Therefore we apply the same training strategy (100 epochs with early stopping) for ResNet and FTTransformer, as well as our Overlap loss based versions ResNet-Overlap and FTTransformer-Overlap. The extra training time is mainly caused by the calculation of Overlap loss, compared to the supervised binary cross entropy loss.

\vspace{-0.2in}


\begin{table*}[h!]
\footnotesize
  \centering
  \caption{AUC-PR results of ablation studies. Overlap-Gaussian refers to the basic method mentioned in Section~\ref{subsec: score distribution estimator}. Overlap-Arbitrary refers to the basic method of Eq.\ref{Eq:overlap_wo_order}. Overlap-Ranking isolates the ranking loss in Eq.\ref{Eq:overlap_w_order}. Overlap-Combined corresponds to the combined loss form of both Overlap-Arbitrary and Overlap-Ranking as illustrated in Eq.\ref{Eq:overlap_w_order}. Overlap-Proposed refers to the final solution in this paper.}
  \label{tab:ablation_AUCPR}%
  \begin{tabular}{lccccc|ccccc|ccccc}
    \toprule
    \multicolumn{1}{c}{\multirow{2}{*}{\textbf{Method}}} & \multicolumn{5}{c}{$\gamma_{l}=5\%$}            & \multicolumn{5}{c}{$\gamma_{l}=10\%$}            & \multicolumn{5}{c}{$\gamma_{l}=20\%$} \\
\cmidrule{2-16}          & \textbf{VAE} & \textbf{MLP} & \textbf{AE} & \textbf{ResNet} & \textbf{FTT} & \textbf{VAE} & \textbf{MLP} & \textbf{AE} & \textbf{ResNet} & \textbf{FTT} & \textbf{VAE} & \textbf{MLP} & \textbf{AE} & \textbf{ResNet} & \textbf{FTT} \\
    \midrule
    Overlap-Gaussian & 0.159  & /     & /     & /     & /     & 0.158  & /     & /     & /     & /     & 0.158  & /     & /     & /     & / \\
    Overlap-Arbitrary & /     & 0.349  & 0.376  & 0.341  & 0.356  & /     & 0.376  & 0.416  & 0.407  & 0.339  & /     & 0.397  & 0.410  & 0.426  & 0.352  \\
    Overlap-Ranking & /     & 0.535  & 0.558  & 0.539  & 0.623  & /     & 0.605  & 0.620  & 0.625  & \textbf{0.687}  & /     & 0.657  & 0.657  & 0.691  & \textbf{0.731}  \\
    Overlap-Combined & /     & \textbf{0.624}  & 0.642  & 0.604  & 0.418  & /     & \textbf{0.682}  & \textbf{0.695}  & 0.686  & 0.439  & /     & \textbf{0.708}  & \textbf{0.719}  & 0.741  & 0.459  \\
    Overlap-Proposed & /     & 0.623  & \textbf{0.652}  & \textbf{0.627}  & \textbf{0.627}  & /     & 0.674  & \textbf{0.695}  & \textbf{0.699}  & 0.686  & /     & 0.696  & 0.713  & \textbf{0.742}  & 0.730  \\
    \bottomrule
    \end{tabular}%
\end{table*}%

\subsubsection{Ablation Study}
\label{further_analysis_alternative_methods}
In Table~\ref{tab:ablation_AUCPR}, we report the AUC-PR results of several basic methods mentioned in Section \ref{setting}. We instantiate the basic method of Overlap-Gaussian by replacing the scoring layer $\eta\left(\cdot ; \Theta_{s}\right)$ with the VAE \cite{VAE} structure, where the anomaly scores of normal and abnormal data are sampled from their corresponding Gaussian distribution via the reparameterization trick \cite{VAE}. The calculated intersection point $c$ of Eq.\ref{Eq:intersection_point_Gaussian} can be used for estimating score distribution overlap via Eq.\ref{Eq:overlap}.

\textit{First}, we observe that Overlap-Gaussian has the worst performance. This is because the scarceness of labeled anomalies makes its score distribution often present a certain arbitrariness, whereas the Gaussian assumption is detrimental to the representation of scoring function.
\textit{Second}, the disorder in anomaly scores leads to performance degradation in the Overlap-Arbitrary. Ranking loss term can be served as an effective way to guarantee the order in anomaly scores, as the Overlap-Combined method significantly improves the AUC-PR performance.
\textit{Third}, the proposed Overlap loss outperforms all basic methods in most cases, since it can effectively estimate arbitrary score distributions of output anomaly scores while avoiding the score disorder problem that occurs in the Overlap-Arbitrary method. Compared to the Overlap-Combined method, Overlap-Proposed achieves better performance, probably because it realizes a unified loss function form, rather than a combination of two different loss parts. Besides, we observe that the multi-task loss form in the Overlap-Combined method fails in more complex network backbones like FTTransformer.
\vspace{-0.15in}

\subsection{Further Exploration into AD Loss Functions}
\label{additional results}
While most of the existing research focuses on proposing and evaluating specific models or architectural designs of AD methods \cite{you2020design}, we manage to go a step further and directly compare different loss functions in the same network architecture. We introduce the decoupling methods in the Neural Architecture Search (NAS) problem \cite{liu2018progressive,elsken2019neural}, where we mainly concern the design space of loss functions instead of other perspectives like architecture settings \cite{li2021autood}. Such an analytical method could eliminate the effects of model configurations such as dropout and activation layers, while fully focusing on the role of loss functions (i.e., training objectives) in the anomaly detection tasks.

\begin{table}[h]
\footnotesize
  \centering
  \caption{Summary of decoupled loss functions. The No Prior column indicates whether the prior anomaly score target is needed in the corresponding loss functions, e.g., the margin hyperparameter $M$ in the Hinge loss.}
  \vspace{-0.1in}
    \begin{tabular}{llc}
    \toprule
    \textbf{Loss} & \textbf{Formula} & \textbf{No Prior} \\
    \midrule
    Minus & $\mathcal{L}=\left|s_{n}\right|+\max \left(0, B N D-\left|s_{a}\right|\right)$      & \XSolidBrush \\
    Inverse & $\mathcal{L}=\left|s_{n}\right|+1 /\left|s_{a}\right|$      & \checkmark \\
    Hinge & $\mathcal{L}=\max \left(0, M+s_{n}-s_{a}\right)$      & \XSolidBrush \\
    Deviation & $\mathcal{L}=\left|s_{n}\right|+\max \left(0, M-s_{a}\right)$      & \XSolidBrush \\
    Ordinal & 
    $\mathcal{L}=\left|s_{n, n}-\mu_{n, n}\right|+\left|s_{a, n}-\mu_{a, n}\right|+
    \left|s_{a, a}-\mu_{a, a}\right|$
         & \XSolidBrush \\
    Overlap & Eq.\ref{Eq:objective}     & \checkmark \\
    
    \bottomrule
    \end{tabular}%
  \label{loss_comparison_formula}%
\end{table}%

We decouple the loss functions in several popular AD models mentioned in Figure~\ref{fig:loss_comparison}, including the Minus loss in Unlearning \cite{Unlearning}, Inverse loss in DeepSAD \cite{ruff2019deep}, Hinge loss in REPEN \cite{pang-ultra-high}, Deviation loss in DevNet \cite{devnet} and FEAWAD \cite{FEAWAD}, Ordinal loss in PReNet \cite{pang2019deep_weak}, and our proposed Overlap loss, as shown in Table~\ref{loss_comparison_formula}. For the consistency of comparison, we replace the original reconstruction error in Minus loss and the Euclidean distance of embedding in Inverse loss with the absolute anomaly score.
A network with one-hidden-layer of 20 neurons is applied to ensure the comparability of different loss functions. The ReLU activation layer is employed in this network. We train the network for 200 epochs of 256 batch size and use the SGD optimizer with 0.01 learning rate and 0.7 momentum. The weight decay is set to 0.01. The hyperparameter $BND$ in the Minus loss is set to 5, and the hyperparameter $M$ in the Hinge and Deviation loss is set to 5. The anomaly scores in Deviation loss are normalized as Z-Score.
Consistent with the original paper, we set $\mu_{n,n}$, $\mu_{n,n}$ and $\mu_{n,n}$ in the Ordinal loss to 0, 4, and 8, respectively. We first investigate different loss functions on \ndatasets real-world datasets and then report their performances in detecting various types of anomalies.




\subsubsection{Exploration of AD Loss Functions on Real-world Datasets}\label{exploration_real_world}
In this subsection, we analyze different loss functions on real-world datasets with respect to the following two perspectives: (i) Embedding transformation. The transformed embedding of input features \cite{GANomaly,Skip-GANomaly} can be seen as a visualization of representation layer variation for realizing the training objective. (ii) Network Parameter Changes. This is often discussed in the continual learning problem, where drastic changes in network parameters may suffer from the problem of catastrophic forgetting \cite{aljundi2018memory,zenke2017continual,kirkpatrick2017overcoming,Unlearning}. Similarly, we investigate the network parameter changes of different loss function based AD models when achieving their corresponding training objectives.





\begin{figure*}[t]
     \centering
     \begin{minipage}[c]{0.12\textwidth}
        \begin{subfigure}[c]{0.96\textwidth}
            \centering
            \includegraphics[width=\textwidth]{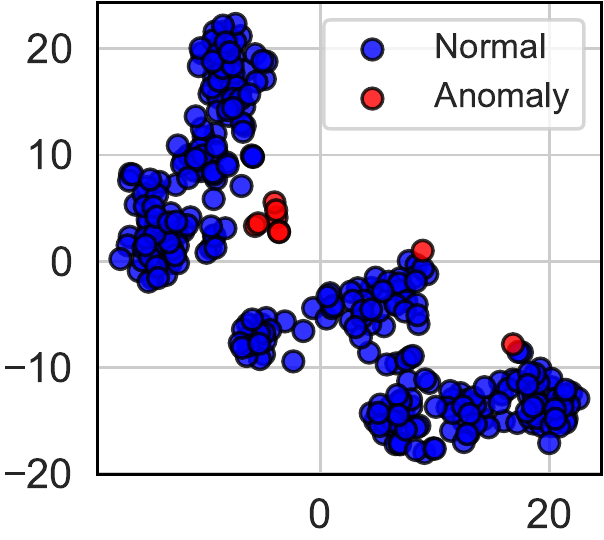}
            \caption{t-SNE \cite{TSNE} plots of the input feature of vowels dataset.}
            \label{fig:Embedding_input}
        \end{subfigure}
     \end{minipage}
     \hspace{0.1in}
     \begin{minipage}[c]{0.85\textwidth}
     \begin{subfigure}[b]{0.3\textwidth}
         \centering
         \includegraphics[width=\textwidth]{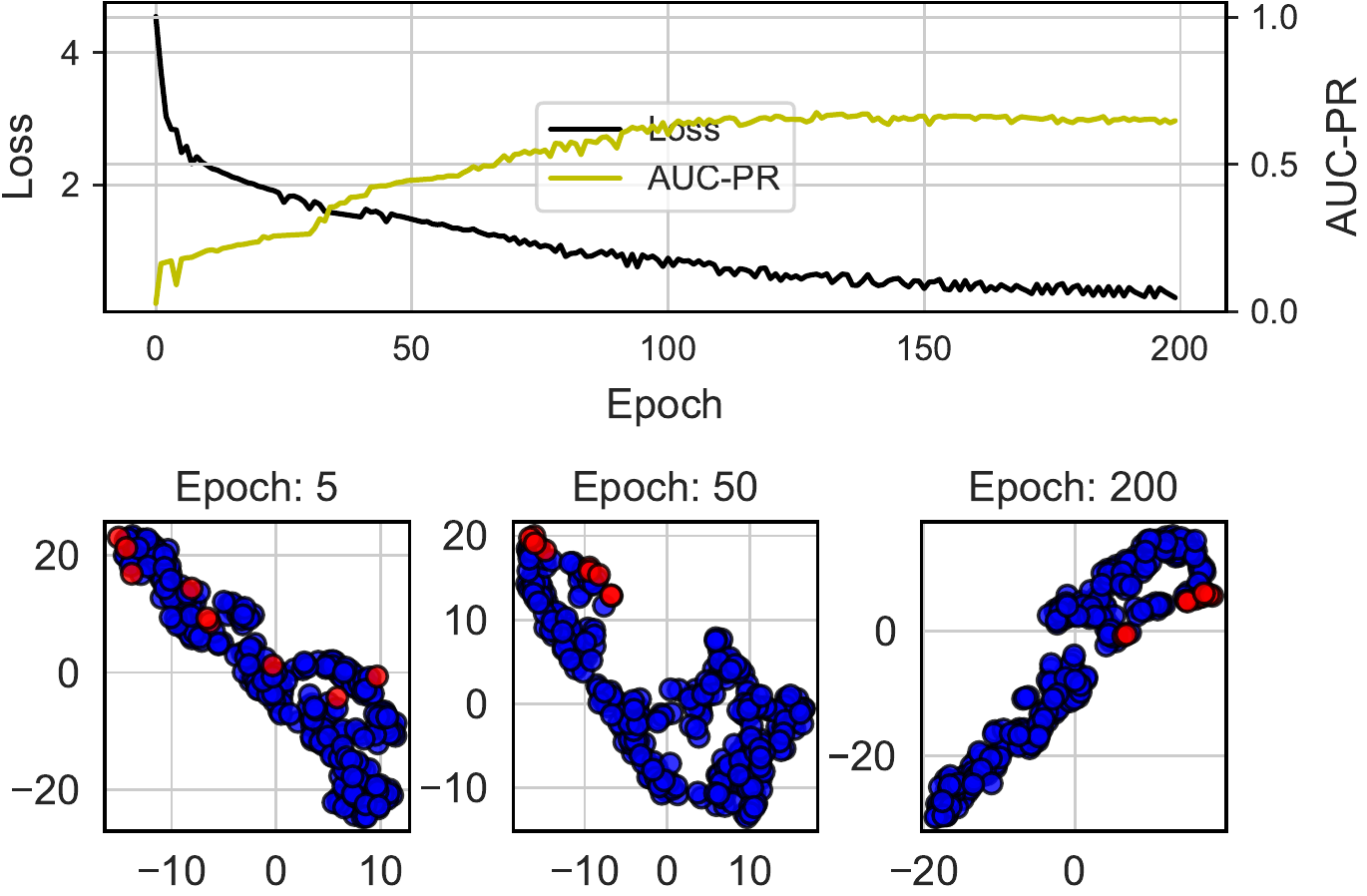}
         \caption{Minus}
         \label{fig:}
     \end{subfigure}
     \hspace{0.2em}
     \begin{subfigure}[b]{0.3\textwidth}
         \centering
         \includegraphics[width=\textwidth]{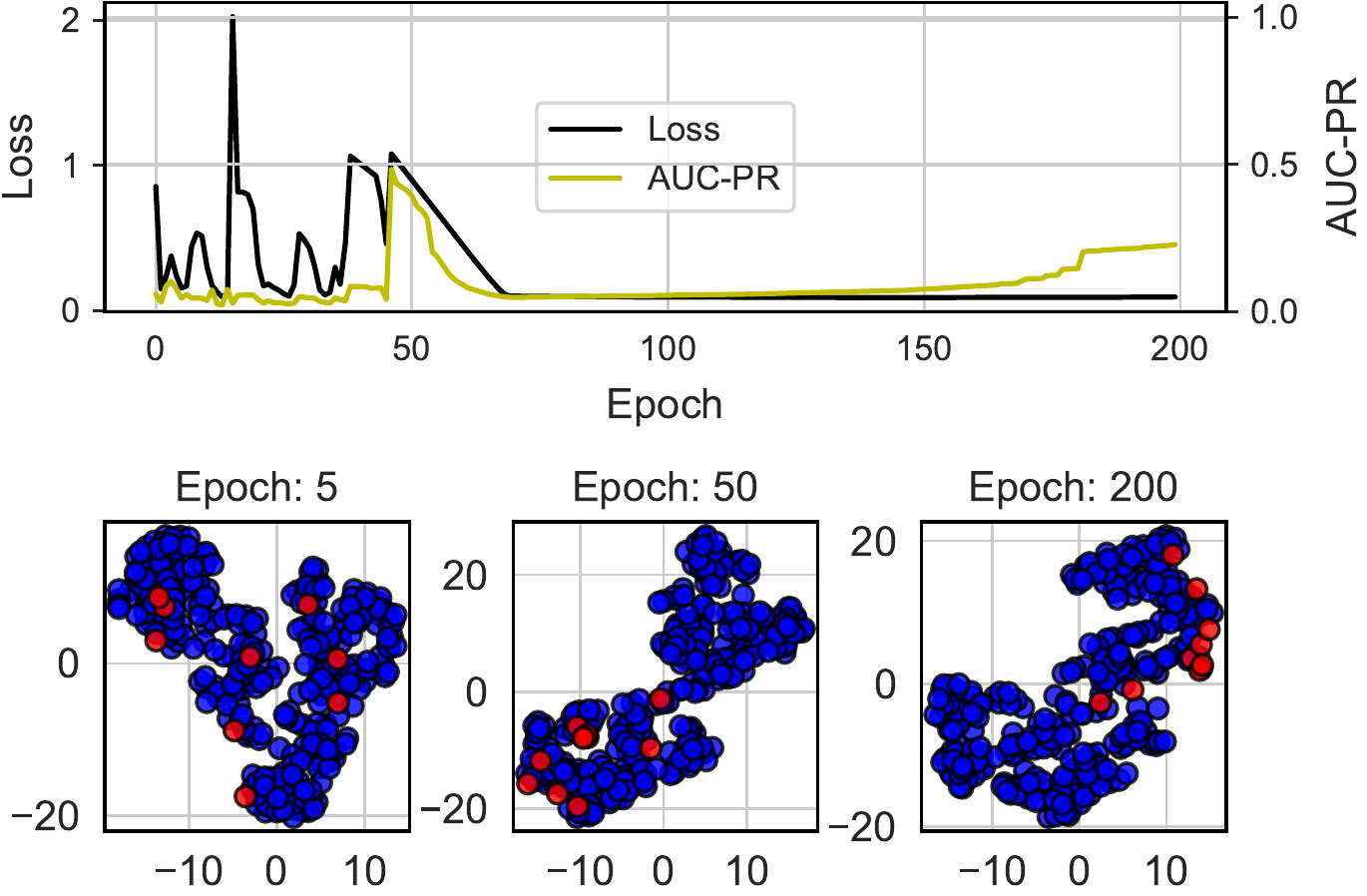}
         \caption{Inverse}
         \label{fig:}
     \end{subfigure}
     \hspace{0.2em}
     \begin{subfigure}[b]{0.3\textwidth}
         \centering
         \includegraphics[width=\textwidth]{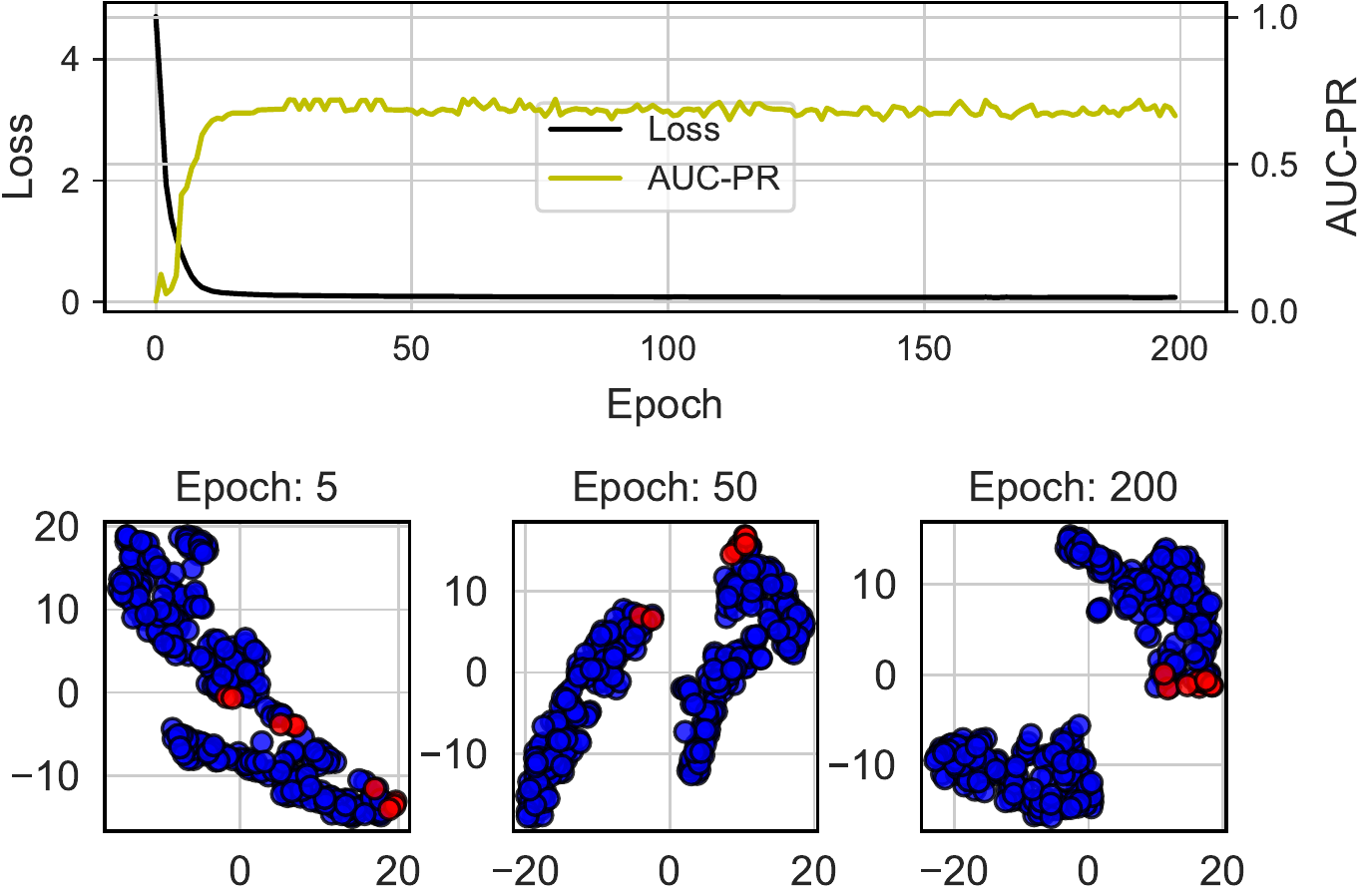}
         \caption{Hinge}
         \label{fig:}
     \end{subfigure}
     \hspace{0.2em}
     
     \vspace{0.5em}
     \begin{subfigure}[b]{0.3\textwidth}
         \centering
         \includegraphics[width=\textwidth]{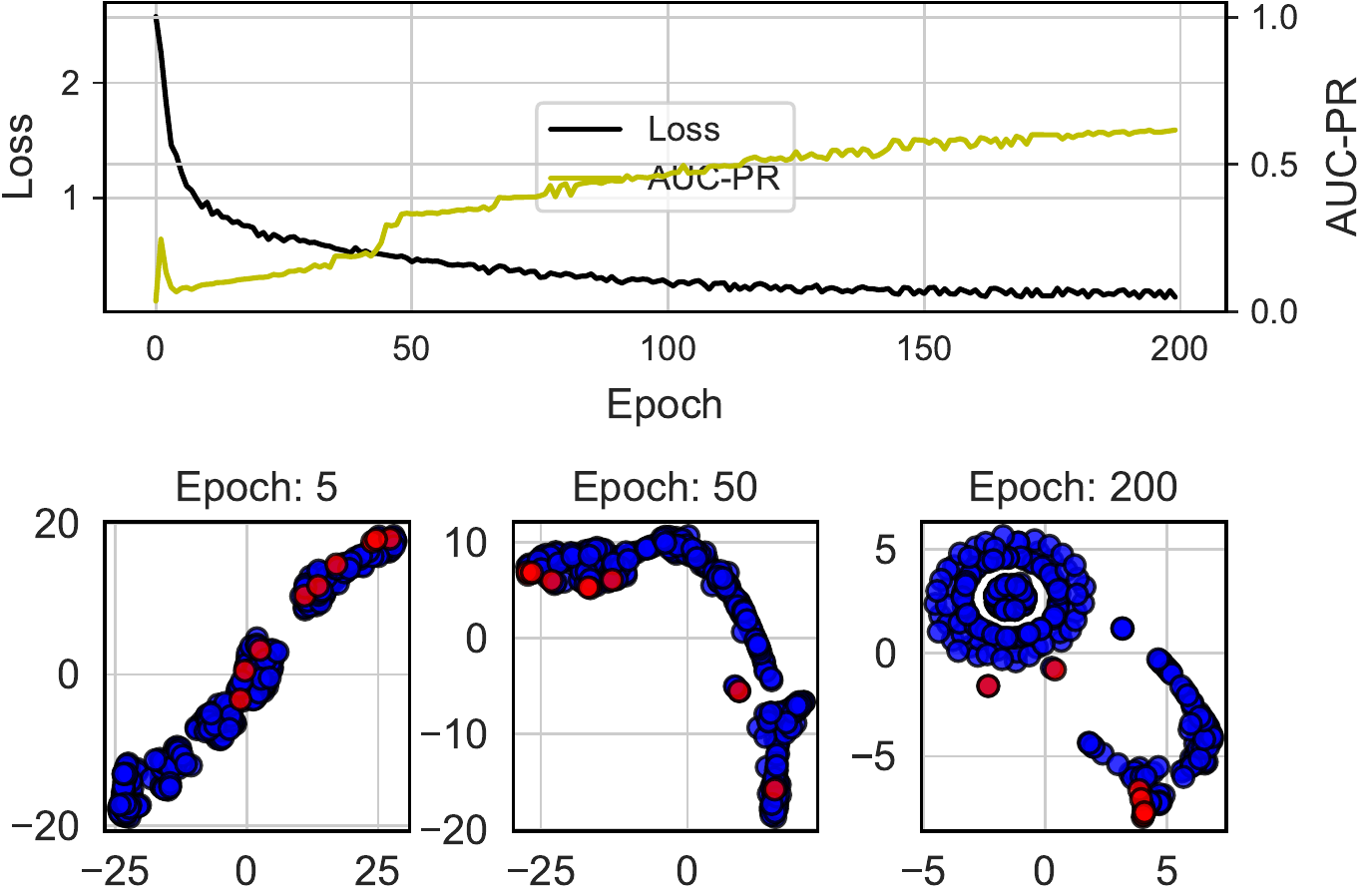}
         \caption{Deviation}
         \label{fig:}
     \end{subfigure}
     \hspace{0.2em}
     \begin{subfigure}[b]{0.3\textwidth}
         \centering
         \includegraphics[width=\textwidth]{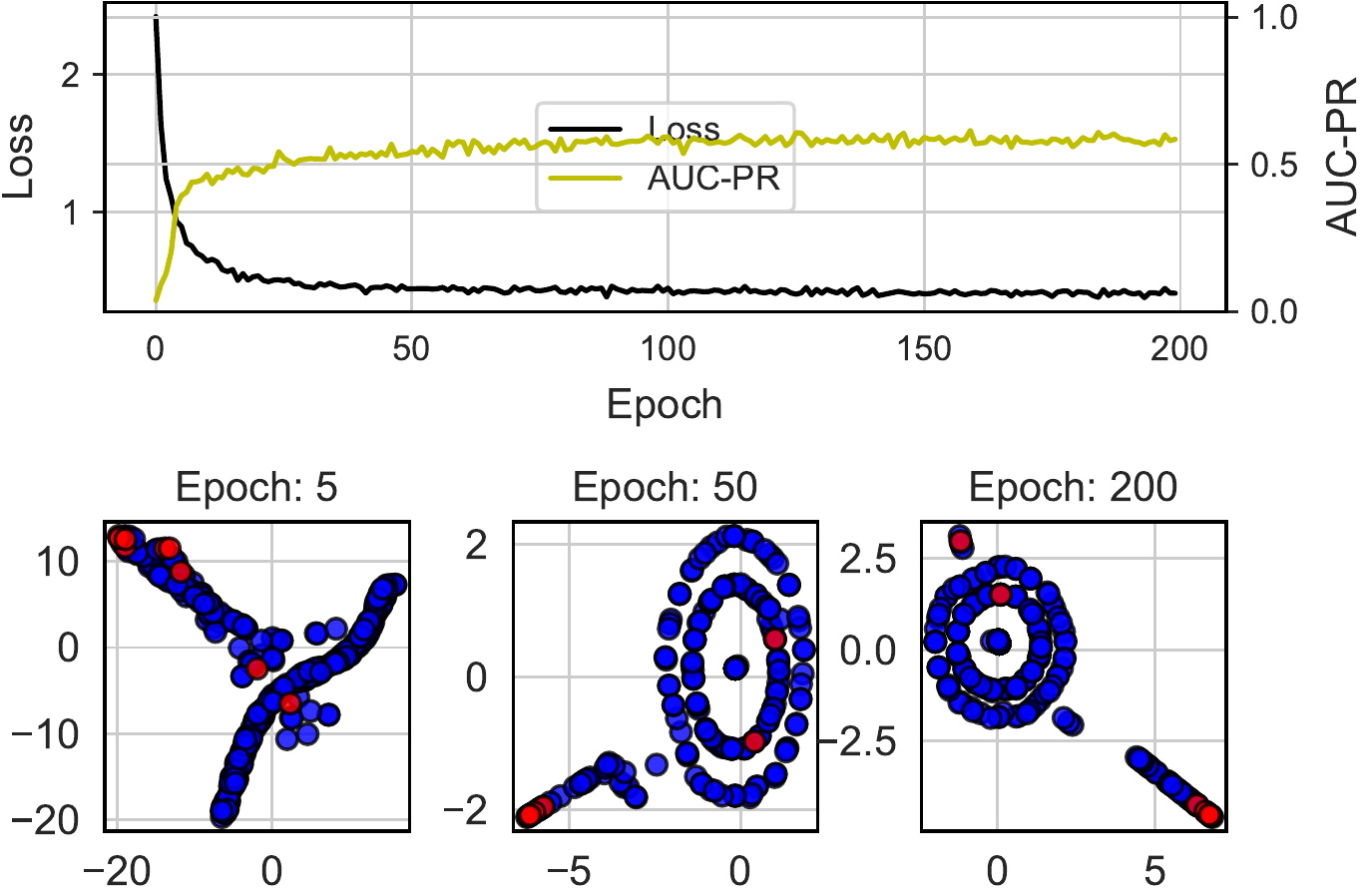}
         \caption{Ordinal}
         \label{fig:}
     \end{subfigure}
     \hspace{0.2em}
     \begin{subfigure}[b]{0.3\textwidth}
         \centering
         \includegraphics[width=\textwidth]{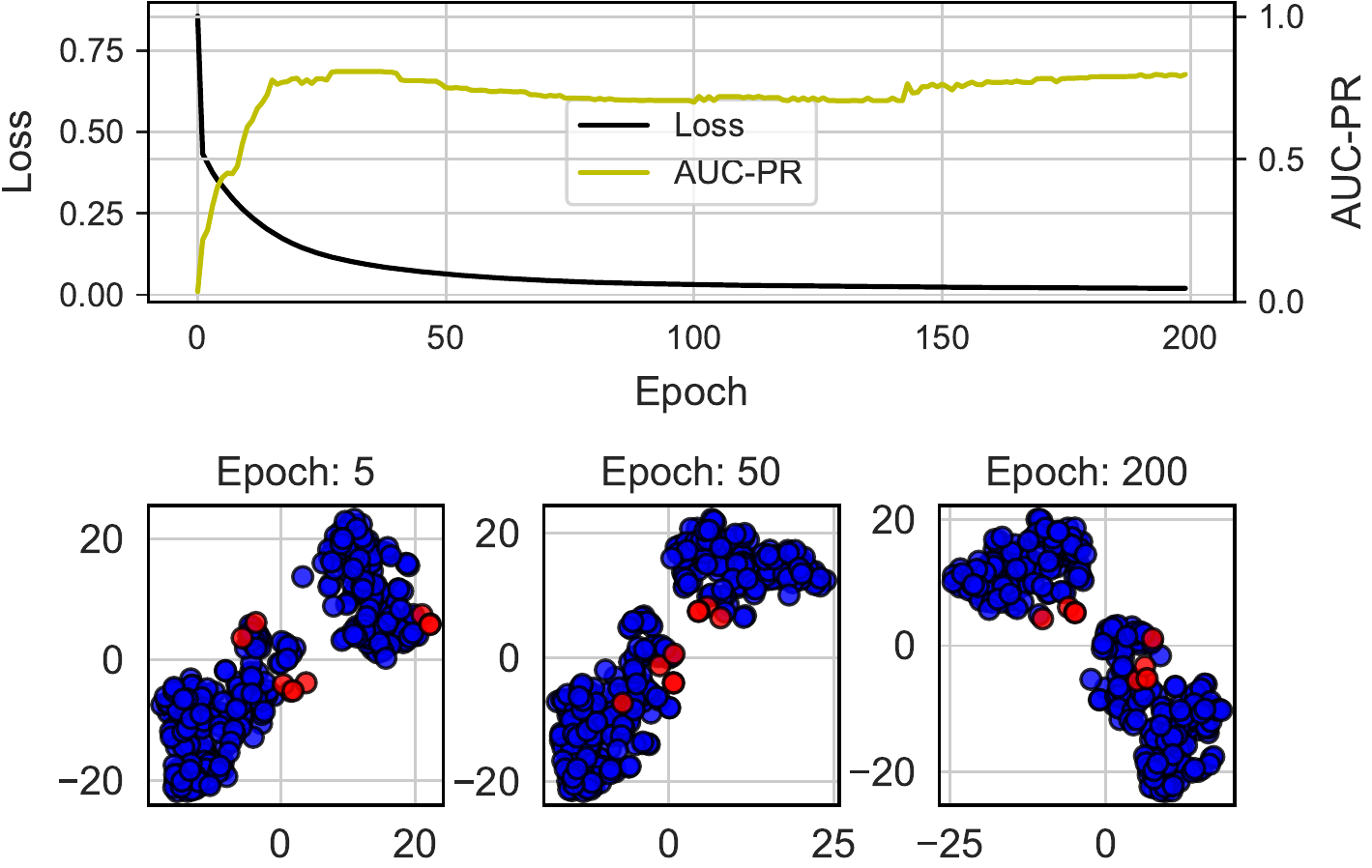}
         \caption{Overlap}
         \label{fig:}
     \end{subfigure}
     \hspace{0.2em}
    \end{minipage}
    \vspace{-0.1in}
    \caption{Training loss along with the AUC-PR performance on testing set of different loss function based AD models, where the vowels dataset is specified for comparison. The transformed embeddings of the input feature are demonstrated, which corresponds to 5, 50, and 200 training epochs, respectively. See the additional results in Appendix.}
    \label{fig:Loss}
\end{figure*}

\textbf{Embedding Transformation during Model Training}. We take the vowels dataset as an example to demonstrate the embedding transformation in the feature representation layer during the training process, as shown in Figure~\ref{fig:Loss}.
Similar experimental results can be observed in other real-world datasets, and we provide these results in the Appendix of supplementary materials.
Figure~\ref{fig:Loss} indicates that the Deviation and Ordinal loss tend to seriously distort the embedding of original input data after a few training epochs. This is due to the fact that these two loss functions explicitly guide networks to map the anomaly score of each instance or pair to one or more fixed score constants or a score margin, thus hindering the diversity of learned feature representation. Besides, the unlabeled normal samples are contaminated by the unlabeled anomalies, and defining an identical training target for these two types of data would limit the representational ability of the learned models.

Our proposed Overlap loss generates a relatively mild transformation of the input features. As mentioned before, Overlap loss based AD models can achieve superior detection performance, therefore we speculate that good detection performance \textit{does not} always require an excessive transformation in the representation space, where the model can merely transform the embeddings that have the most impact on the score distribution discrimination and remain more fine-grained information of input data. 

\begin{figure}[h]
    \begin{minipage}[c]{0.25\textwidth}
        \centering
        \includegraphics[width=4.5cm]{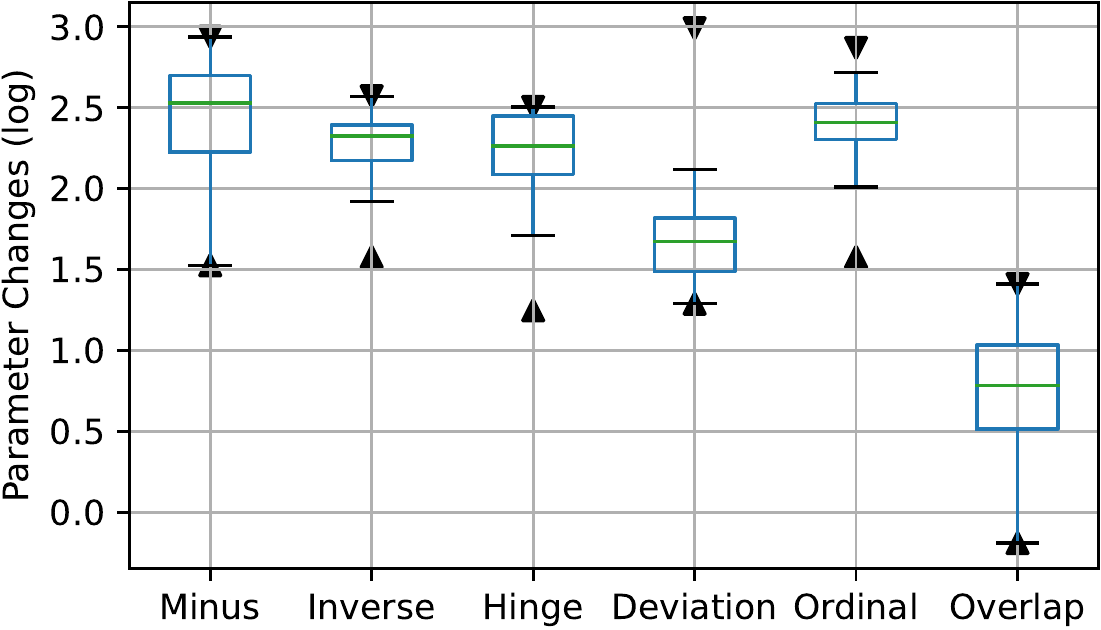}
    \end{minipage}
    \hfill
    \begin{minipage}[c]{0.21\textwidth}
        \caption{Network parameter changes in the training stage.}
        \label{fig:model_parameter}
    \end{minipage}
\end{figure}

\textbf{Network Parameter Changes}. We show the results of network parameter changes on the \ndatasets real-world datasets in Figure~\ref{fig:model_parameter}, where the sum of norms of parameter differences in each layer are calculated between the initialized model and its updated version as $\sum_{l}\left\|M-M_{0}\right\|_{2}^{2}$. The result indicates that compared to the other loss functions, the AD model based on our proposed Overlap loss inherits smaller network parameter changes.
This result corresponds to the good properties of Overlap loss, where (i) Overlap loss is naturally bounded, avoiding drastic updating of anomaly scores (and network parameters). (ii) Overlap loss does not require the prior target of anomaly score, therefore reducing unnecessary scoring function updates in the training stage and being capable of adapting to different datasets with minimum adjustment of the score distribution.

\subsubsection{Exploration of AD Loss Functions on Different Types of Anomalies}
While extensive AD methods have been proven to be effective on real-world datasets, previous studies often neglect to discuss the pros and cons of AD methods regarding specific types of anomalies \cite{gopalan2019pidforest, steinbuss2021benchmarking}. In fact, public datasets often consist of a mixture of different types of anomalies. We follow \cite{steinbuss2021benchmarking, han2022adbench} to create realistic synthetic datasets based on the above \ndatasets datasets by injecting four types (namely local, global, clustered, and dependency) of anomalies to evaluate different loss functions. Appendix~\ref{subsec: appendix_further_explore} provides detailed information of generated synthetic anomalies.
\vspace{-0.1in}

\begin{table}[h!]
\small
  \centering
  \caption{Loss comparison on different types of anomalies.}
  \vspace{-0.1in}
  \label{tab:type_AUCPR}
  \centering
    \begin{tabular}{lcccc}
    \toprule
    \textbf{Loss}&
    \multicolumn{1}{l}{\textbf{Local}} &
    \multicolumn{1}{l}{\textbf{Global}} & \multicolumn{1}{l}{\textbf{Clustered}} & \multicolumn{1}{l}{\textbf{Dependency}} \\
    \midrule
    Minus & 0.255 & 0.822 & 0.992 & 0.369 \\
    Inverse & 0.235 & 0.647 & 0.900   & 0.198 \\
    Hinge & 0.271 & 0.853 & 0.996 & 0.413 \\
    Deviation & 0.246 & 0.851 & 0.987 & 0.303 \\
    Ordinal & 0.247 & 0.849 & 0.991 & 0.327 \\
    Overlap & \textbf{0.439} & \textbf{0.929} & \textbf{0.998} & \textbf{0.571} \\
    \bottomrule
    \end{tabular}%
\end{table}%
\vspace{-0.1in}
Table~\ref{tab:type_AUCPR} shows the AUC-PR results of loss function comparison on different types of anomalies. These results are consistent with the findings in \cite{han2022adbench}, where the other loss functions devised for semi- or weakly-supervised AD algorithms perform relatively poorly on local and dependency anomalies. Unlike clustered anomalies, the partially labeled anomalies of local and dependency anomalies can not well capture all characteristics of specific types of anomalies, and learning such decision boundaries for separating normal and abnormal data is often challenging (see Figure~\ref{fig:local_minus}\textasciitilde\ref{fig:local_ordinal} in the Appendix). Therefore, the incomplete label information may bias the learning process of these loss functions, which explains their relatively inferior performances compared to the Overlap loss.

In contrast, Overlap loss performs significantly better on local, global, and dependency anomalies, and achieves satisfactory results on clustered anomalies. For example, the average AUC-PR of Overlap loss on the local anomalies is 0.439, compared to the second-best Hinge loss of 0.271. Similar result can be observed for the dependency anomalies, where the AUC-PR of Overlap loss is 0.571, compared to the second-best Hinge loss of 0.413.

Such results verify that Overlap loss can effectively leverage the prior knowledge of both partial labels and anomaly types. That is to say, Overlap loss based AD models (e.g., ResNet-Overlap) can achieve superior performance when only a limited number of labeled anomalies are available during the training stage. Furthermore, if one could get access to the valuable prior knowledge of anomaly types \cite{han2022adbench}, Overlap loss can be served as an effective solution to learn the pattern of this specific type (e.g., dependency) of anomalies.

\vspace{-0.08in}
\section{Conclusion} \label{conclusion}

In this paper, we propose a novel loss function called Overlap loss for AD tasks. Overlap loss liberates the AD models from the predefined anomaly score targets, e.g., predefined constant or margin hyperparameter(s), thus adapting well to various datasets. By directly optimizing distribution overlap to realize score distribution discrimination, Overlap loss can retain more fine-grained information of input data, and also avoids dramatic changes in network parameters which may lead to overfitting or catastrophic forgetting problem.
Extensive experimental results verify that the proposed Overlap loss can be effectively instantiated to different network architectures, including MLP, AutoEncoder, ResNet, and Transformer. Moreover, Overlap loss significantly outperforms other popular AD loss functions on various types of anomalies.

For the future, we plan to improve the optimization process of Overlap loss by leveraging more complex score distribution estimators, such as the Gaussian Mixture Model (GMM) \cite{reynolds2009gaussian}. Besides, we will extend our work to more general scenarios in weakly-supervised AD tasks \cite{zhou2018brief}, such as the inaccurate supervision \cite{zhao2022admoe} and inexact supervision \cite{sultani2018real,lee2021weakly} problems.

\vspace{-0.068in}
\section{Acknowledgments}
We thank anonymous reviewers for their insightful comments and discussions. We appreciate the paper suggestions from all SUFE AI Lab members. Hailiang Huang is supported by the National Natural Science Foundation of China under Grant No. 72271151. We thank the financial support provided by FlagInfo-SHUFE Joint Laboratory. 

\clearpage
\bibliographystyle{ACM-Reference-Format}
\balance
\bibliography{reference}

\clearpage
\appendix
\section*{Appendix}

\setcounter{table}{0}
\setcounter{figure}{0}

\renewcommand{\thetable}{\Alph{section}A\arabic{table}}
\renewcommand{\thefigure}{\Alph{section}A\arabic{figure}}

\subsection{Details of Dataset Description}
\label{subsec: appendix_dataset_description}
We show the detailed description of \ndatasets publicly available real-world datasets in Table~\ref{tab:data description}, which include several domains such as disease diagnosis, speech recognition, and image identification.
\vspace{-0.1in}
\begin{table}[h!]
\footnotesize
  \centering
  \caption{Dataset description.}
  \vspace{-0.1in}
    \begin{tabular}{lcccc}
    \toprule
    \textbf{Dataset} & $N$ & $D$ & \#anomalies & \#anomaly ratio (\%) \\
    \midrule
    ALOI  & 49534 & 27    & 1508  & 3.04  \\
    annthyroid & 7200  & 6     & 534   & 7.42  \\
    Cardiotocography & 2114  & 21    & 466   & 22.04  \\
    fault & 1941  & 27    & 673   & 34.67  \\
    http  & 567498 & 3     & 2211  & 0.39  \\
    landsat & 6435  & 36    & 1333  & 20.71  \\
    letter & 1600  & 32    & 100   & 6.25  \\
    magic.gamma & 19020 & 10    & 6688  & 35.16  \\
    mammography & 11183 & 6     & 260   & 2.32  \\
    mnist & 7603  & 100   & 700   & 9.21  \\
    musk  & 3062  & 166   & 97    & 3.17  \\
    optdigits & 5216  & 64    & 150   & 2.88  \\
    PageBlocks & 5393  & 10    & 510   & 9.46  \\
    pendigits & 6870  & 16    & 156   & 2.27  \\
    satellite & 6435  & 36    & 2036  & 31.64  \\
    satimage-2 & 5803  & 36    & 71    & 1.22  \\
    shuttle & 49097 & 9     & 3511  & 7.15  \\
    skin  & 245057 & 3     & 50859 & 20.75  \\
    SpamBase & 4207  & 57    & 1679  & 39.91  \\
    speech & 3686  & 400   & 61    & 1.65  \\
    thyroid & 3772  & 6     & 93    & 2.47  \\
    vowels & 1456  & 12    & 50    & 3.43  \\
    Waveform & 3443  & 21    & 100   & 2.90  \\
    Wilt  & 4819  & 5     & 257   & 5.33  \\
    yeast & 1484  & 8     & 507   & 34.16  \\
    \bottomrule
    \end{tabular}%
  \label{tab:data description}%
\end{table}%
\vspace{-0.1in}

\subsection{Additional Training Details} \label{subsec: appendix_training_details}
For unsupervised baselines, Iforest, ECOD, and DeepSVDD are built using the PyOD \cite{zhao2019pyod} library. Labeled anomalies are combined with unlabeled data for constructing the validation set, in order to tune the hyperparameters of these unsupervised methods via the grid search method, since tuning their hyperparameters on a small validation set often yields better performance than using the default settings \cite{effect_hyper_unsupervise}. Table \ref{tab:hyper-parameter} shows the hyperparameter grids, where ECOD is not considered since it is a parameter-free method.
\begin{table}[h!]
\small
  \centering
  \caption{Hyperparameter grid of the unsupervised models.}
  \vspace{-0.1in}
    \begin{tabular}{cc}
    \toprule
    \textbf{Model} & \textbf{Hyperparameter} \\
    \midrule
    Iforest & n\_estimators: [10, 50, 100, 500] \\
    DeepSVDD & epochs: [20, 50, 100, 200] \\
    \bottomrule
    \end{tabular}%
  \label{tab:hyper-parameter}%
\end{table}%

We replace the convolutional layer in the original GANomaly with the dense layer for evaluating it on the tabular data, where the hidden size of the encoder-decoder-encoder structure of GANomaly is set to half of the input dimension. We realize the PReNet in PyTorch as we do not find the open-source codes, and set the hyperparameters in PReNet according to its original paper.
Other models are built based on their corresponding source codes. If not specified, we train these baseline models according to their default hyperparameters mentioned in the original papers.

\subsection{Experimental Results of Selecting Intersection Points}
We demonstrate the experimental results of different intersection point selection strategies in Table~\ref{tab:appendix_intersection}. The detection performance of a randomly selected intersection point is very close to that of ensembling all the calculated intersection points w.r.t. different ratios of labeled anomalies $\gamma_{l}=5\%$, $\gamma_{l}=10\%$ and $\gamma_{l}=20\%$. This is due to the fact that the calculation of the overlap area repeats many times (epochs$\times$batchsize), which is essentially similar to the average results of the ensemble strategy. Moreover, random sampling of the intersection points can improve computational efficiency since the overlap area only needs to be estimated once in a training batch.
\begin{table}[htbp]
\small
  \centering
  \caption{AUC-ROC and AUC-PR results of different intersection point selection strategies. MLP-Overlap corresponds to the default strategy that randomly chooses one of the intersection points for estimating the overlap area via Eq.\ref{Eq:objective}. MLP-Overlap-E refers to the ensemble strategy by taking the average of the overlap areas calculated based on each intersection point.}
  \label{tab:appendix_intersection}
  \vspace{-0.1in}
  \subcaption{AUC-ROC results.}
  \begin{subtable}{0.4\textwidth}
  \centering
    \begin{tabular}{llll}
    \toprule
          & $\gamma_{l}=5\%$ & $\gamma_{l}=10\%$ & $\gamma_{l}=20\%$ \\
    \midrule
    MLP-Overlap & 0.847±0.145 & 0.880±0.132 & 0.893±0.133 \\
    MLP-Overlap-E & 0.849±0.146 & 0.879±0.132 & 0.894±0.131 \\
    \bottomrule
    \end{tabular}%
  \end{subtable}

  \vspace{0.03in}
  \subcaption{AUC-PR results.}
  \begin{subtable}{0.4\textwidth}
  \centering
    \begin{tabular}{llll}
    \toprule
          & $\gamma_{l}=5\%$ & $\gamma_{l}=10\%$ & $\gamma_{l}=20\%$ \\
    \midrule
    MLP-Overlap & 0.623±0.291 & 0.674±0.286 & 0.696±0.288 \\
    MLP-Overlap-E & 0.626±0.291 & 0.673±0.285 & 0.694±0.287 \\
    \bottomrule
    \end{tabular}%
  \end{subtable}
\end{table}%

\subsection{Detailed Experimental Results}
We show the AUC-ROC results of model performance in Table~\ref{tab:model_comparison_AUCROC} and ablation study in Table~\ref{tab:ablation_AUCROC}, corresponding to Section \ref{model_performance} and \ref{further_analysis_alternative_methods} in the main paper, respectively. These experimental results are basically consistent with the main paper. Besides, we show the detailed results of model comparison on \ndatasets real-world datasets w.r.t. $\gamma_{l}=5\%$, $\gamma_{l}=10\%$ and $\gamma_{l}=20\%$ in Table \ref{tab:all_aucroc_0.05}\textasciitilde\ref{tab:all_aucpr_0.2}. The best performing method(s) is marked in \textbf{bold}.

\begin{table*}[h!]
\footnotesize
\caption{Average AUC-ROC performance over \ndatasets real-world datasets. Each experiment is repeated 5 times. $\gamma_{l}$ stands for the ratio of labeled anomalies to all true anomalies in the training set. $\Delta$ Perf. shows the relative improvement of Overlap loss based models over their corresponding counterparts. $^{***}$, $^{**}$ and $^{*}$ denote statistical significance at $1\%$, $5\%$ and $10\%$ of Wilcoxon signed rank test, respectively. The best results are in \textbf{bold}.}
\label{tab:model_comparison_AUCROC}
  \centering
    \begin{tabular}{ccc|cc|cc|cc}
    \toprule
    \multirow{2}{*}{\textbf{Architecture}} & \multirow{2}{*}{\textbf{Model}} & \multirow{2}{*}{\textbf{Supervision}} & \multicolumn{2}{c}{$\gamma_{l}=5\%$} & \multicolumn{2}{c}{$\gamma_{l}=10\%$} & \multicolumn{2}{c}{$\gamma_{l}=20\%$} \\
\cmidrule{4-9}          &       &       & AUC-ROC & $\Delta$ Perf. & AUC-ROC & $\Delta$ Perf. & AUC-ROC & $\Delta$ Perf. \\
    \midrule
    \multirow{6}[4]{*}{\textbf{Typical}} & Iforest &Unsup       & 0.737±0.187 & /     & 0.737±0.187 & /     & 0.737±0.187 & / \\
          & ECOD  &Unsup       & 0.701±0.208 & /     & 0.701±0.208 & /     & 0.701±0.208 & / \\
          & DeepSVDD &Unsup       & 0.504±0.028 & /     & 0.504±0.028 & /     & 0.504±0.028 & / \\
\cmidrule{2-9}          & GANomaly &Semi       & 0.655±0.162 & /     & 0.648±0.153 & /     & 0.665±0.152 & / \\
          & DeepSAD &Semi       & 0.823±0.142 & /     & 0.859±0.136 & /     & 0.888±0.129 & / \\
          & REPEN &Weak       & 0.810±0.166 & /     & 0.832±0.165 & /     & 0.848±0.163 & / \\
    \midrule
    \multirow{3}[1]{*}{\textbf{MLP}} & DevNet &Weak       & 0.842±0.148 & $+0.57\%$ & 0.861±0.135 & $+2.16\%$ & 0.873±0.129 & $+2.40\%$ \\
          & PReNet &Weak       & 0.846±0.146 & $+0.18\%$ & 0.866±0.132 & $+1.61\%$ & 0.876±0.127 & $+1.97\%$ \\
          & MLP-Overlap (ours) &Weak       & \textbf{0.847±0.145} & /     & \textbf{0.880±0.132} & /     & \textbf{0.893±0.133} & / \\
    \midrule
    \multirow{3}[0]{*}{\textbf{AutoEncoder}} & FEAWAD & Sup      & 0.771±0.211 & $+11.72\%^{***}$ & 0.849±0.133 & $+4.34\%^{***}$ & 0.876±0.133 & $+2.42\%^{***}$ \\
    & FEAWAD & Weak      & 0.808±0.154 & $+6.60\%^{***}$ & 0.848±0.145 & $+4.51\%^{***}$ & 0.876±0.129 & $+2.43\%^{***}$ \\
          & AE-Overlap (ours) &Weak       & \textbf{0.862±0.144} & /     & \textbf{0.886±0.137} & /     & \textbf{0.897±0.132} & / \\
    \midrule
    \multirow{2}[0]{*}{\textbf{ResNet}} & ResNet & Sup      & 0.651±0.158 & $+28.48\%^{***}$ & 0.736±0.124 & $+19.74\%^{***}$ & 0.816±0.127 & $+11.03\%^{***}$ \\
          & ResNet-Overlap (ours) &Weak       & \textbf{0.836±0.146} & /     & \textbf{0.882±0.134} & /     & \textbf{0.906±0.122} & / \\
    \midrule
    \multirow{2}[1]{*}{\textbf{Transformer}} & FTTransformer & Sup      & 0.827±0.159 & $+3.00\%^{**}$ & 0.859±0.146 & $+1.80\%$ & 0.889±0.129 & $+1.23\%$ \\
          & FTTransformer-Overlap (ours) & Weak      & \textbf{0.851±0.138} & /     & \textbf{0.874±0.130} & /     & \textbf{0.900±0.127} & / \\
    \bottomrule
    \end{tabular}%
\end{table*}%

\begin{table*}[h!]
\footnotesize
  \centering
  \caption{AUC-ROC results of ablation studies. Overlap-Gaussian refers to the basic method mentioned in Section~\ref{subsec: score distribution estimator}. Overlap-Arbitrary refers to the basic method of Eq.\ref{Eq:overlap_wo_order}. Overlap-Ranking isolates the ranking loss in Eq.\ref{Eq:overlap_w_order}. Overlap-Combined corresponds to the combined loss form of both Overlap-Arbitrary and Overlap-Ranking as illustrated in Eq.\ref{Eq:overlap_w_order}. Overlap-Proposed refers to the final solution in this paper.}
  \label{tab:ablation_AUCROC}%
  \begin{tabular}{lccccc|ccccc|ccccc}
    \toprule
    \multicolumn{1}{c}{\multirow{2}{*}{\textbf{Method}}} & \multicolumn{5}{c}{$\gamma_{l}=5\%$}            & \multicolumn{5}{c}{$\gamma_{l}=10\%$}            & \multicolumn{5}{c}{$\gamma_{l}=20\%$} \\
\cmidrule{2-16}          & \textbf{VAE} & \textbf{MLP} & \textbf{AE} & \textbf{ResNet} & \textbf{FTT} & \textbf{VAE} & \textbf{MLP} & \textbf{AE} & \textbf{ResNet} & \textbf{FTT} & \textbf{VAE} & \textbf{MLP} & \textbf{AE} & \textbf{ResNet} & \textbf{FTT} \\
    
    \midrule
    Overlap-Gaussian & 0.539  & /     & /     & /     & /     & 0.540  & /     & /     & /     & /     & 0.541  & /     & /     & /     & / \\
    Overlap-Arbitrary & /     & 0.496  & 0.531  & 0.493  & 0.521  & /     & 0.498  & 0.543  & 0.534  & 0.482  & /     & 0.502  & 0.516  & 0.541  & 0.483  \\
    Overlap-Ranking & /     & 0.810  & 0.822  & 0.807  & \textbf{0.862}  & /     & 0.845  & 0.857  & 0.855  & \textbf{0.888}  & /     & 0.873  & 0.874  & 0.890  & \textbf{0.906}  \\
    Overlap-Combined & /     & 0.843  & 0.854  & 0.820  & 0.610  & /     & \textbf{0.881}  & 0.885  & 0.874  & 0.602  & /     & \textbf{0.898}  & \textbf{0.901}  & \textbf{0.908}  & 0.589  \\
    Overlap-Proposed & /     & \textbf{0.847}  & \textbf{0.862}  & \textbf{0.836}  & 0.851  & /     & 0.880  & \textbf{0.886}  & \textbf{0.882}  & 0.874  & /     & 0.893  & 0.897  & 0.906  & 0.900  \\
    \bottomrule
    \end{tabular}%
\end{table*}%

\clearpage
\begin{table*}[h]
\scriptsize
  \centering
  \caption{AUC-ROC results of model comparison on \ndatasets real-world datasets w.r.t. $\gamma_{l}=5\%$.}
    \begin{tabular}{lcccccc|ccc|ccc|cc|cc}
    \toprule
      \multicolumn{1}{l}{\multirow{2}{*}{\textbf{Dataset}}} & \multicolumn{6}{c}{\textbf{Typical}}                   & \multicolumn{3}{c}{\textbf{MLP}} & \multicolumn{3}{c}{\textbf{AutoEncoder}} & \multicolumn{2}{c}{\textbf{ResNet}} & \multicolumn{2}{c}{\textbf{Transformer}} \\
      \cmidrule{2-17}
       & Iforest & ECOD & \makecell{Deep\\SVDD} & \makecell{GAN\\omaly} & \makecell{Deep\\SAD} & REPEN & DevNet & PReNet & \makecell{MLP-\\Overlap} & \makecell{FEAWAD\\(Sup)} & \makecell{FEAWAD\\(Weak)} & \makecell{AE-\\Overlap} & ResNet & \makecell{ResNet-\\Overlap} & FTT & \makecell{FTT-\\Overlap} \\
      \midrule
    ALOI  & 0.548  & 0.560  & 0.525  & 0.552  & \textbf{0.576 } & 0.528  & 0.486  & 0.492  & 0.506  & 0.497  & 0.547  & 0.532  & 0.483  & 0.500  & 0.501  & 0.515  \\
    annthyroid & 0.828  & 0.783  & 0.502  & 0.769  & 0.818  & 0.803  & 0.805  & 0.822  & 0.882  & 0.798  & 0.903  & 0.914  & 0.761  & 0.872  & 0.922  & \textbf{0.985 } \\
    Cardiotocography & 0.692  & 0.684  & 0.514  & 0.575  & 0.807  & 0.898  & \textbf{0.911 } & 0.908  & 0.874  & 0.818  & 0.791  & 0.883  & 0.627  & 0.884  & 0.842  & 0.835  \\
    fault & 0.569  & 0.444  & 0.485  & 0.628  & \textbf{0.704 } & 0.694  & 0.696  & 0.683  & 0.645  & 0.671  & 0.657  & 0.668  & 0.502  & 0.665  & 0.657  & 0.656  \\
    http  & 0.999  & 0.981  & 0.509  & 0.980  & 0.998  & 0.999  & 0.999  & 0.999  & \textbf{1.000 } & 0.000  & 0.999  & 0.999  & 0.835  & \textbf{1.000 } & 0.999  & \textbf{1.000 } \\
    landsat & 0.483  & 0.571  & 0.501  & 0.514  & 0.854  & 0.584  & 0.776  & 0.779  & 0.833  & 0.767  & 0.802  & 0.816  & 0.671  & 0.735  & \textbf{0.864 } & \textbf{0.864 } \\
    letter & 0.635  & 0.526  & 0.468  & 0.673  & 0.703  & 0.629  & 0.590  & 0.582  & 0.591  & 0.557  & 0.562  & 0.566  & \textbf{0.752 } & 0.634  & 0.513  & 0.560  \\
    magic.gamma & 0.732  & 0.648  & 0.498  & 0.580  & 0.819  & 0.799  & 0.827  & 0.832  & 0.840  & 0.646  & 0.818  & 0.842  & 0.695  & 0.824  & 0.812  & \textbf{0.843 } \\
    mammography & 0.861  & 0.909  & 0.520  & 0.781  & 0.915  & 0.919  & 0.924  & 0.919  & 0.912  & 0.853  & 0.919  & \textbf{0.931 } & 0.748  & 0.867  & 0.900  & 0.921  \\
    mnist & 0.803  & 0.846  & 0.523  & 0.705  & 0.862  & 0.917  & \textbf{0.949 } & 0.925  & 0.823  & 0.935  & 0.869  & 0.925  & 0.586  & 0.720  & 0.916  & 0.870  \\
    musk  & 0.999  & 0.952  & 0.550  & 0.781  & 0.917  & 0.915  & \textbf{1.000 } & \textbf{1.000 } & \textbf{1.000 } & 0.998  & 0.927  & \textbf{1.000 } & 0.426  & \textbf{1.000 } & 0.999  & 0.989  \\
    optdigits & 0.674  & 0.612  & 0.522  & 0.384  & 0.934  & 0.986  & \textbf{1.000 } & 0.999  & 0.995  & 0.966  & 0.988  & 0.999  & 0.654  & 0.994  & 0.977  & 0.916  \\
    PageBlocks & 0.894  & 0.913  & 0.522  & 0.654  & \textbf{0.934 } & 0.912  & 0.864  & 0.883  & 0.896  & 0.785  & 0.895  & 0.890  & 0.714  & 0.910  & 0.842  & 0.888  \\
    pendigits & 0.955  & 0.910  & 0.485  & 0.707  & 0.965  & 0.997  & 0.996  & 0.993  & 0.995  & 0.958  & 0.997  & \textbf{0.999 } & 0.682  & 0.994  & 0.989  & 0.984  \\
    satellite & 0.699  & 0.750  & 0.503  & 0.722  & 0.883  & 0.807  & 0.853  & 0.852  & 0.852  & 0.766  & 0.834  & \textbf{0.910 } & 0.740  & 0.907  & 0.874  & 0.880  \\
    satimage-2 & \textbf{0.992 } & 0.966  & 0.525  & 0.969  & 0.981  & 0.986  & 0.991  & 0.989  & 0.970  & 0.921  & 0.967  & 0.988  & 0.367  & 0.973  & 0.942  & 0.932  \\
    shuttle & \textbf{0.996 } & 0.995  & 0.508  & 0.744  & 0.990  & 0.989  & 0.979  & 0.978  & 0.981  & 0.976  & 0.980  & 0.982  & 0.973  & 0.979  & 0.976  & 0.977  \\
    skin  & 0.684  & 0.391  & 0.500  & 0.542  & 0.995  & 0.919  & 0.951  & 0.954  & 0.982  & \textbf{0.999 } & 0.978  & 0.987  & 0.998  & 0.994  & 0.993  & 0.965  \\
    SpamBase & 0.633  & 0.660  & 0.500  & 0.534  & 0.690  & 0.838  & 0.902  & 0.909  & 0.876  & 0.748  & 0.768  & 0.841  & 0.598  & 0.769  & 0.870  & \textbf{0.917 } \\
    speech & 0.498  & 0.510  & 0.549  & 0.481  & 0.531  & 0.582  & 0.604  & 0.631  & 0.589  & 0.587  & 0.475  & 0.624  & 0.532  & 0.581  & 0.551  & \textbf{0.648 } \\
    thyroid & 0.981  & 0.979  & 0.521  & 0.919  & 0.941  & 0.990  & 0.994  & \textbf{0.995 } & 0.981  & 0.863  & 0.818  & 0.988  & 0.387  & 0.954  & 0.980  & 0.968  \\
    vowels & 0.765  & 0.440  & 0.407  & 0.792  & 0.767  & 0.812  & 0.847  & 0.891  & 0.860  & 0.777  & 0.665  & \textbf{0.916 } & 0.646  & 0.787  & 0.735  & 0.823  \\
    Waveform & 0.693  & 0.723  & 0.495  & 0.545  & 0.691  & 0.763  & 0.806  & 0.809  & 0.764  & \textbf{0.882 } & 0.626  & 0.780  & 0.563  & 0.765  & 0.772  & 0.814  \\
    Wilt  & 0.427  & 0.395  & 0.502  & 0.388  & 0.799  & 0.529  & 0.689  & 0.695  & 0.922  & 0.894  & 0.819  & 0.945  & 0.795  & \textbf{0.950 } & 0.658  & 0.910  \\
    yeast & 0.382  & 0.387  & 0.477  & 0.449  & 0.512  & 0.467  & 0.625  & 0.627  & 0.611  & 0.625  & 0.610  & 0.619  & 0.530  & \textbf{0.642 } & 0.583  & 0.627  \\
    \bottomrule
    \end{tabular}%
  \label{tab:all_aucroc_0.05}%
\end{table*}%
\begin{table*}[htbp]
\scriptsize
  \centering
  \caption{AUC-ROC results of model comparison on \ndatasets real-world datasets w.r.t. $\gamma_{l}=10\%$.}
    \begin{tabular}{lcccccc|ccc|ccc|cc|cc}
    \toprule
      \multicolumn{1}{l}{\multirow{2}{*}{\textbf{Dataset}}} & \multicolumn{6}{c}{\textbf{Typical}}                   & \multicolumn{3}{c}{\textbf{MLP}} & \multicolumn{3}{c}{\textbf{AutoEncoder}} & \multicolumn{2}{c}{\textbf{ResNet}} & \multicolumn{2}{c}{\textbf{Transformer}} \\
      \cmidrule{2-17}
       & Iforest & ECOD & \makecell{Deep\\SVDD} & \makecell{GAN\\omaly} & \makecell{Deep\\SAD} & REPEN & DevNet & PReNet & \makecell{MLP-\\Overlap} & \makecell{FEAWAD\\(Sup)} & \makecell{FEAWAD\\(Weak)} & \makecell{AE-\\Overlap} & ResNet & \makecell{ResNet-\\Overlap} & FTT & \makecell{FTT-\\Overlap} \\
      \midrule
    ALOI  & 0.548  & 0.560  & 0.525  & 0.556  & \textbf{0.574 } & 0.546  & 0.510  & 0.514  & 0.523  & 0.487  & 0.531  & 0.519  & 0.476  & 0.496  & 0.506  & 0.524  \\
    annthyroid & 0.828  & 0.783  & 0.502  & 0.733  & 0.884  & 0.824  & 0.826  & 0.834  & 0.939  & 0.880  & 0.897  & 0.968  & 0.905  & 0.932  & \textbf{0.990 } & \textbf{0.990 } \\
    Cardiotocography & 0.692  & 0.684  & 0.514  & 0.578  & 0.868  & 0.916  & \textbf{0.931 } & \textbf{0.931 } & 0.927  & 0.849  & 0.835  & 0.920  & 0.689  & 0.928  & 0.893  & 0.885  \\
    fault & 0.569  & 0.444  & 0.485  & 0.631  & \textbf{0.728 } & 0.720  & 0.724  & 0.719  & 0.695  & 0.692  & 0.674  & 0.695  & 0.570  & 0.721  & 0.694  & 0.699  \\
    http  & 0.999  & 0.981  & 0.509  & 0.785  & 0.999  & \textbf{1.000 } & 0.999  & \textbf{1.000 } & \textbf{1.000 } & \textbf{1.000 } & \textbf{1.000 } & \textbf{1.000 } & 0.829  & \textbf{1.000 } & \textbf{1.000 } & \textbf{1.000 } \\
    landsat & 0.483  & 0.571  & 0.501  & 0.522  & \textbf{0.897 } & 0.561  & 0.779  & 0.789  & 0.878  & 0.805  & 0.805  & 0.839  & 0.743  & 0.836  & 0.891  & 0.891  \\
    letter & 0.635  & 0.526  & 0.468  & 0.673  & 0.723  & \textbf{0.753 } & 0.699  & 0.713  & 0.694  & 0.699  & 0.613  & 0.656  & 0.749  & 0.720  & 0.612  & 0.636  \\
    magic.gamma & 0.732  & 0.648  & 0.498  & 0.611  & 0.847  & 0.809  & 0.827  & 0.833  & 0.870  & 0.688  & 0.841  & 0.874  & 0.755  & 0.871  & 0.834  & \textbf{0.879 } \\
    mammography & 0.861  & 0.909  & 0.520  & 0.774  & 0.907  & 0.925  & 0.926  & 0.924  & 0.931  & 0.793  & 0.918  & \textbf{0.935 } & 0.812  & 0.932  & 0.908  & 0.919  \\
    mnist & 0.803  & 0.846  & 0.523  & 0.707  & 0.916  & 0.966  & \textbf{0.974 } & 0.963  & 0.915  & 0.957  & 0.940  & 0.962  & 0.753  & 0.911  & 0.947  & 0.939  \\
    musk  & 0.999  & 0.952  & 0.550  & 0.879  & 0.968  & 0.971  & \textbf{1.000 } & \textbf{1.000 } & \textbf{1.000 } & 0.997  & \textbf{1.000 } & \textbf{1.000 } & 0.654  & \textbf{1.000 } & 0.996  & 0.995  \\
    optdigits & 0.674  & 0.612  & 0.522  & 0.383  & 0.979  & 0.993  & \textbf{1.000 } & 0.998  & 0.996  & 0.928  & 0.982  & 0.999  & 0.652  & \textbf{1.000 } & 0.985  & 0.903  \\
    PageBlocks & 0.894  & 0.913  & 0.522  & 0.681  & \textbf{0.945 } & 0.925  & 0.863  & 0.880  & 0.919  & 0.829  & 0.939  & 0.927  & 0.777  & 0.919  & 0.894  & 0.897  \\
    pendigits & 0.955  & 0.910  & 0.485  & 0.713  & 0.993  & 0.996  & 0.995  & 0.995  & 0.995  & 0.981  & 0.992  & \textbf{0.999 } & 0.761  & \textbf{0.999 } & 0.997  & 0.981  \\
    satellite & 0.699  & 0.750  & 0.503  & 0.726  & 0.909  & 0.807  & 0.851  & 0.849  & 0.899  & 0.840  & 0.852  & \textbf{0.919 } & 0.759  & 0.910  & 0.908  & 0.891  \\
    satimage-2 & \textbf{0.992 } & 0.966  & 0.525  & 0.969  & 0.986  & 0.989  & 0.988  & 0.986  & 0.954  & 0.959  & 0.971  & 0.978  & 0.801  & 0.987  & 0.974  & 0.949  \\
    shuttle & \textbf{0.996 } & 0.995  & 0.508  & 0.650  & 0.992  & 0.987  & 0.979  & 0.980  & 0.982  & 0.983  & 0.985  & 0.982  & 0.981  & 0.979  & 0.981  & 0.979  \\
    skin  & 0.684  & 0.391  & 0.500  & 0.516  & 0.997  & 0.910  & 0.951  & 0.954  & 0.992  & \textbf{0.999 } & 0.992  & 0.996  & 0.998  & 0.993  & 0.995  & 0.985  \\
    SpamBase & 0.633  & 0.660  & 0.500  & 0.538  & 0.801  & 0.861  & 0.915  & 0.928  & 0.921  & 0.812  & 0.862  & 0.916  & 0.677  & 0.893  & 0.914  & \textbf{0.951 } \\
    speech & 0.498  & 0.510  & 0.549  & 0.481  & 0.538  & 0.541  & 0.659  & 0.689  & 0.616  & 0.668  & 0.512  & 0.647  & 0.548  & 0.614  & 0.610  & \textbf{0.698 } \\
    thyroid & 0.981  & 0.979  & 0.521  & 0.920  & 0.965  & 0.994  & \textbf{0.996 } & \textbf{0.996 } & 0.994  & 0.974  & 0.897  & 0.988  & 0.698  & \textbf{0.996 } & 0.993  & 0.995  \\
    vowels & 0.765  & 0.440  & 0.407  & 0.794  & 0.805  & 0.896  & 0.926  & 0.942  & 0.925  & 0.906  & 0.899  & \textbf{0.964 } & 0.674  & 0.952  & 0.753  & 0.917  \\
    Waveform & 0.693  & 0.723  & 0.495  & 0.545  & 0.765  & 0.846  & 0.881  & 0.884  & 0.836  & \textbf{0.887 } & 0.728  & 0.854  & 0.666  & 0.786  & 0.863  & 0.774  \\
    Wilt  & 0.427  & 0.395  & 0.502  & 0.391  & 0.918  & 0.597  & 0.691  & 0.698  & 0.944  & 0.950  & 0.876  & 0.969  & 0.890  & \textbf{0.988 } & 0.685  & 0.942  \\
    yeast & 0.382  & 0.387  & 0.477  & 0.448  & 0.576  & 0.460  & 0.638  & 0.644  & 0.648  & 0.663  & 0.649  & 0.639  & 0.596  & \textbf{0.682 } & 0.655  & 0.642  \\
    \bottomrule
    \end{tabular}%
  \label{tab:addlabel}%
\end{table*}%

\begin{table*}[htbp]
\scriptsize
  \centering
  \caption{AUC-ROC results of model comparison on \ndatasets real-world datasets w.r.t. $\gamma_{l}=20\%$.}
    \begin{tabular}{lcccccc|ccc|ccc|cc|cc}
    \toprule
      \multicolumn{1}{l}{\multirow{2}{*}{\textbf{Dataset}}} & \multicolumn{6}{c}{\textbf{Typical}}                   & \multicolumn{3}{c}{\textbf{MLP}} & \multicolumn{3}{c}{\textbf{AutoEncoder}} & \multicolumn{2}{c}{\textbf{ResNet}} & \multicolumn{2}{c}{\textbf{Transformer}} \\
      \cmidrule{2-17}
       & Iforest & ECOD & \makecell{Deep\\SVDD} & \makecell{GAN\\omaly} & \makecell{Deep\\SAD} & REPEN & DevNet & PReNet & \makecell{MLP-\\Overlap} & \makecell{FEAWAD\\(Sup)} & \makecell{FEAWAD\\(Weak)} & \makecell{AE-\\Overlap} & ResNet & \makecell{ResNet-\\Overlap} & FTT & \makecell{FTT-\\Overlap} \\
      \midrule
    ALOI  & 0.548  & 0.560  & 0.525  & 0.550  & 0.591  & 0.532  & 0.522  & 0.520  & 0.525  & 0.504  & \textbf{0.596 } & 0.544  & 0.495  & 0.535  & 0.526  & 0.515  \\
    annthyroid & 0.828  & 0.783  & 0.502  & 0.781  & 0.930  & 0.825  & 0.825  & 0.835  & 0.951  & 0.955  & 0.941  & 0.970  & 0.957  & 0.958  & \textbf{0.992 } & 0.989  \\
    Cardiotocography & 0.692  & 0.684  & 0.514  & 0.583  & 0.913  & 0.930  & 0.946  & 0.945  & 0.942  & 0.896  & 0.894  & 0.946  & 0.778  & \textbf{0.948 } & 0.930  & 0.933  \\
    fault & 0.569  & 0.444  & 0.485  & 0.633  & \textbf{0.749 } & 0.743  & 0.733  & 0.738  & 0.698  & 0.669  & 0.718  & 0.673  & 0.642  & 0.745  & 0.713  & 0.717  \\
    http  & 0.999  & 0.981  & 0.509  & 0.783  & \textbf{1.000 } & \textbf{1.000 } & \textbf{1.000 } & \textbf{1.000 } & \textbf{1.000 } & \textbf{1.000 } & \textbf{1.000 } & \textbf{1.000 } & \textbf{1.000 } & \textbf{1.000 } & \textbf{1.000 } & \textbf{1.000 } \\
    landsat & 0.483  & 0.571  & 0.501  & 0.538  & \textbf{0.921 } & 0.560  & 0.789  & 0.782  & 0.899  & 0.872  & 0.846  & 0.806  & 0.806  & 0.884  & 0.913  & 0.901  \\
    letter & 0.635  & 0.526  & 0.468  & 0.675  & 0.751  & \textbf{0.806 } & 0.748  & 0.767  & 0.739  & 0.717  & 0.677  & 0.683  & 0.739  & 0.779  & 0.705  & 0.720  \\
    magic.gamma & 0.732  & 0.648  & 0.498  & 0.662  & 0.874  & 0.812  & 0.827  & 0.831  & 0.874  & 0.746  & 0.828  & 0.881  & 0.803  & \textbf{0.889 } & 0.858  & 0.860  \\
    mammography & 0.861  & 0.909  & 0.520  & 0.755  & 0.934  & 0.932  & 0.927  & 0.925  & 0.945  & 0.856  & 0.934  & \textbf{0.948 } & 0.864  & 0.946  & 0.923  & 0.934  \\
    mnist & 0.803  & 0.846  & 0.523  & 0.705  & 0.959  & \textbf{0.984 } & \textbf{0.984 } & 0.983  & 0.969  & 0.967  & 0.965  & 0.982  & 0.846  & 0.980  & 0.968  & 0.979  \\
    musk  & 0.999  & 0.952  & 0.550  & 0.891  & 0.996  & \textbf{1.000 } & \textbf{1.000 } & \textbf{1.000 } & \textbf{1.000 } & \textbf{1.000 } & \textbf{1.000 } & \textbf{1.000 } & 0.752  & \textbf{1.000 } & \textbf{1.000 } & \textbf{1.000 } \\
    optdigits & 0.674  & 0.612  & 0.522  & 0.385  & 0.997  & 0.996  & \textbf{1.000 } & \textbf{1.000 } & 0.998  & 0.987  & \textbf{1.000 } & 0.999  & 0.777  & \textbf{1.000 } & 0.998  & 0.999  \\
    PageBlocks & 0.894  & 0.913  & 0.522  & 0.775  & \textbf{0.959 } & 0.929  & 0.885  & 0.897  & 0.935  & 0.901  & 0.948  & 0.949  & 0.905  & 0.933  & 0.925  & 0.933  \\
    pendigits & 0.955  & 0.910  & 0.485  & 0.718  & 0.999  & 0.996  & 0.997  & 0.997  & 0.998  & 0.993  & 0.996  & \textbf{1.000 } & 0.923  & 0.999  & 0.997  & 0.993  \\
    satellite & 0.699  & 0.750  & 0.503  & 0.739  & 0.932  & 0.806  & 0.856  & 0.849  & 0.903  & 0.867  & 0.866  & 0.928  & 0.836  & \textbf{0.937 } & 0.935  & 0.914  \\
    satimage-2 & 0.992  & 0.966  & 0.525  & 0.971  & 0.992  & 0.992  & \textbf{0.993 } & \textbf{0.993 } & 0.969  & 0.959  & 0.977  & 0.986  & 0.912  & 0.988  & 0.969  & 0.974  \\
    shuttle & \textbf{0.996 } & 0.995  & 0.508  & 0.757  & 0.992  & 0.987  & 0.979  & 0.979  & 0.983  & 0.987  & 0.987  & 0.984  & 0.982  & 0.980  & 0.984  & 0.978  \\
    skin  & 0.684  & 0.391  & 0.500  & 0.499  & 0.998  & 0.903  & 0.951  & 0.955  & 0.995  & 0.999  & 0.983  & 0.998  & 0.999  & \textbf{1.000 } & 0.996  & 0.976  \\
    SpamBase & 0.633  & 0.660  & 0.500  & 0.544  & 0.887  & 0.889  & 0.913  & 0.931  & 0.938  & 0.868  & 0.886  & 0.946  & 0.807  & 0.945  & 0.928  & \textbf{0.953 } \\
    speech & 0.498  & 0.510  & 0.549  & 0.482  & 0.559  & 0.611  & 0.713  & \textbf{0.740 } & 0.624  & 0.643  & 0.598  & 0.694  & 0.584  & 0.685  & 0.678  & 0.734  \\
    thyroid & 0.981  & 0.979  & 0.521  & 0.918  & 0.986  & 0.996  & \textbf{0.997 } & \textbf{0.997 } & 0.994  & 0.983  & 0.996  & 0.996  & 0.879  & \textbf{0.997 } & 0.993  & \textbf{0.997 } \\
    vowels & 0.765  & 0.440  & 0.407  & 0.798  & 0.871  & 0.971  & 0.970  & 0.979  & 0.977  & 0.953  & 0.961  & 0.995  & 0.837  & \textbf{0.996 } & 0.976  & 0.991  \\
    Waveform & 0.693  & 0.723  & 0.495  & 0.546  & 0.811  & 0.898  & \textbf{0.909 } & 0.903  & 0.880  & 0.899  & 0.765  & 0.885  & 0.700  & 0.839  & 0.885  & 0.885  \\
    Wilt  & 0.427  & 0.395  & 0.502  & 0.479  & 0.960  & 0.646  & 0.689  & 0.696  & 0.954  & 0.981  & 0.876  & 0.970  & 0.922  & \textbf{0.991 } & 0.732  & 0.976  \\
    yeast & 0.382  & 0.387  & 0.477  & 0.458  & 0.644  & 0.459  & 0.663  & 0.666  & 0.646  & 0.703  & 0.663  & 0.672  & 0.660  & 0.699  & \textbf{0.705 } & 0.650  \\
    \bottomrule
    \end{tabular}%
  \label{tab:addlabel}%
\end{table*}%

\begin{table*}[htbp]
\scriptsize
  \centering
  \caption{AUC-PR results of model comparison on \ndatasets real-world datasets w.r.t. $\gamma_{l}=5\%$.}
    \begin{tabular}{lcccccc|ccc|ccc|cc|cc}
    \toprule
      \multicolumn{1}{l}{\multirow{2}{*}{\textbf{Dataset}}} & \multicolumn{6}{c}{\textbf{Typical}}                   & \multicolumn{3}{c}{\textbf{MLP}} & \multicolumn{3}{c}{\textbf{AutoEncoder}} & \multicolumn{2}{c}{\textbf{ResNet}} & \multicolumn{2}{c}{\textbf{Transformer}} \\
      \cmidrule{2-17}
       & Iforest & ECOD & \makecell{Deep\\SVDD} & \makecell{GAN\\omaly} & \makecell{Deep\\SAD} & REPEN & DevNet & PReNet & \makecell{MLP-\\Overlap} & \makecell{FEAWAD\\(Sup)} & \makecell{FEAWAD\\(Weak)} & \makecell{AE-\\Overlap} & ResNet & \makecell{ResNet-\\Overlap} & FTT & \makecell{FTT-\\Overlap} \\
      \midrule
    ALOI  & 0.036  & 0.036  & 0.042  & 0.038  & \textbf{0.058 } & 0.041  & 0.042  & 0.045  & 0.046  & 0.043  & 0.040  & 0.048  & 0.037  & 0.040  & 0.035  & 0.036  \\
    annthyroid & 0.336  & 0.260  & 0.078  & 0.359  & 0.402  & 0.400  & 0.425  & 0.458  & 0.541  & 0.482  & 0.609  & 0.602  & 0.363  & 0.575  & 0.658  & \textbf{0.799 } \\
    Cardiotocography & 0.441  & 0.436  & 0.254  & 0.345  & 0.602  & 0.760  & 0.767  & \textbf{0.776 } & 0.708  & 0.610  & 0.652  & 0.720  & 0.433  & 0.736  & 0.649  & 0.691  \\
    fault & 0.418  & 0.317  & 0.355  & 0.477  & 0.531  & 0.543  & 0.557  & \textbf{0.565 } & 0.514  & 0.501  & 0.527  & 0.525  & 0.425  & 0.521  & 0.495  & 0.538  \\
    http  & 0.808  & 0.158  & 0.022  & 0.648  & 0.659  & 0.833  & 0.869  & 0.845  & 0.982  & 0.003  & 0.847  & 0.898  & 0.805  & \textbf{1.000 } & 0.867  & \textbf{1.000 } \\
    landsat & 0.199  & 0.262  & 0.210  & 0.210  & 0.617  & 0.322  & 0.448  & 0.447  & 0.572  & 0.502  & 0.575  & 0.537  & 0.407  & 0.497  & \textbf{0.682 } & 0.633  \\
    letter & 0.098  & 0.076  & 0.165  & 0.141  & 0.176  & 0.148  & 0.158  & 0.159  & 0.201  & 0.134  & 0.109  & 0.137  & \textbf{0.342 } & 0.169  & 0.118  & 0.126  \\
    magic.gamma & 0.648  & 0.551  & 0.357  & 0.449  & 0.709  & 0.730  & 0.725  & 0.755  & 0.771  & 0.535  & 0.732  & 0.774  & 0.548  & 0.750  & 0.703  & \textbf{0.792 } \\
    mammography & 0.195  & 0.414  & 0.055  & 0.127  & 0.510  & 0.592  & \textbf{0.610 } & 0.603  & 0.427  & 0.418  & 0.545  & 0.546  & 0.372  & 0.469  & 0.528  & 0.589  \\
    mnist & 0.286  & 0.329  & 0.127  & 0.196  & 0.544  & 0.717  & \textbf{0.784 } & 0.738  & 0.617  & 0.670  & 0.643  & 0.760  & 0.286  & 0.496  & 0.644  & 0.554  \\
    musk  & 0.970  & 0.343  & 0.105  & 0.464  & 0.597  & 0.653  & \textbf{1.000 } & \textbf{1.000 } & \textbf{1.000 } & 0.961  & 0.895  & \textbf{1.000 } & 0.246  & \textbf{1.000 } & 0.974  & 0.951  \\
    optdigits & 0.047  & 0.035  & 0.032  & 0.028  & 0.510  & 0.951  & \textbf{0.991 } & 0.989  & 0.966  & 0.708  & 0.961  & 0.987  & 0.260  & 0.960  & 0.811  & 0.625  \\
    PageBlocks & 0.469  & 0.517  & 0.146  & 0.346  & 0.714  & 0.678  & 0.654  & 0.679  & 0.642  & 0.539  & \textbf{0.719 } & 0.680  & 0.456  & 0.696  & 0.602  & 0.638  \\
    pendigits & 0.283  & 0.216  & 0.032  & 0.185  & 0.716  & 0.955  & 0.928  & 0.922  & 0.952  & 0.760  & 0.954  & \textbf{0.980 } & 0.367  & 0.970  & 0.890  & 0.908  \\
    satellite & 0.662  & 0.662  & 0.323  & 0.666  & 0.782  & 0.795  & 0.828  & 0.825  & 0.768  & 0.663  & 0.763  & 0.861  & 0.628  & \textbf{0.869 } & 0.805  & 0.838  \\
    satimage-2 & 0.913  & 0.629  & 0.034  & 0.523  & 0.671  & 0.899  & 0.908  & 0.912  & 0.918  & 0.560  & 0.912  & \textbf{0.930 } & 0.161  & 0.880  & 0.885  & 0.790  \\
    shuttle & \textbf{0.975 } & 0.957  & 0.081  & 0.464  & 0.954  & 0.974  & 0.968  & 0.967  & 0.962  & 0.956  & 0.967  & 0.966  & 0.959  & 0.966  & 0.953  & 0.960  \\
    skin  & 0.257  & 0.156  & 0.204  & 0.223  & 0.950  & 0.555  & 0.660  & 0.673  & 0.866  & \textbf{0.994 } & 0.832  & 0.905  & 0.984  & 0.950  & 0.963  & 0.793  \\
    SpamBase & 0.496  & 0.528  & 0.400  & 0.409  & 0.584  & 0.774  & 0.846  & 0.863  & 0.836  & 0.657  & 0.754  & 0.806  & 0.550  & 0.748  & 0.813  & \textbf{0.889 } \\
    speech & 0.021  & 0.019  & 0.049  & 0.017  & 0.024  & 0.045  & 0.052  & 0.057  & 0.042  & 0.044  & 0.027  & \textbf{0.060 } & 0.040  & 0.046  & 0.037  & 0.054  \\
    thyroid & 0.574  & 0.526  & 0.068  & 0.466  & 0.464  & 0.753  & 0.880  & \textbf{0.888 } & 0.814  & 0.487  & 0.581  & 0.852  & 0.108  & 0.777  & 0.737  & 0.814  \\
    vowels & 0.193  & 0.039  & 0.105  & 0.244  & 0.150  & 0.345  & 0.384  & 0.431  & 0.438  & 0.289  & 0.283  & \textbf{0.602 } & 0.336  & 0.354  & 0.336  & 0.386  \\
    Waveform & 0.063  & 0.064  & 0.032  & 0.043  & 0.180  & 0.121  & 0.135  & 0.177  & 0.182  & \textbf{0.216 } & 0.156  & 0.181  & 0.141  & 0.154  & 0.182  & 0.190  \\
    Wilt  & 0.043  & 0.042  & 0.054  & 0.045  & 0.188  & 0.065  & 0.087  & 0.089  & 0.381  & 0.545  & 0.390  & 0.494  & 0.380  & 0.582  & 0.082  & \textbf{0.633 } \\
    yeast & 0.293  & 0.305  & 0.333  & 0.314  & 0.348  & 0.338  & 0.437  & 0.439  & 0.436  & 0.445  & 0.429  & 0.438  & 0.391  & \textbf{0.467 } & 0.409  & 0.447  \\
    \bottomrule
    \end{tabular}%
  \label{tab:addlabel}%
\end{table*}%

\begin{table*}[htbp]
\scriptsize
  \centering
  \caption{AUC-PR results of model comparison on \ndatasets real-world datasets w.r.t. $\gamma_{l}=10\%$.}
    \begin{tabular}{lcccccc|ccc|ccc|cc|cc}
    \toprule
      \multicolumn{1}{l}{\multirow{2}{*}{\textbf{Dataset}}} & \multicolumn{6}{c}{\textbf{Typical}}                   & \multicolumn{3}{c}{\textbf{MLP}} & \multicolumn{3}{c}{\textbf{AutoEncoder}} & \multicolumn{2}{c}{\textbf{ResNet}} & \multicolumn{2}{c}{\textbf{Transformer}} \\
      \cmidrule{2-17}
       & Iforest & ECOD & \makecell{Deep\\SVDD} & \makecell{GAN\\omaly} & \makecell{Deep\\SAD} & REPEN & DevNet & PReNet & \makecell{MLP-\\Overlap} & \makecell{FEAWAD\\(Sup)} & \makecell{FEAWAD\\(Weak)} & \makecell{AE-\\Overlap} & ResNet & \makecell{ResNet-\\Overlap} & FTT & \makecell{FTT-\\Overlap} \\
      \midrule
    ALOI  & 0.036  & 0.036  & 0.042  & 0.039  & \textbf{0.068 } & 0.052  & 0.047  & 0.052  & 0.049  & 0.044  & 0.041  & 0.046  & 0.036  & 0.037  & 0.037  & 0.042  \\
    annthyroid & 0.336  & 0.260  & 0.078  & 0.327  & 0.506  & 0.437  & 0.459  & 0.476  & 0.660  & 0.615  & 0.574  & 0.729  & 0.591  & 0.674  & \textbf{0.835 } & 0.833  \\
    Cardiotocography & 0.441  & 0.436  & 0.254  & 0.347  & 0.708  & 0.799  & 0.805  & \textbf{0.817 } & 0.784  & 0.676  & 0.710  & 0.779  & 0.501  & 0.799  & 0.695  & 0.755  \\
    fault & 0.418  & 0.317  & 0.355  & 0.480  & 0.560  & 0.577  & 0.586  & \textbf{0.596 } & 0.542  & 0.521  & 0.543  & 0.537  & 0.490  & 0.575  & 0.538  & 0.567  \\
    http  & 0.808  & 0.158  & 0.022  & 0.451  & 0.829  & 0.928  & 0.891  & \textbf{1.000 } & \textbf{1.000 } & \textbf{1.000 } & \textbf{1.000 } & \textbf{1.000 } & 0.801  & \textbf{1.000 } & \textbf{1.000 } & \textbf{1.000 } \\
    landsat & 0.199  & 0.262  & 0.210  & 0.215  & 0.704  & 0.357  & 0.454  & 0.486  & 0.632  & 0.568  & 0.516  & 0.597  & 0.474  & 0.614  & \textbf{0.719 } & 0.708  \\
    letter & 0.098  & 0.076  & 0.165  & 0.142  & 0.183  & 0.209  & 0.153  & 0.170  & 0.225  & 0.167  & 0.140  & 0.144  & \textbf{0.305 } & 0.190  & 0.139  & 0.142  \\
    magic.gamma & 0.648  & 0.551  & 0.357  & 0.473  & 0.754  & 0.748  & 0.718  & 0.748  & 0.817  & 0.594  & 0.758  & 0.822  & 0.622  & 0.819  & 0.735  & \textbf{0.839 } \\
    mammography & 0.195  & 0.414  & 0.055  & 0.127  & 0.562  & 0.606  & \textbf{0.621 } & 0.606  & 0.515  & 0.370  & 0.520  & 0.566  & 0.475  & 0.573  & 0.565  & 0.544  \\
    mnist & 0.286  & 0.329  & 0.127  & 0.198  & 0.678  & \textbf{0.847 } & 0.846  & 0.820  & 0.752  & 0.737  & 0.737  & 0.844  & 0.443  & 0.781  & 0.724  & 0.799  \\
    musk  & 0.970  & 0.343  & 0.105  & 0.733  & 0.835  & 0.886  & \textbf{1.000 } & \textbf{1.000 } & \textbf{1.000 } & 0.972  & \textbf{1.000 } & \textbf{1.000 } & 0.405  & \textbf{1.000 } & 0.983  & 0.988  \\
    optdigits & 0.047  & 0.035  & 0.032  & 0.027  & 0.811  & 0.973  & 0.994  & 0.991  & 0.979  & 0.713  & 0.961  & 0.992  & 0.323  & \textbf{0.995 } & 0.885  & 0.746  \\
    PageBlocks & 0.469  & 0.517  & 0.146  & 0.349  & 0.747  & 0.719  & 0.645  & 0.662  & 0.666  & 0.616  & \textbf{0.774 } & 0.695  & 0.526  & 0.695  & 0.689  & 0.711  \\
    pendigits & 0.283  & 0.216  & 0.032  & 0.186  & 0.893  & 0.963  & 0.914  & 0.914  & 0.955  & 0.881  & 0.946  & \textbf{0.985 } & 0.427  & 0.980  & 0.946  & 0.865  \\
    satellite & 0.662  & 0.662  & 0.323  & 0.653  & 0.829  & 0.789  & 0.830  & 0.828  & 0.846  & 0.749  & 0.772  & \textbf{0.877 } & 0.636  & 0.872  & 0.829  & 0.851  \\
    satimage-2 & 0.913  & 0.629  & 0.034  & 0.527  & 0.868  & 0.910  & 0.897  & 0.886  & 0.911  & 0.851  & 0.907  & \textbf{0.919 } & 0.525  & 0.916  & 0.873  & 0.877  \\
    shuttle & \textbf{0.975 } & 0.957  & 0.081  & 0.363  & 0.964  & 0.971  & 0.968  & 0.967  & 0.966  & 0.971  & 0.970  & 0.971  & 0.967  & 0.968  & 0.961  & 0.965  \\
    skin  & 0.257  & 0.156  & 0.204  & 0.212  & 0.971  & 0.532  & 0.658  & 0.675  & 0.937  & \textbf{0.989 } & 0.945  & 0.959  & 0.982  & 0.944  & 0.970  & 0.904  \\
    SpamBase & 0.496  & 0.528  & 0.400  & 0.411  & 0.701  & 0.802  & 0.857  & 0.879  & 0.884  & 0.745  & 0.834  & 0.886  & 0.635  & 0.868  & 0.862  & \textbf{0.934 } \\
    speech & 0.021  & 0.019  & 0.049  & 0.017  & 0.025  & 0.056  & 0.065  & 0.068  & \textbf{0.085 } & 0.066  & 0.044  & 0.065  & 0.064  & 0.067  & 0.061  & 0.072  \\
    thyroid & 0.574  & 0.526  & 0.068  & 0.475  & 0.606  & 0.870  & \textbf{0.903 } & 0.883  & 0.870  & 0.750  & 0.810  & 0.882  & 0.368  & 0.895  & 0.840  & 0.893  \\
    vowels & 0.193  & 0.039  & 0.105  & 0.247  & 0.192  & 0.477  & 0.646  & 0.705  & 0.676  & 0.506  & 0.501  & \textbf{0.770 } & 0.410  & 0.676  & 0.418  & 0.711  \\
    Waveform & 0.063  & 0.064  & 0.032  & 0.043  & \textbf{0.263 } & 0.146  & 0.160  & 0.189  & 0.221  & 0.245  & 0.195  & 0.248  & 0.137  & 0.218  & 0.232  & 0.220  \\
    Wilt  & 0.043  & 0.042  & 0.054  & 0.046  & 0.387  & 0.077  & 0.087  & 0.089  & 0.436  & 0.663  & 0.462  & 0.611  & 0.492  & \textbf{0.831 } & 0.090  & 0.752  \\
    yeast & 0.293  & 0.305  & 0.333  & 0.313  & 0.392  & 0.349  & 0.434  & 0.440  & 0.443  & \textbf{0.489 } & 0.462  & 0.439  & 0.429  & 0.486  & 0.472  & 0.443  \\
    \bottomrule
    \end{tabular}%
  \label{tab:addlabel}%
\end{table*}%

\begin{table*}[htbp]
\scriptsize
  \centering
  \caption{AUC-PR results of model comparison on \ndatasets real-world datasets w.r.t. $\gamma_{l}=20\%$.}
    \begin{tabular}{lcccccc|ccc|ccc|cc|cc}
    \toprule
      \multicolumn{1}{l}{\multirow{2}{*}{\textbf{Dataset}}} & \multicolumn{6}{c}{\textbf{Typical}}                   & \multicolumn{3}{c}{\textbf{MLP}} & \multicolumn{3}{c}{\textbf{AutoEncoder}} & \multicolumn{2}{c}{\textbf{ResNet}} & \multicolumn{2}{c}{\textbf{Transformer}} \\
      \cmidrule{2-17}
       & Iforest & ECOD & \makecell{Deep\\SVDD} & \makecell{GAN\\omaly} & \makecell{Deep\\SAD} & REPEN & DevNet & PReNet & \makecell{MLP-\\Overlap} & \makecell{FEAWAD\\(Sup)} & \makecell{FEAWAD\\(Weak)} & \makecell{AE-\\Overlap} & ResNet & \makecell{ResNet-\\Overlap} & FTT & \makecell{FTT-\\Overlap} \\
      \midrule
    ALOI  & 0.036  & 0.036  & 0.042  & 0.038  & \textbf{0.069 } & 0.046  & 0.047  & 0.045  & 0.045  & 0.051  & 0.047  & 0.044  & 0.040  & 0.052  & 0.041  & 0.039  \\
    annthyroid & 0.336  & 0.260  & 0.078  & 0.372  & 0.624  & 0.441  & 0.457  & 0.478  & 0.677  & 0.703  & 0.622  & 0.739  & 0.704  & 0.748  & \textbf{0.858 } & 0.818  \\
    Cardiotocography & 0.441  & 0.436  & 0.254  & 0.354  & 0.793  & 0.830  & \textbf{0.838 } & 0.837  & 0.826  & 0.743  & 0.768  & 0.835  & 0.600  & \textbf{0.838 } & 0.794  & 0.828  \\
    fault & 0.418  & 0.317  & 0.355  & 0.485  & 0.582  & \textbf{0.610 } & 0.581  & 0.604  & 0.510  & 0.517  & 0.570  & 0.520  & 0.536  & 0.592  & 0.548  & 0.565  \\
    http  & 0.808  & 0.158  & 0.022  & 0.451  & 0.891  & 0.928  & \textbf{1.000 } & \textbf{1.000 } & \textbf{1.000 } & \textbf{1.000 } & \textbf{1.000 } & \textbf{1.000 } & \textbf{1.000 } & \textbf{1.000 } & \textbf{1.000 } & \textbf{1.000 } \\
    landsat & 0.199  & 0.262  & 0.210  & 0.228  & \textbf{0.763 } & 0.389  & 0.528  & 0.487  & 0.688  & 0.695  & 0.563  & 0.523  & 0.587  & 0.680  & 0.753  & 0.745  \\
    letter & 0.098  & 0.076  & 0.165  & 0.143  & 0.197  & 0.276  & 0.216  & 0.260  & 0.239  & 0.206  & 0.195  & 0.175  & \textbf{0.344 } & 0.268  & 0.197  & 0.211  \\
    magic.gamma & 0.648  & 0.551  & 0.357  & 0.518  & 0.799  & 0.746  & 0.720  & 0.743  & 0.817  & 0.674  & 0.726  & 0.826  & 0.684  & \textbf{0.850 } & 0.767  & 0.819  \\
    mammography & 0.195  & 0.414  & 0.055  & 0.119  & 0.600  & 0.617  & 0.614  & 0.613  & 0.513  & 0.477  & 0.618  & 0.600  & 0.510  & 0.602  & 0.601  & \textbf{0.629 } \\
    mnist & 0.286  & 0.329  & 0.127  & 0.195  & 0.810  & \textbf{0.908 } & 0.871  & 0.888  & 0.850  & 0.808  & 0.786  & 0.883  & 0.598  & 0.902  & 0.798  & 0.872  \\
    musk  & 0.970  & 0.343  & 0.105  & 0.752  & 0.981  & 0.994  & \textbf{1.000 } & \textbf{1.000 } & \textbf{1.000 } & 0.999  & \textbf{1.000 } & \textbf{1.000 } & 0.592  & \textbf{1.000 } & \textbf{1.000 } & 0.999  \\
    optdigits & 0.047  & 0.035  & 0.032  & 0.028  & 0.961  & 0.989  & 0.996  & 0.994  & 0.987  & 0.887  & 0.993  & 0.992  & 0.517  & \textbf{0.999 } & 0.967  & 0.982  \\
    PageBlocks & 0.469  & 0.517  & 0.146  & 0.395  & \textbf{0.787 } & 0.724  & 0.672  & 0.692  & 0.696  & 0.725  & 0.785  & 0.752  & 0.720  & 0.710  & 0.708  & 0.738  \\
    pendigits & 0.283  & 0.216  & 0.032  & 0.188  & 0.984  & 0.973  & 0.933  & 0.940  & 0.982  & 0.919  & 0.965  & \textbf{0.990 } & 0.733  & 0.986  & 0.969  & 0.958  \\
    satellite & 0.662  & 0.662  & 0.323  & 0.646  & 0.874  & 0.790  & 0.832  & 0.829  & 0.854  & 0.794  & 0.801  & 0.888  & 0.741  & \textbf{0.903 } & 0.870  & 0.882  \\
    satimage-2 & 0.913  & 0.629  & 0.034  & 0.535  & 0.906  & 0.923  & \textbf{0.928 } & 0.919  & 0.926  & 0.877  & 0.926  & 0.920  & 0.752  & 0.927  & 0.907  & 0.911  \\
    shuttle & \textbf{0.975 } & 0.957  & 0.081  & 0.415  & 0.967  & 0.972  & 0.967  & 0.967  & 0.966  & 0.973  & 0.971  & 0.972  & 0.968  & 0.969  & 0.964  & 0.969  \\
    skin  & 0.257  & 0.156  & 0.204  & 0.202  & 0.981  & 0.513  & 0.658  & 0.679  & 0.954  & 0.988  & 0.871  & 0.988  & 0.990  & \textbf{0.998 } & 0.974  & 0.857  \\
    SpamBase & 0.496  & 0.528  & 0.400  & 0.416  & 0.811  & 0.838  & 0.858  & 0.885  & 0.905  & 0.814  & 0.854  & 0.923  & 0.759  & 0.928  & 0.879  & \textbf{0.938 } \\
    speech & 0.021  & 0.019  & 0.049  & 0.017  & 0.027  & \textbf{0.160 } & 0.101  & 0.117  & 0.084  & 0.075  & 0.074  & 0.070  & 0.119  & 0.108  & 0.063  & 0.104  \\
    thyroid & 0.574  & 0.526  & 0.068  & 0.475  & 0.785  & 0.866  & 0.911  & 0.898  & 0.859  & 0.751  & 0.871  & 0.897  & 0.574  & 0.891  & \textbf{0.916 } & 0.914  \\
    vowels & 0.193  & 0.039  & 0.105  & 0.253  & 0.341  & 0.833  & 0.807  & 0.838  & 0.825  & 0.654  & 0.803  & 0.921  & 0.621  & 0.925  & 0.757  & \textbf{0.941 } \\
    Waveform & 0.063  & 0.064  & 0.032  & 0.043  & \textbf{0.335 } & 0.185  & 0.217  & 0.227  & 0.282  & 0.286  & 0.224  & 0.310  & 0.192  & 0.282  & 0.328  & 0.265  \\
    Wilt  & 0.043  & 0.042  & 0.054  & 0.053  & 0.573  & 0.087  & 0.086  & 0.088  & 0.476  & 0.799  & 0.537  & 0.579  & 0.590  & \textbf{0.874 } & 0.110  & 0.822  \\
    yeast & 0.293  & 0.305  & 0.333  & 0.318  & 0.445  & 0.349  & 0.462  & 0.467  & 0.438  & \textbf{0.535 } & 0.477  & 0.476  & 0.493  & 0.509  & 0.514  & 0.453  \\
    \bottomrule
    \end{tabular}%
  \label{tab:all_aucpr_0.2}%
\end{table*}%

\clearpage
\subsection{Additional Results of AD Loss Function Exploration}
\label{subsec: appendix_further_explore}
\begin{figure*}[ht!]
     \centering
     \begin{minipage}[c]{0.12\textwidth}
        \begin{subfigure}[c]{0.96\textwidth}
            \centering
            \includegraphics[width=\textwidth]{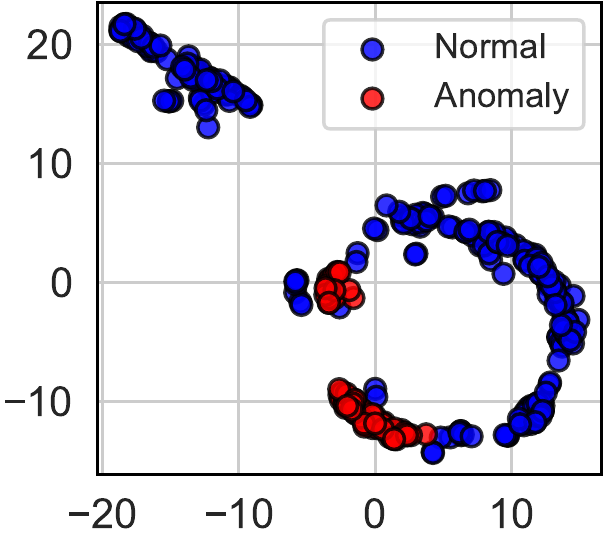}
            \caption{t-SNE \cite{TSNE} plots of the input feature of skin dataset.}
            \label{fig:Embedding_input}
        \end{subfigure}
     \end{minipage}
     \hspace{0.1in}
     \begin{minipage}[c]{0.85\textwidth}
     \begin{subfigure}[b]{0.3\textwidth}
         \centering
         \includegraphics[width=\textwidth]{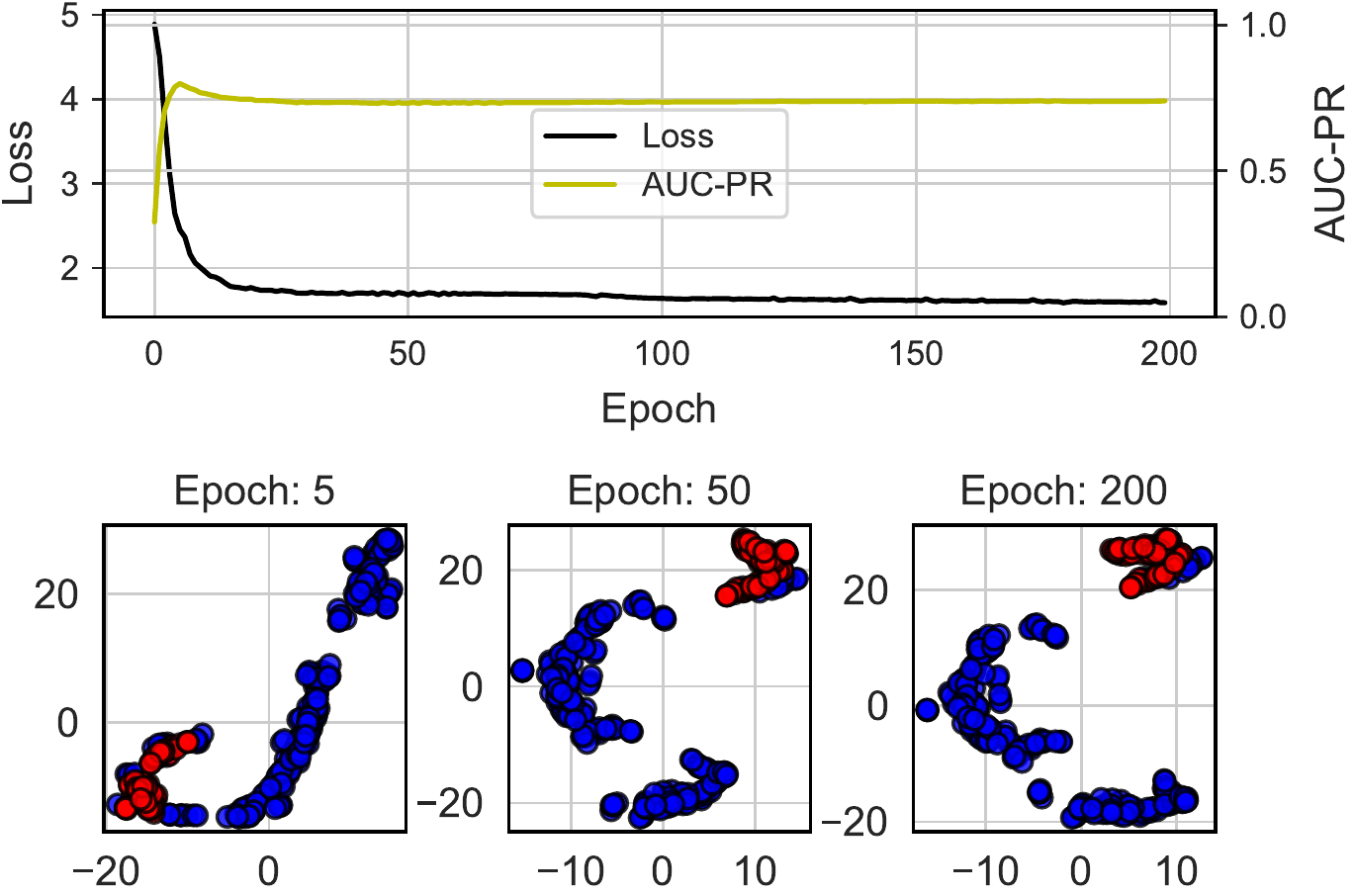}
         \caption{Minus}
         \label{fig:}
     \end{subfigure}
     \hspace{0.2em}
     \begin{subfigure}[b]{0.3\textwidth}
         \centering
         \includegraphics[width=\textwidth]{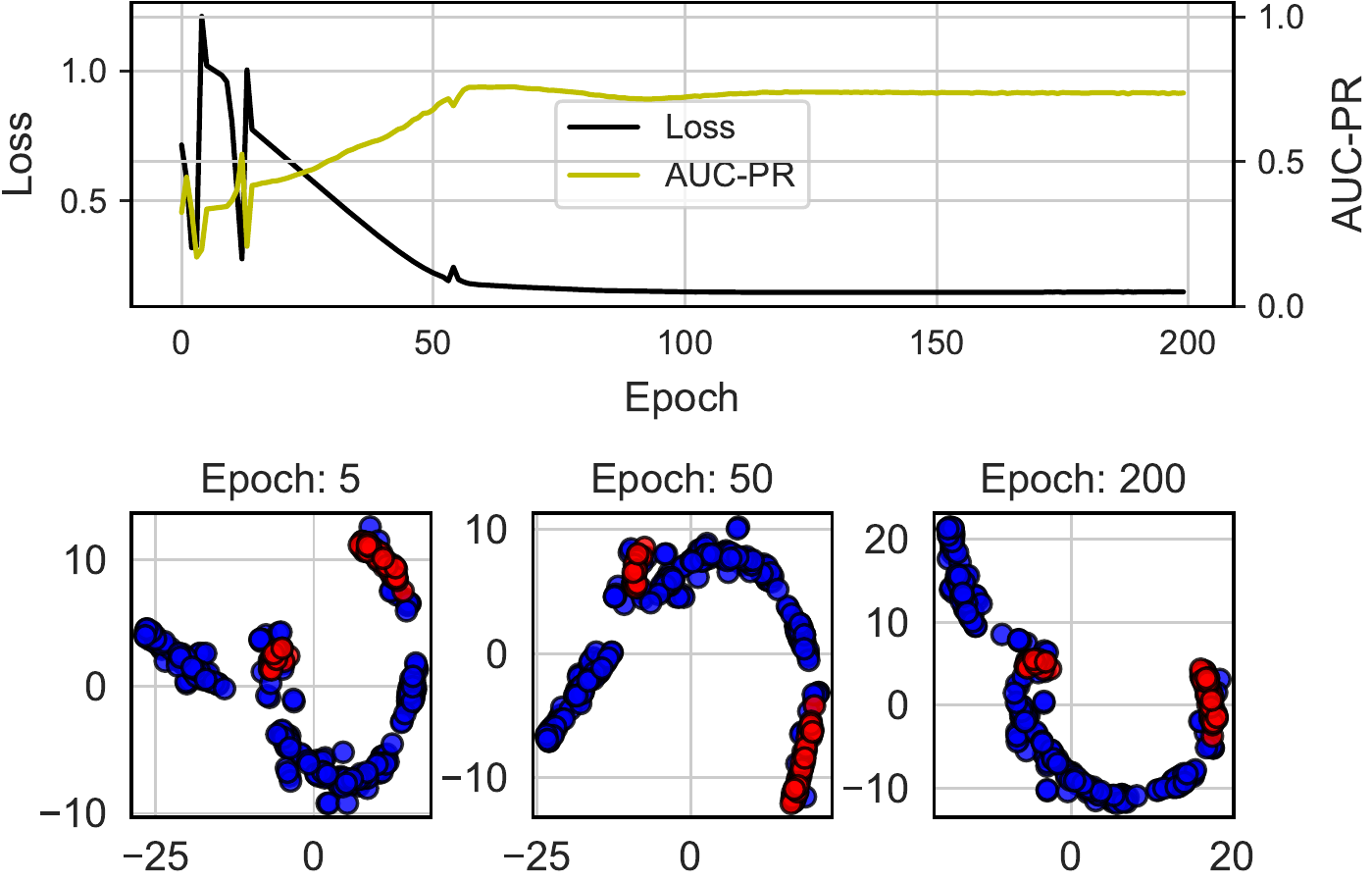}
         \caption{Inverse}
         \label{fig:}
     \end{subfigure}
     \hspace{0.2em}
     \begin{subfigure}[b]{0.3\textwidth}
         \centering
         \includegraphics[width=\textwidth]{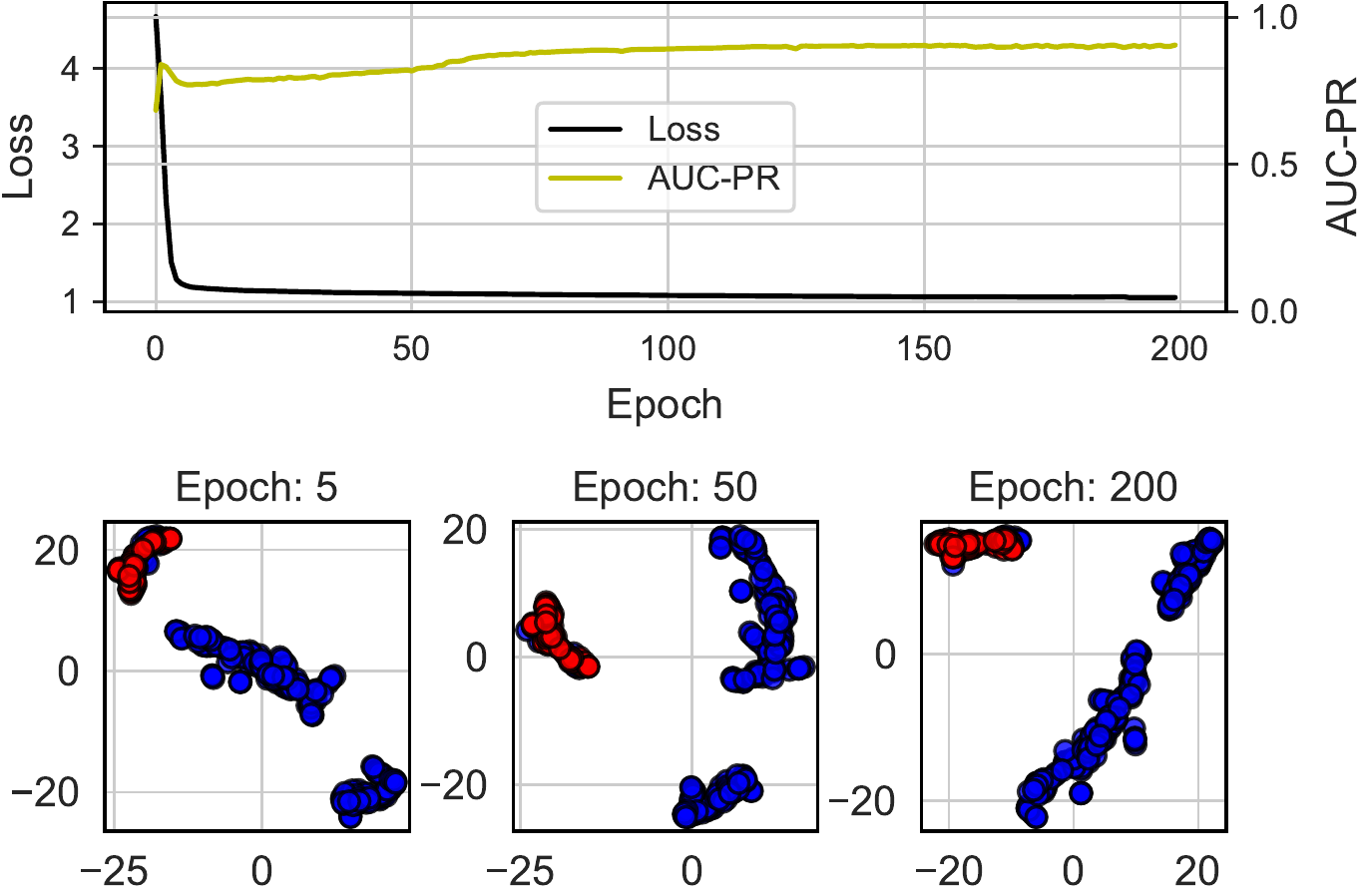}
         \caption{Hinge}
         \label{fig:}
     \end{subfigure}
     \hspace{0.2em}
     
     \vspace{0.5em}
     \begin{subfigure}[b]{0.3\textwidth}
         \centering
         \includegraphics[width=\textwidth]{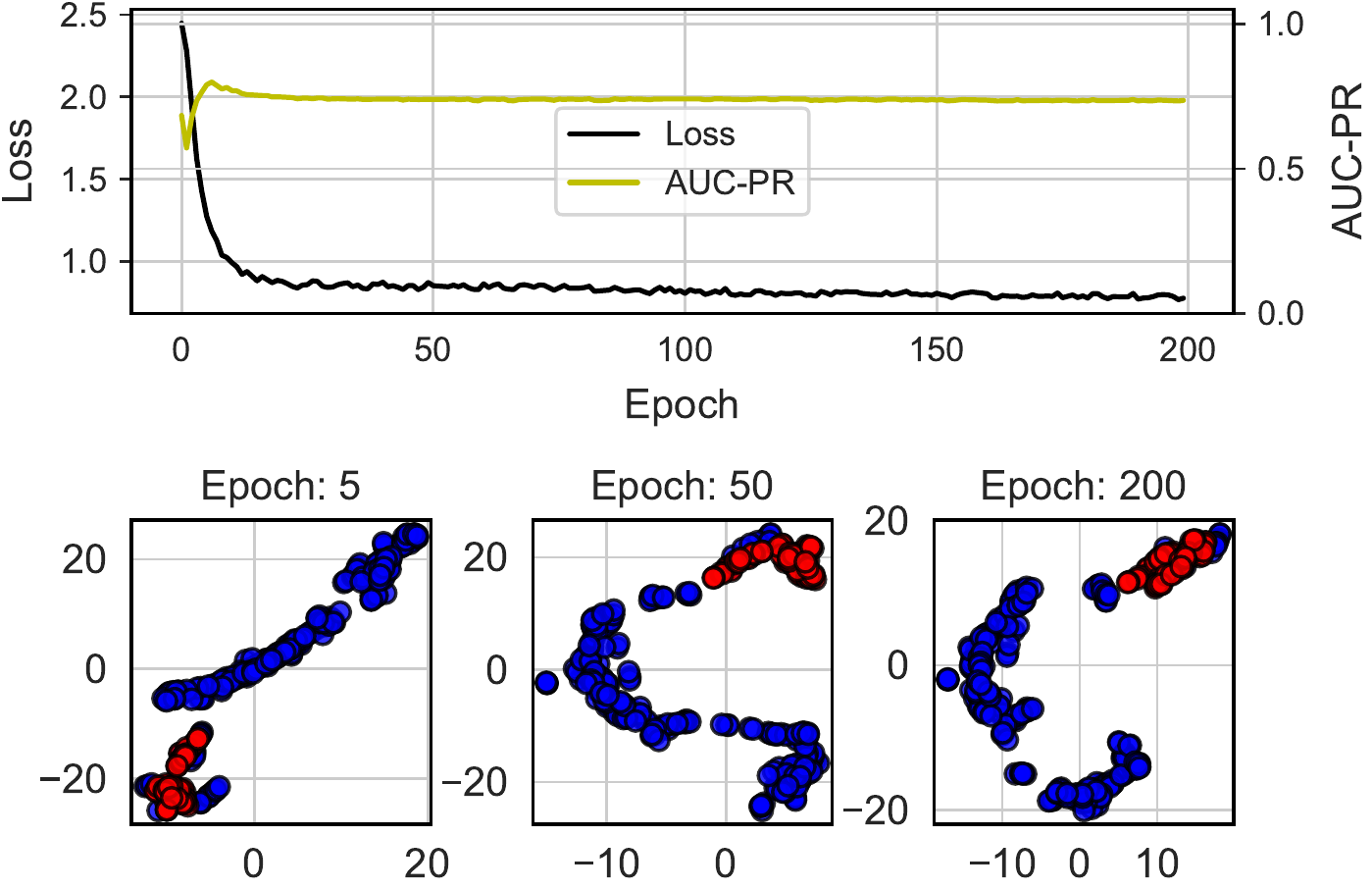}
         \caption{Deviation}
         \label{fig:}
     \end{subfigure}
     \hspace{0.2em}
     \begin{subfigure}[b]{0.3\textwidth}
         \centering
         \includegraphics[width=\textwidth]{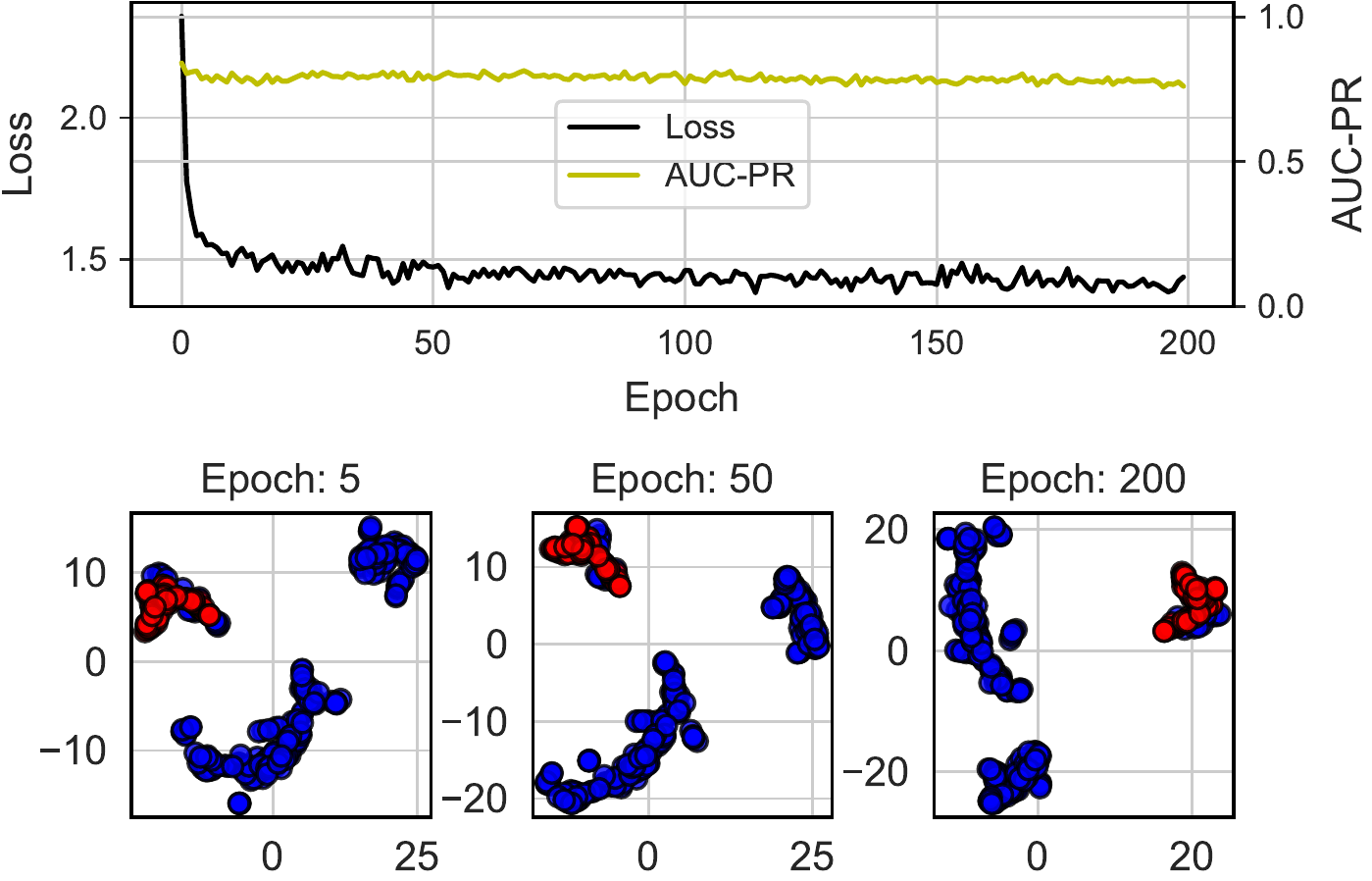}
         \caption{Ordinal}
         \label{fig:}
     \end{subfigure}
     \hspace{0.2em}
     \begin{subfigure}[b]{0.3\textwidth}
         \centering
         \includegraphics[width=\textwidth]{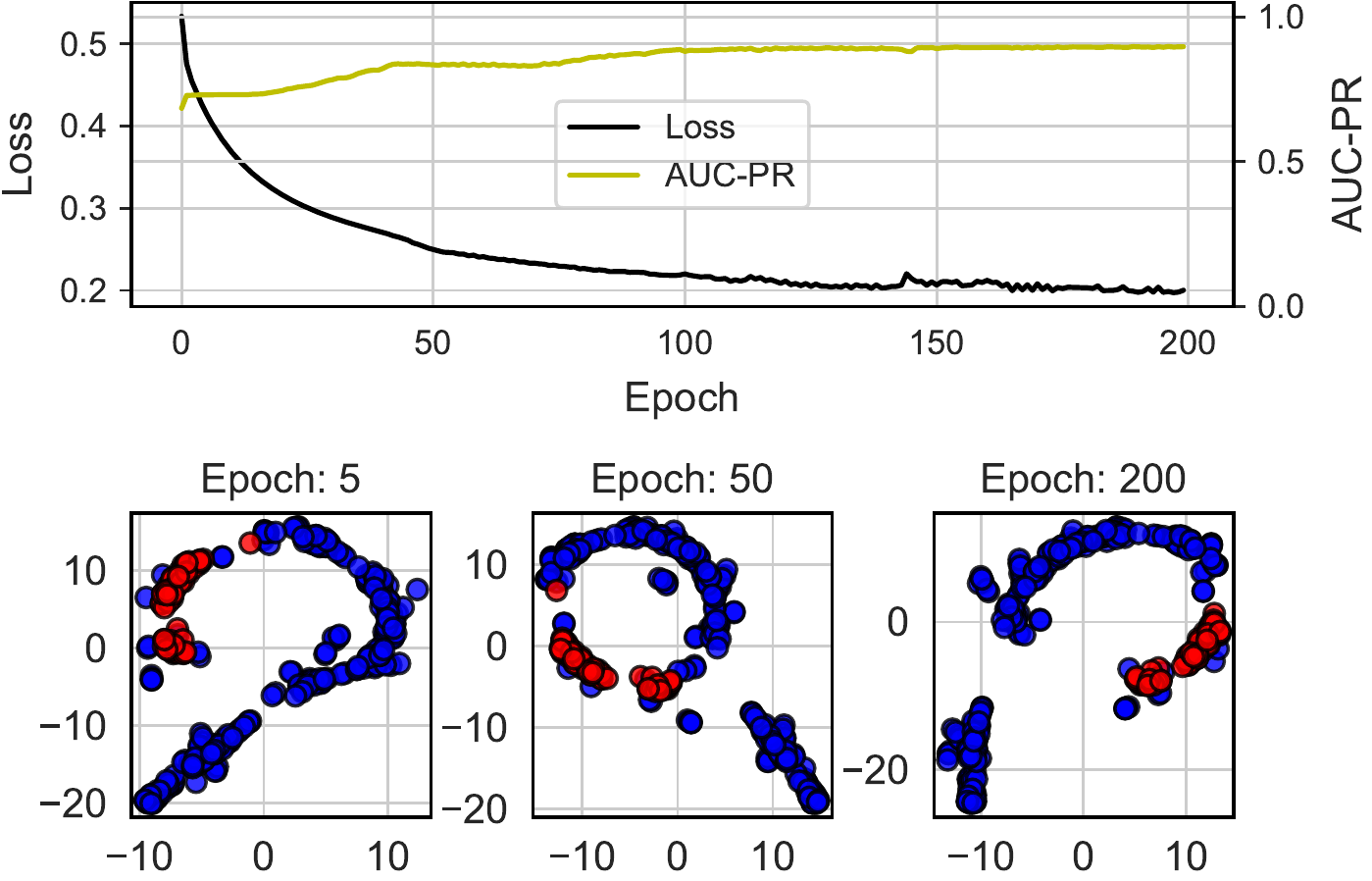}
         \caption{Overlap}
         \label{fig:}
     \end{subfigure}
     \hspace{0.2em}
    \end{minipage}
    \caption{Training loss along with the AUC-PR performance on testing set of different loss function based AD models, where the skin dataset is specified for comparison. The transformed embeddings of the input feature are demonstrated, which corresponds to 5, 50, and 200 training epochs, respectively.}
    \label{fig:skin_embedding}
\end{figure*}

In addition to the main paper that demonstrates the embedding variations on the vowels dataset in Section~\ref{exploration_real_world}, here we provide another example of the skin dataset, as shown in Figure~\ref{fig:skin_embedding}. Compared to the other loss functions, our proposed Overlap loss better retains the ringlike shape in the embedding of input feature while achieving satisfactory detection performance.

We follow \cite{steinbuss2021benchmarking, han2022adbench} to generate the following four types of synthetic anomalies, which are further used to evaluate different AD loss functions. The AUC-ROC results of loss function comparison on different types of anomalies also indicate that our proposed Overlap loss significantly outperforms other counterparts, as is shown in Table~\ref{tab:type_AUCROC}.

\setlist{nolistsep}
\begin{itemize}[leftmargin=*,noitemsep]
\item \textbf{Local anomalies} refer to the anomalies deviant from their local neighborhoods \cite{LOF}. GMM procedure \cite{milligan1985algorithm, steinbuss2021benchmarking} is used to generate synthetic normal samples, and then scale the covariance matrix $\hat{\boldsymbol{\Sigma}}=\alpha \hat{\boldsymbol{\Sigma}}$ by a scaling parameter $\alpha=5$ to generate local anomalies.

\item \textbf{Global anomalies} are generated from a uniform distribution $\text{Unif}\left(\alpha \cdot \min \left(\boldsymbol{x^{k}}\right), \alpha \cdot \max \left(\boldsymbol{x^{k}}\right)\right)$, where the boundaries are defined as the \textit{min} and \textit{max} of an input feature, e.g., $k$-th feature $\boldsymbol{x^{k}}$, and $\alpha=1.1$ controls the outlyingness of anomalies. 

\item \textbf{Dependency anomalies} refer to the samples that do not follow the dependency structure that normal data follows \cite{martinez2016fault}, i.e., the input features of dependency anomalies are assumed to be independent of each other. Vine Copula \cite{aas2009pair} method is applied to model the dependency structure of original data, where the probability density function of generated anomalies is set to complete independence by removing the modeled dependency (see \cite{martinez2016fault}).
KDE method estimates the probability density function of features and generates normal samples.

\item \textbf{Clustered anomalies}, also known as group anomalies \cite{lee2021gen}, 
exhibit similar characteristics \cite{meta_analysis, liu2022unsupervised}. We scale the mean feature vector of normal samples by $\alpha=5$, i.e., $\hat{\boldsymbol{\mu}}=\alpha \hat{\boldsymbol{\mu}}$, where $\alpha$ controls the distance between anomaly clusters and the normals, and use the scaled GMM to generate anomalies.
\end{itemize}

\begin{figure*}[h!]
     \centering
     \begin{center}
        \small
        \textbf{Clustered Anomaly}
        \vspace{0.02in}
     \end{center}
     \begin{subfigure}[b]{0.15\textwidth}
         \centering
         \includegraphics[width=\textwidth]{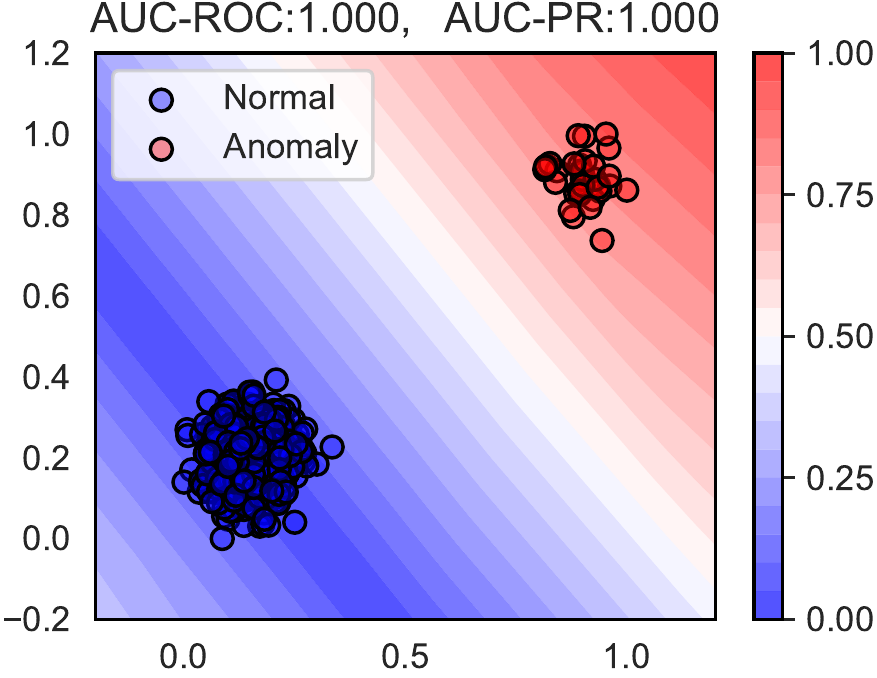}
         \caption{Minus}
         \label{fig:}
     \end{subfigure}
     \hspace{0.2em}
     \begin{subfigure}[b]{0.15\textwidth}
         \centering
         \includegraphics[width=\textwidth]{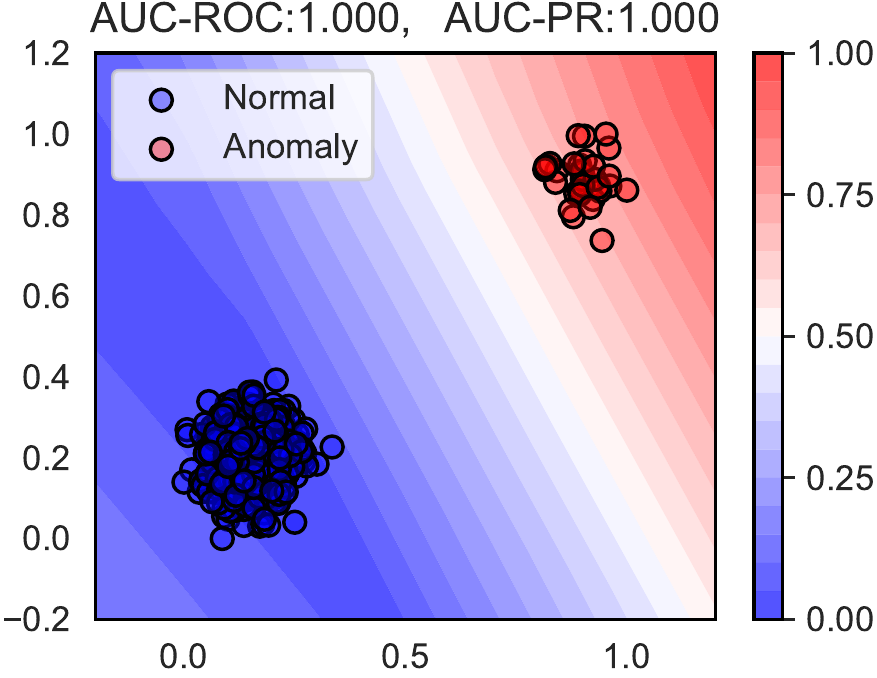}
         \caption{Inverse}
         \label{fig:}
     \end{subfigure}
     \hspace{0.2em}
     \begin{subfigure}[b]{0.15\textwidth}
         \centering
         \includegraphics[width=\textwidth]{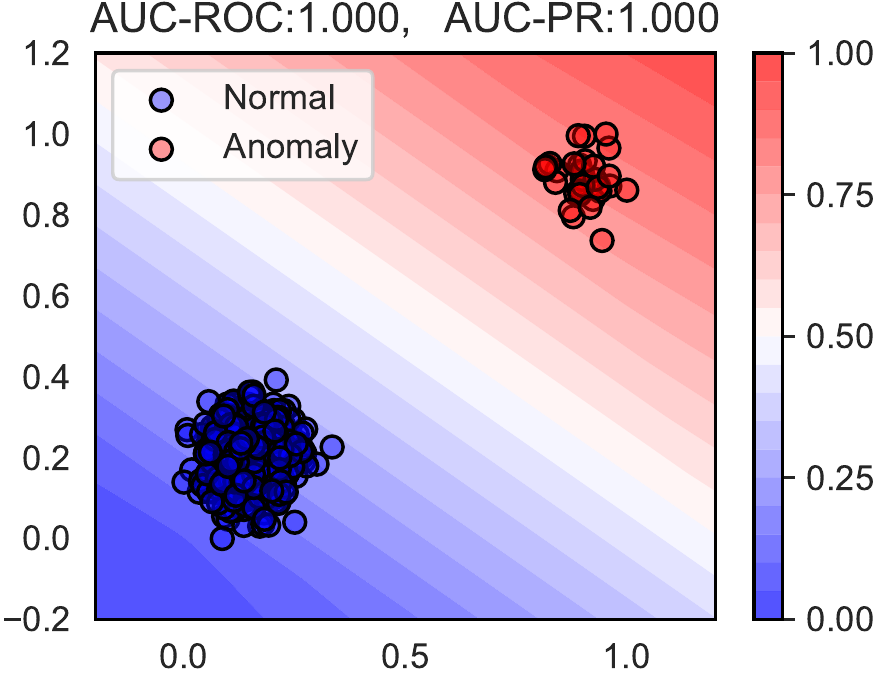}
         \caption{Hinge}
         \label{fig:}
     \end{subfigure}
     \hspace{0.2em}
     \begin{subfigure}[b]{0.15\textwidth}
         \centering
         \includegraphics[width=\textwidth]{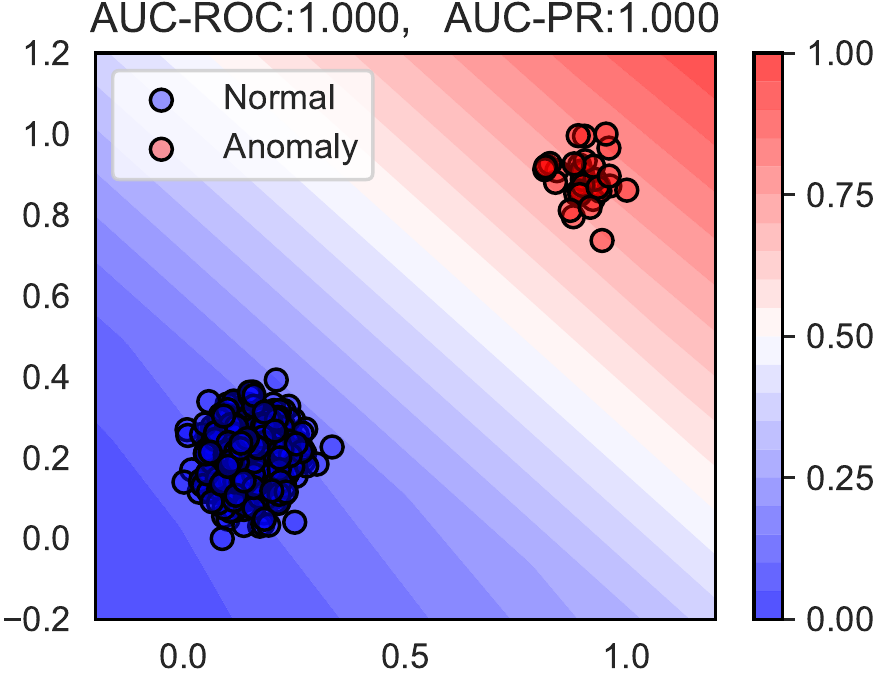}
         \caption{Deviation}
         \label{fig:}
     \end{subfigure}
     \hspace{0.2em}
     \begin{subfigure}[b]{0.15\textwidth}
         \centering
         \includegraphics[width=\textwidth]{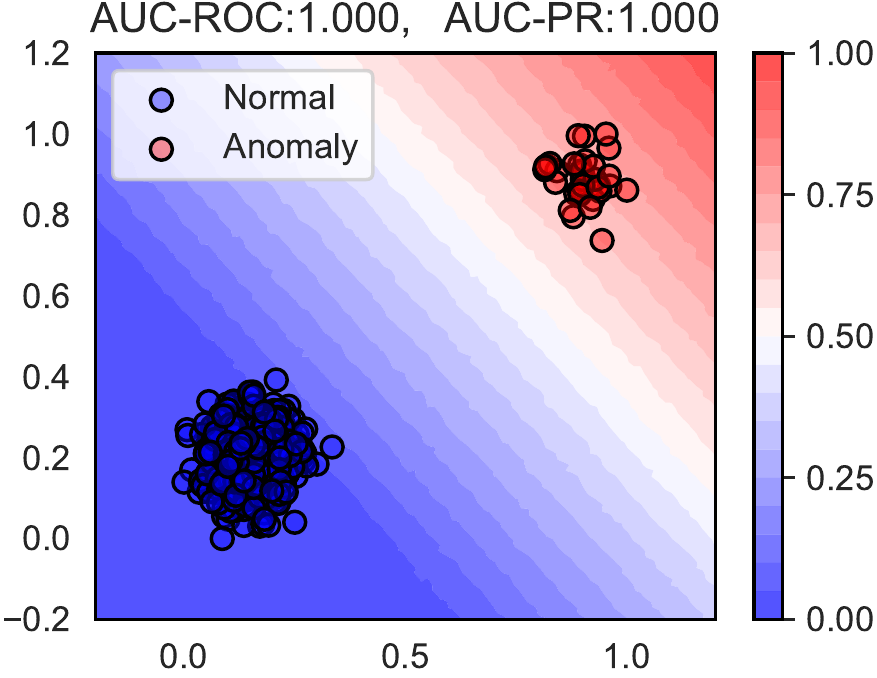}
         \caption{Ordinal}
         \label{fig:}
     \end{subfigure}
     \hspace{0.2em}
     \begin{subfigure}[b]{0.15\textwidth}
         \centering
         \includegraphics[width=\textwidth]{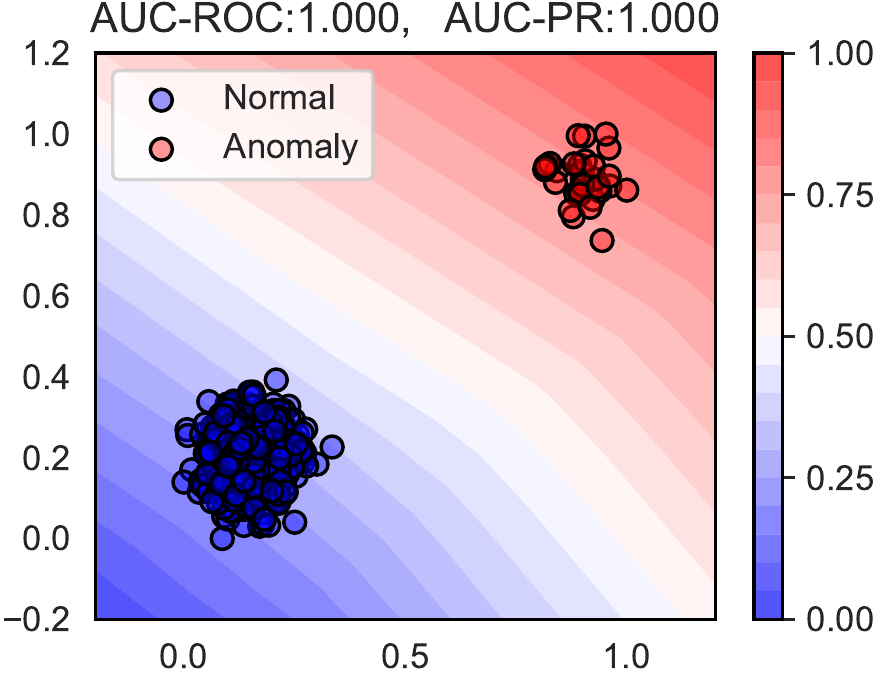}
         \caption{Overlap}
         \label{fig:}
     \end{subfigure}
     \hspace{0.2em}
     \vspace{0.06in}
     
     \begin{center}
        \small
        \textbf{Local Anomaly}
        \vspace{0.02in}
     \end{center}
     \begin{subfigure}[b]{0.15\textwidth}
         \centering
         \includegraphics[width=\textwidth]{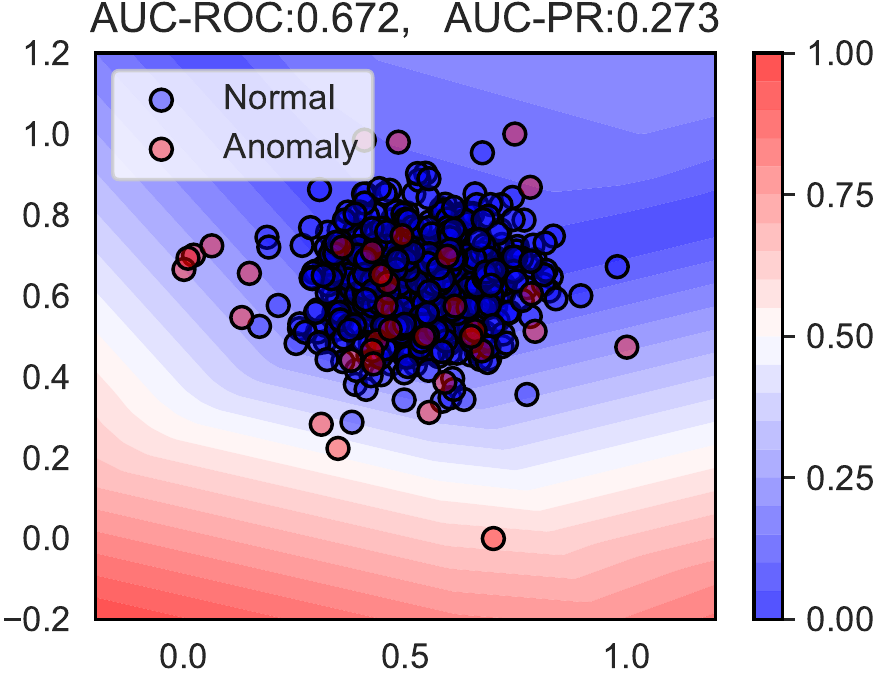}
         \caption{Minus}
         \label{fig:local_minus}
     \end{subfigure}
     \hspace{0.2em}
     \begin{subfigure}[b]{0.15\textwidth}
         \centering
         \includegraphics[width=\textwidth]{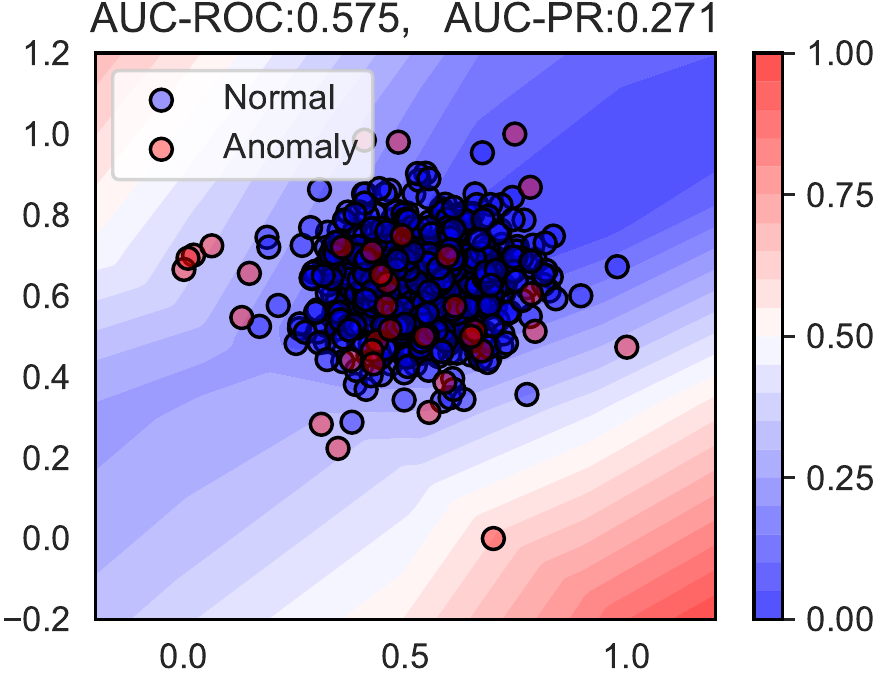}
         \caption{Inverse}
         \label{fig:}
     \end{subfigure}
     \hspace{0.2em}
     \begin{subfigure}[b]{0.15\textwidth}
         \centering
         \includegraphics[width=\textwidth]{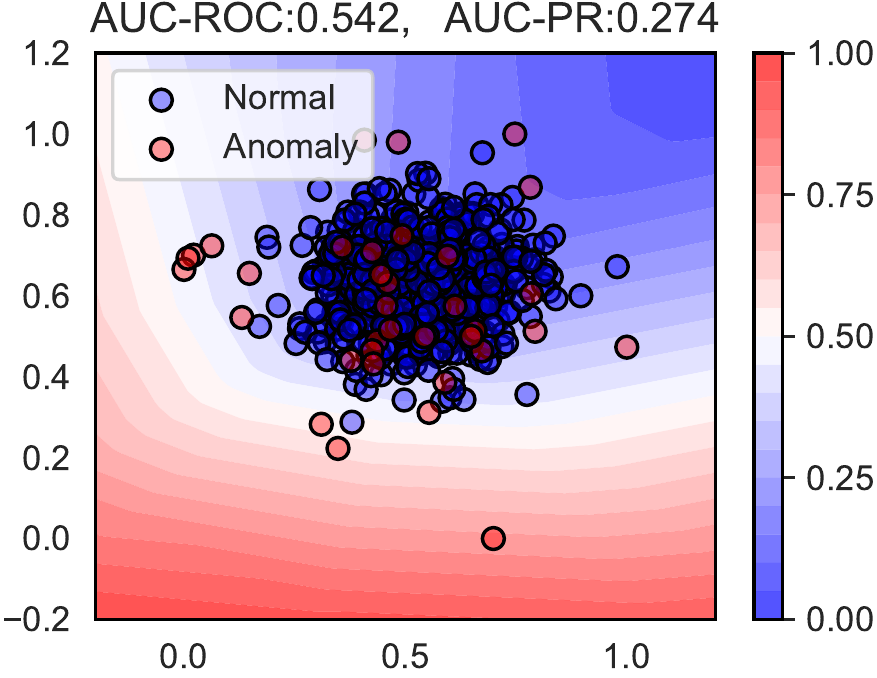}
         \caption{Hinge}
         \label{fig:}
     \end{subfigure}
     \hspace{0.2em}
     \begin{subfigure}[b]{0.15\textwidth}
         \centering
         \includegraphics[width=\textwidth]{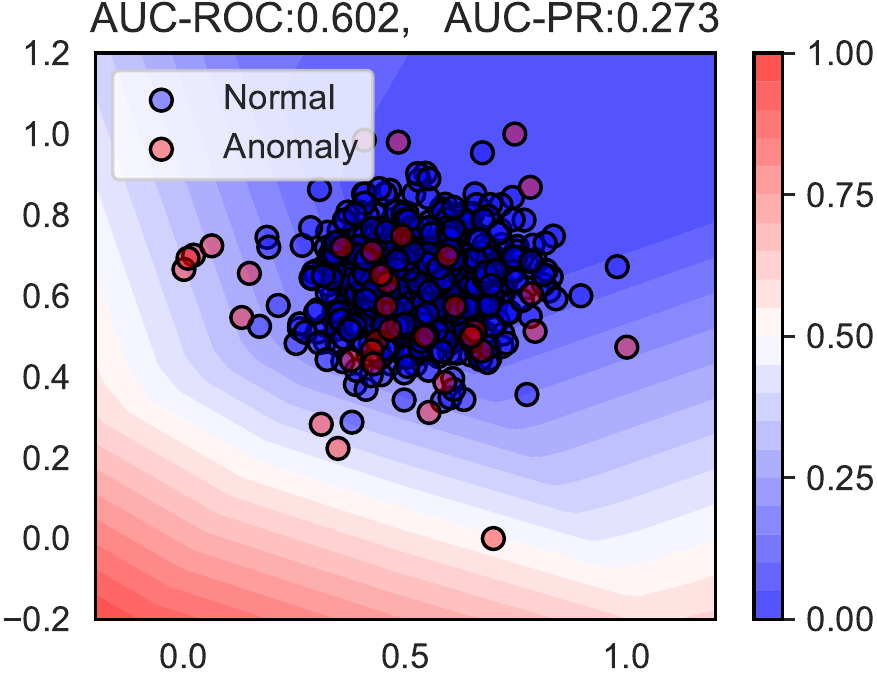}
         \caption{Deviation}
         \label{fig:}
     \end{subfigure}
     \hspace{0.2em}
     \begin{subfigure}[b]{0.15\textwidth}
         \centering
         \includegraphics[width=\textwidth]{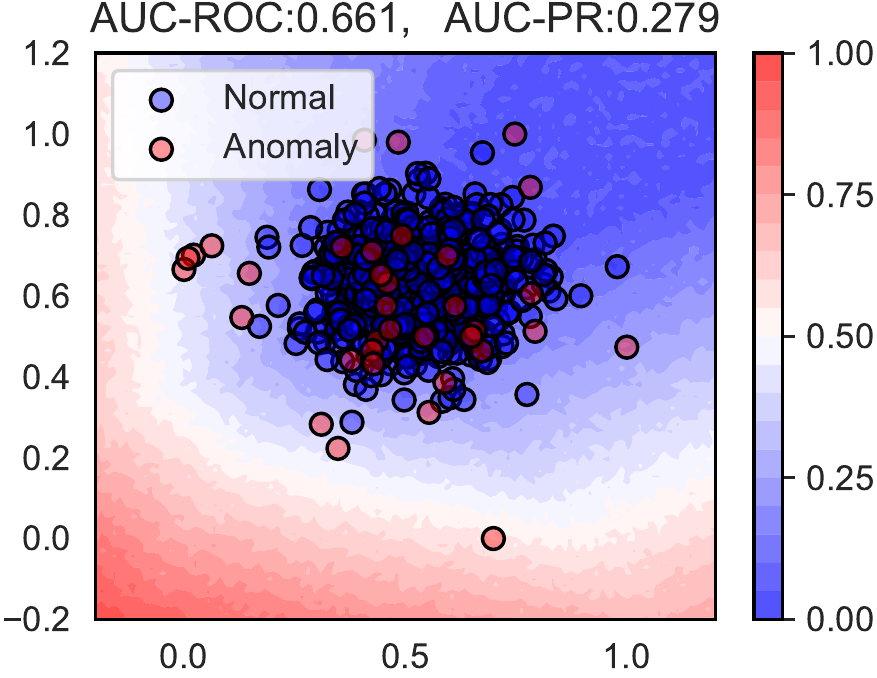}
         \caption{Ordinal}
         \label{fig:local_ordinal}
     \end{subfigure}
     \hspace{0.2em}
     \begin{subfigure}[b]{0.15\textwidth}
         \centering
         \includegraphics[width=\textwidth]{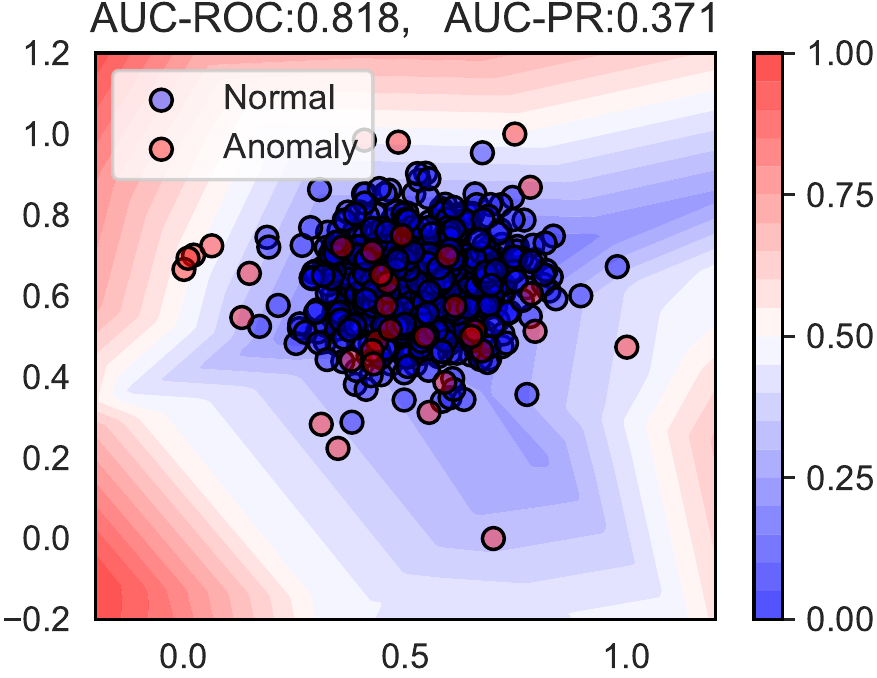}
         \caption{Overlap}
         \label{fig:local_score}
     \end{subfigure}
     \hspace{0.2em}
    
    \caption{Decision boundaries of different loss functions on the local anomalies. The output anomaly scores are normalized to $[0,1]$ for comparison. Both AUC-ROC and AUC-PR performances are displayed in the title above each subfigure.}
    \label{fig:synthetic local}
\end{figure*}

\begin{table}[h!]
\small
  \centering
  \caption{Loss function comparison on different types of anomalies generated based on the \ndatasets real-world datasets.}
  \label{tab:type_AUCROC}
  \centering
    \begin{tabular}{lcccc}
    \toprule
    \textbf{Loss}&
    \multicolumn{1}{l}{\textbf{Local}} &
    \multicolumn{1}{l}{\textbf{Global}} & \multicolumn{1}{l}{\textbf{Clustered}} & \multicolumn{1}{l}{\textbf{Dependency}} \\
    \midrule
    Minus & 0.629 & 0.936 & 0.996 & 0.738 \\
    Inverse & 0.547 & 0.823 & 0.937 & 0.570 \\
    Hinge & 0.607 & 0.938 & 0.997 & 0.761 \\
    Deviation & 0.588 & 0.959 & 0.990  & 0.652 \\
    Ordinal & 0.604 & 0.954 & 0.994 & 0.687 \\
    Overlap & 0.\textbf{742} & \textbf{0.981} & \textbf{0.998} & \textbf{0.847} \\
    \bottomrule
    \end{tabular}%
\end{table}%
We further investigate two case studies by generating visualized two-dimensional synthetic samples of the above local and clustered anomalies, as shown in Figure~\ref{fig:synthetic local}. The anomaly ratios of these two datasets are set to 5\%.
The results indicate that all the compared loss functions can correctly detect anomalies for the two-dimensional clustered anomalies (with 1.000 AUC-ROC and AUC-PR). This result can be expected since few labeled clustered anomalies can already represent similar behaviors of the entire clustered anomalies. For the local anomalies, however, we observe most of the compared loss functions perform poorly. In contrast, Overlap loss achieves better detection performance, and successfully learns a suitable decision boundary (see Figure~\ref{fig:local_score}), where the learned decision boundary fits well with the local anomalies that are often overlapped or surrounded by the normal samples.


\end{document}